\documentclass[11pt, a4paper, logo, twocolumn, copyright]{googledeepmind}

\pdftrailerid{redacted}

\makeatletter
\renewcommand\bibentry[1]{\nocite{#1}{\frenchspacing\@nameuse{BR@r@#1\@extra@b@citeb}}}
\makeatother

\usepackage{longtable}
\usepackage{dsfont}
\usepackage{gdm-colors}
\usepackage{multirow}
\usepackage{graphicx}

\usepackage[numbers, sort&compress, square]{natbib}

\makeatletter
\DeclareRobustCommand\onedot{\futurelet\@let@token\@onedot}
\def\@onedot{\ifx\@let@token.\else.\null\fi}

\def\eg/{\emph{e.g}\onedot} \def\Eg/{\emph{E.g}\onedot}
\def\ie/{\emph{i.e}\onedot} \def\Ie/{\emph{I.e}\onedot}
\def\cf/{\emph{c.f}\onedot} \def\Cf/{\emph{C.f}\onedot}
\def\etc/{\emph{etc}\onedot} \def\vs/{\emph{vs}\onedot}
\def\wrt/{w.r.t\onedot} \def\dof/{d.o.f\onedot}
\def\etal/{\emph{et al}\onedot}
\makeatother

\newcolumntype{L}[1]{>{\raggedright\let\newline\\\arraybackslash\hspace{0pt}}m{#1}} 

\title{PaliGemma: A versatile 3B VLM for transfer}

\correspondingauthor{lbeyer,xzhai@google.com}

\renewcommand{\today}{July 2024}

\author[*,$\dagger$]{Lucas Beyer}
\author[*]{Andreas Steiner}
\author[*]{André Susano Pinto}
\author[*]{Alexander Kolesnikov}
\author[*]{Xiao Wang}
\author[ \hspace{-0.7ex}]{Daniel Salz}
\author[ \hspace{-0.7ex}]{Maxim Neumann}
\author[ \hspace{-0.7ex}]{Ibrahim Alabdulmohsin}
\author[ \hspace{-0.7ex}]{Michael Tschannen}
\author[ \hspace{-0.7ex}]{Emanuele Bugliarello}
\author[ \hspace{-0.7ex}]{Thomas Unterthiner}
\author[ \hspace{-0.7ex}]{Daniel Keysers}
\author[ \hspace{-0.7ex}]{Skanda Koppula}
\author[ \hspace{-0.7ex}]{Fangyu Liu}
\author[ \hspace{-0.7ex}]{Adam Grycner}
\author[ \hspace{-0.7ex}]{Alexey Gritsenko}
\author[ \hspace{-0.7ex}]{Neil Houlsby}
\author[ \hspace{-0.7ex}]{Manoj Kumar}
\author[ \hspace{-0.7ex}]{Keran Rong}
\author[ \hspace{-0.7ex}]{Julian Eisenschlos}
\author[ \hspace{-0.7ex}]{Rishabh Kabra}
\author[ \hspace{-0.7ex}]{Matthias Bauer}
\author[ \hspace{-0.7ex}]{Matko Bošnjak}
\author[ \hspace{-0.7ex}]{Xi Chen}
\author[ \hspace{-0.7ex}]{Matthias Minderer}
\author[ \hspace{-0.7ex}]{Paul Voigtlaender}
\author[ \hspace{-0.7ex}]{Ioana Bica}
\author[ \hspace{-0.7ex}]{Ivana Balazevic}
\author[ \hspace{-0.7ex}]{Joan Puigcerver}
\author[ \hspace{-0.7ex}]{Pinelopi Papalampidi}
\author[ \hspace{-0.7ex}]{Olivier Henaff}
\author[ \hspace{-0.7ex}]{Xi Xiong}
\author[ \hspace{-0.7ex}]{Radu Soricut}
\author[ \hspace{-0.7ex}]{Jeremiah Harmsen}
\author[*,$\dagger$]{Xiaohua Zhai}

\affil[*]{Core team}
\affil[$\dagger$]{Project lead}

\begin{abstract}
PaliGemma is an open Vision-Language Model (VLM) that is based on the SigLIP-So400m vision encoder and the Gemma-2B language model.
It is trained to be a versatile and broadly knowledgeable base model that is effective to transfer. It achieves strong performance on a wide variety of open-world tasks.
We evaluate PaliGemma on almost 40 diverse tasks including standard VLM benchmarks, but also more specialized tasks such as remote-sensing and segmentation.
\end{abstract}

\begin{document}

\maketitle

\section{Introduction}

\begin{figure*}[t]
  \centering
  \includegraphics[width=0.9\textwidth]{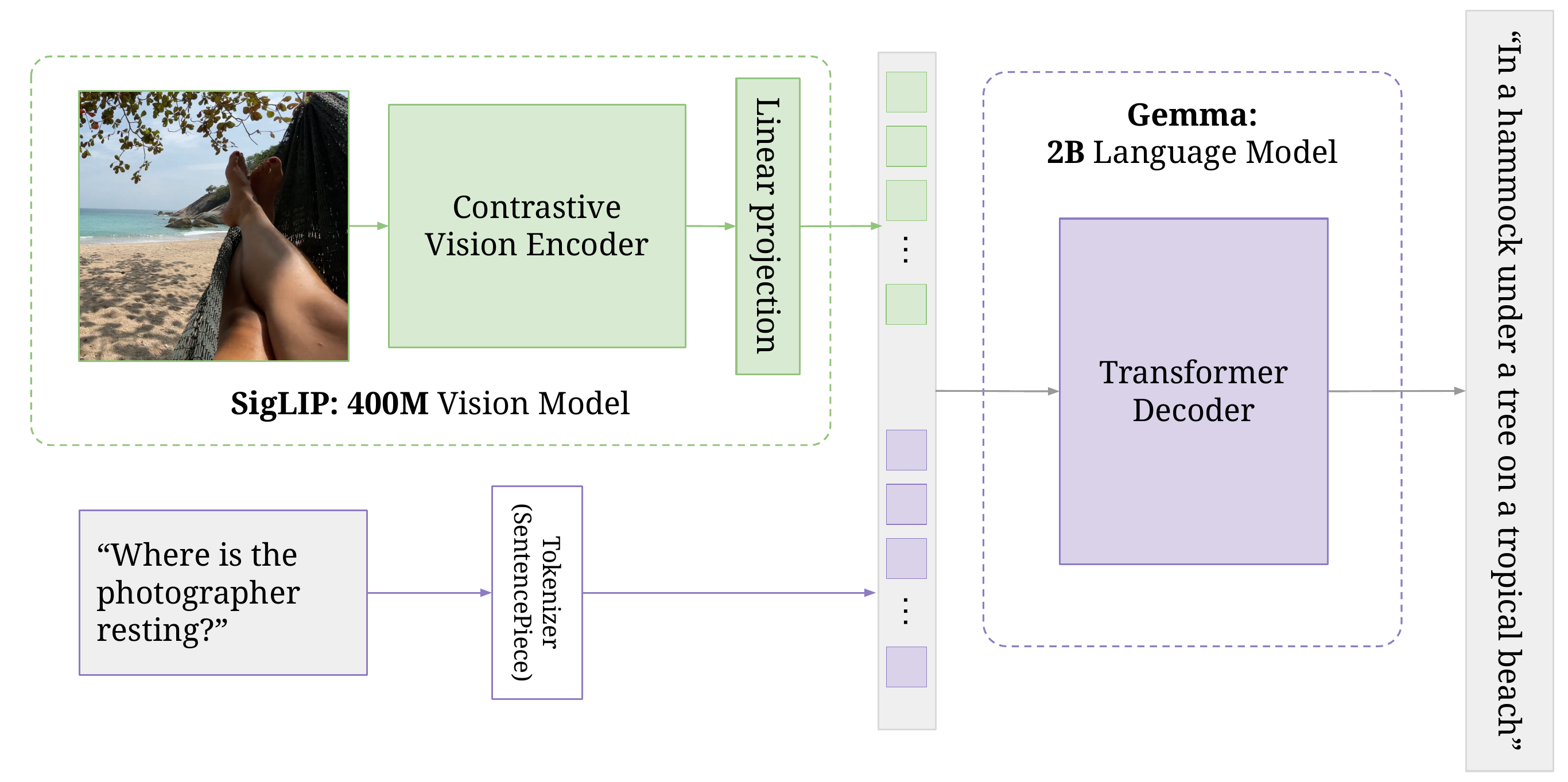}
  \caption{PaliGemma's architecture: a SigLIP image encoder feeds into a Gemma decoder LM.}
  \label{fig:architecture}
\end{figure*}

PaliGemma is an open model, continuing the line of PaLI vision-language models in a combination with the Gemma family of language models.

PaLI is a series of state-of-the-art vision-language models, starting with the first PaLI~\cite{pali} showing promising scaling results up to 17\,B, using classification pretrained ViT~\cite{scalingvit} and mT5~\cite{xue2020mt5} language model.
PaLI-X~\cite{palix} and PaLM-E~\cite{palm_e} then pushed this further, combining ViT-22\,B~\cite{vit22b} and a 32\,B UL2~\cite{ul_2} language model or the 540\,B PaLM~\cite{palm} language model, respectively, and getting further increased performance on vision-language tasks, albeit saturating performance on standard image classification and retrieval tasks.
Finally, PaLI-3~\cite{pali3} demonstrates that through better pretraining with SigLIP~\cite{siglip} and more careful multimodal data curation, a 2\,B vision and 3\,B language model (\ie/ a 5\,B vision-language model) matches the 10x larger PaLI-X and 100x larger PaLM-E across most benchmarks.

PaliGemma continues this trend, combining the 400\,M SigLIP and the 2\,B Gemma models~\cite{gemma} into a sub-3\,B VLM that still maintains performance comparable to PaLI-X, PaLM-E, and PaLI-3.

Gemma~\cite{gemma} is a family of auto-regressive decoder-only open large language models built from the same research and technology used to create the Gemini~\cite{gemini} models.
The models come in different sizes (2\,B, 7\,B), both pretrained and instruction fine-tuned. PaliGemma uses the 2\,B pretrained version.

The main goal of our work is to provide a versatile base VLM.
Hence, we show that it reaches state-of-the-art results not only on standard COCO captions, VQAv2, InfographicVQA and others, but also on more exotic Remote-Sensing VQA, TallyVQA, several video captioning and QA tasks, as well as referring expression \emph{segmentation} (see full task list in Appendix~\ref{sec:app:tasks}).

\section{Related work}

Over the course of the past few years, vision-language models have gained considerable importance in computer vision.
The first generation, spearheaded by CLIP~\cite{clip} and ALIGN~\cite{align} by scaling up ConVIRT~\cite{zhang2020contrastive} and VirTex~\cite{desai2021virtex}, is an extension of large-scale classification pretraining~\cite{bit,scalingvit}, to leverage all data from the web without the need for onerous human labeling, replacing a fixed and large set of classes by a caption embedding instead.
The caption embeddings are mostly obtained using language encoders (similar to BERT~\cite{bert}) and allow to open up the vocabulary of classification and retrieval tasks. 
The second generation, akin to T5~\cite{t5} in language, is a unification of captioning and question-answering tasks via generative encoder-decoder modeling~\cite{cho21unifying,wang2021simvlm,zhou20unified,tsimpoukelli21frozen}, often backed by the progress in generative language models.
These were then further scaled up by, among others, Flamingo~\cite{alayrac2022flamingo}, BLIP-2~\cite{blip2} and, PaLI~\cite{pali}.
Finally, most recent works~\cite{openai2024gpt4,llava1,gemini,moondream} perform an additional ``instruction tuning'' step that is intended to make the raw model more user-friendly.
In addition to building systems, several recent more systematic studies~\cite{mm1,idefics2,cambrian1} aim to find out what really matters in VLMs.
PaliGemma is an open base VLM without instruction tuning, and this report answers a few more questions regarding what matters.
More discussion in Appendix~\ref{sec:app:relwork}.

\section{Model}

In this section we present details about PaliGemma's architecture and training.
Several of our decisions are further ablated in Section~\ref{sec:abl}.

At a high level, PaliGemma is a VLM, taking as input one or more images, and a textual description of the task (the prompt or question, which we often refer to as the {\tt prefix}).
PaliGemma then autoregressively generates a prediction in the form of a text string (the answer, which we often refer to as the {\tt suffix}).

This simple image+text in, text out API is flexible enough to cover many standard tasks, such as image classification, captioning, visual question-answering and dialogue.
Additionally, as shown in the literature, by converting more complex structured outputs into ``text'', this API can also cover more tasks such as: 
detection~\cite{pix2seq}, instance segmentation~\cite{pali3,locca}, panoptic segmentation, depth prediction, colorization, and many more~\cite{uvim,unifiedio,xdecoder}.
This conversion can be hand-engineered and task-specific, such as done in pix2seq~\cite{pix2seq} for detection, or learned as is the case for segmentation~\cite{uvim} and dense output tasks in general.

During PaliGemma's pretraining, we limit ourselves to ``text'' covering natural language, object detection, and instance segmentation, but this API remains versatile and the pretrained models can be finetuned for other output types.

\subsection{Architecture}\label{sec:model:architecture}
PaliGemma consists of three components:

\begin{itemize}

\item An image encoder, for which we use a publicly available SigLIP~\cite{siglip} checkpoint, specifically the ``shape optimized''~\cite{sovit} ViT-So400m image encoder. This model was contrastively pretrained at large scale via the sigmoid loss, and has shown state-of-the-art performance, especially for its small size.

\item A decoder-only language model, for which we use the publicly available Gemma-2B v1.0~\cite{gemma} raw pretrained checkpoint, which strikes a great balance between performance and size.
As we will show, this language model is good enough to match or surpass the performance of VLMs using much larger language models, including previous PaLIs.

\item A linear layer projecting SigLIP's output tokens into the same dimensions as Gemma-2B's vocab tokens, so they can be concatenated.
In early experiments, we found that more complicated alternatives (\eg/ MLPs) do not provide a clear advantage, and hence decided to use the simplest option (Sec~\ref{sec:abl:connector}).

\end{itemize}

\begin{figure}[t]
  \centering
  \includegraphics[width=0.9\columnwidth]{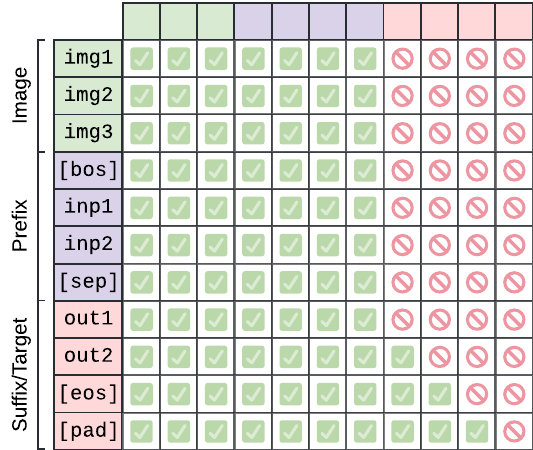}
  \caption{PaliGemma's Prefix-LM masking: block attention throughout image and prefix, autoregressive attention on the suffix. Each square indicates whether the row can attend to the column.}
  \label{fig:attention_mask}
\end{figure}

The image is passed through the image encoder, which turns it into a sequence of $N_{\text{img}}$ tokens.
The text is converted into $N_{\text{txt}}$ tokens using Gemma's SentencePiece~\cite{kudo-richardson-2018-sentencepiece} tokenizer, and embedded with Gemma's vocabulary embedding layer.
The image tokens are projected with the (zero initialized) linear projection.
Then the sequence of input tokens to the decoder is created as follows (and also as visible in Figure~\ref{fig:attention_mask}):

{\footnotesize
\begin{samepage}\begin{verbatim}  tokens = [image tokens...,
            BOS, prefix tokens..., SEP,
            suffix tokens..., EOS, PAD...]
\end{verbatim}\end{samepage}
}


We always resize the image to a fixed square size (224, 448, or 896 pixels).
This leads to a fixed number of image tokens per model variant (respectively 256, 1024, or 4096 tokens), which we place in the front, making image tokens straightforward to interpret without the need for special location markers.
The BOS token then marks the start of text tokens.
We use \texttt{\textbackslash n} as SEP token, it does not appear in any of our prefixes.
We also tokenize SEP separately to avoid it being merged (by the tokenizer) with either the end of the prefix or the beginning of the suffix.
In order to maximize model capacity for such a small model, we have full (unmasked) attention on the whole input, \ie/ the image and prefix tokens.
In this way, image tokens can also "lookahead" at the task at hand (prefix) in order to update their representation.
The suffix is our output and necessarily covered by an auto-regressive mask, including the PAD tokens.
When we mention sequence length ($N_{\text{txt}}$), we typically mean prefix and suffix combined, ignoring image tokens.

\subsection{Pretraining}

The training of PaliGemma follows the same steps as previous PaLI models, with only small modifications. Training consists of several stages, which we detail in this section:

\begin{itemize}
\item {\bf Stage0:} Unimodal pretraining - we use existing off-the-shelf components.
\item {\bf Stage1:} Multimodal pretraining - long pretraining on a carefully chosen mixture of multimodal tasks. Notably, nothing is frozen.
\item {\bf Stage2:} Resolution increase - short continued pretraining at higher resolution.
\item {\bf Stage3:} Transfer - turn the base model into a task-specific specialist.
\end{itemize}

\subsubsection{Stage0: Unimodal pretraining}

First, the unimodal components of the model are pretrained individually, in order to benefit from their well-studied and scaled training recipes.
For PaliGemma specifically, we do not perform any custom unimodal pretraining, instead relying on existing publicly available checkpoints.

Following PaLI-3's strong experimental results, we use a SigLIP image encoder.
While PaLI-3 (and others~\cite{alayrac2022flamingo,chen2023internvl}) use a large image model such as ViT-G, we use the much smaller but similarly strong ``shape optimized'' ViT-So400m model.

PaLI traditionally uses an encoder-decoder language model; however all recently publicly released language models are decoder-only Transformers.
We opt for the Gemma-2B model, which strikes a good balance between size and performance.
Larger language models, such as the popular 7\,B or 70\,B sizes, are often significantly better at tasks like mathematical reasoning.
However, PaLI-3 has shown that across a wide range of vision-language tasks, a well-trained small 5\,B model (2\,B vision + 3\,B language) can attain the same performance as the much larger 55\,B PaLI-X (22\,B vision + 32\,B language) and 562\,B PaLM-E (22\,B vision + 540\,B language), including tasks such as ScienceQA.
With PaliGemma we continue this push for smaller models and show that we can keep the same performance with less than 3\,B total parameters.

\subsubsection{Stage1: Multimodal pretraining}

In this stage, we combine the unimodal models as explained in Section~\ref{sec:model:architecture} and train the whole model on a broad mixture of large-scale vision-language tasks.
Contrary to most recent VLMs, our core goal is to train a base model that fine-tunes well to a wide range of tasks, not merely to align the modalities.
Intuitively, we want a mix of tasks which force the model to acquire a broad range of ``skills'', regardless of the task's user (or benchmark) friendliness out of the box.
More on this in Section~\ref{sec:pretrain_mix}.

It is common practice, also followed by previous PaLI versions, to keep the image encoder frozen during the first multimodal pretraining stage. This is partially due to findings as in LiT~\cite{lit} reporting multimodal tuning of pretrained image encoders degrading their representations.
However, more recent work such as CapPa~\cite{cappa} and LocCa~\cite{locca} have shown that captioning and other harder-to-learn tasks can provide valuable signal to image encoders, allowing them to learn spatial and relational understanding capabilities which contrastive models like CLIP or SigLIP typically lack.
Hence, again in the spirit of learning more skills during pretraining, we depart from common practice and do not freeze the image encoder.
However, the challenges outlined in LiT remain.
In order to avoid destructive supervision signal from the initially unaligned language model,
we use a slow linear warm-up for the image encoder's learning-rate (Figure~\ref{fig:schedule}), which ensures that the image encoder's quality is not deteriorated from the initially misaligned gradients coming through the LLM.

We train Stage1 at resolution 224px (hence, $N_\text{img}=256$ image tokens) and sequence length $N_\text{txt}=128$ for a total of 1 billion examples.
While we provide an ablation in Section~\ref{sec:abl:duration} showing that a 10x to 30x shorter Stage1 still provides good results on popular benchmarks,
we wish to imbue as much visual knowledge to the base model as possible, and cover a broad set of concepts, cultures, and languages~\cite{liu-etal-2021-visually,bugliarello-etal-2022-iglue,qiu-etal-2022-multilingual,nofilter,nguyen2024multilingual,geigle2024african,zhang2024vlmbadclf}.

\subsubsection{Stage2: Resolution increase}

The model resulting from Stage1 is already a useful base model for many tasks (see example images in Appendix~\ref{sec:app:tasks}).
However, it only understands images at $224 \times 224$ pixel resolution, which is too small for several tasks.
For instance, detection and segmentation of smaller objects, and tasks related to reading smaller texts such as charts, infographics, or documents, all strongly benefit from higher resolution (see Table~\ref{tbl:main_results}).
Hence, we train two further model checkpoints for increased resolution, first to $448 \times 448$ and then to $896 \times 896$ pixel resolution.

Since stage1 took care of providing the model with a broad set of knowledge and skill, stage2 can focus on extending the model's ability to parse higher-resolution images.
We thus run Stage2 with fewer total examples, while increasing the cost and information density of each example.
For resolution 448, we train for an additional 50\,M examples, and for resolution 896, we add another 10\,M examples.

For simplicity, Stage2 consists of the exact same mixture of tasks and datasets as Stage1, but with significantly increased sampling of tasks that require high resolution.
Additionally, these upweighted tasks all can be modified to provide much longer suffix sequence lengths.
For instance, for OCR tasks, we can simply request the model to read \emph{all} text on the image in left-to-right, top-to-bottom order.
For detection and segmentation tasks, we can request the model to detect or segment \emph{all} objects for which annotation is provided.
Hence, we also increase the text sequence length to $N_\text{txt}=512$ tokens.

While PaLI has always had this resolution increasing stage, and for image classification the importance of resolution is long known~\cite{touvron2022fixres,bit}, several recent works~\cite{visheratin2024curse,wu2023vstar,mm1} have raised the importance of resolution in VLMs too.
We add to this body of knowledge by providing several ablation studies regarding Stage2 in Section~\ref{sec:abl:res}.

\subsubsection{Stage3: Transfer}\label{sec:transfer}

The result of Stages~1 and~2 is a family of three PaliGemma checkpoints, at 224px, 448px, and 896px resolution,
which are pre-equipped with broad visual knowledge.
However, these checkpoints are not ``user (or benchmark) friendly'' as their pretraining has focused solely on density of learning signal, as opposed to usable interface.

These base models need to be transferred to serve their intended final purpose.
That could take the form of fine-tuning on a specific, specialized task, such as COCO Captions, Remote Sensing VQA, Video Captioning, or InfographicQA. Adapt to new inputs such as multiple images (NLVR2) or bounding boxes draw in the image (WidgetCap).
Or it could take the form of instruction~\cite{llava1} or even chat~\cite{huang2024sparkles} tuning.

To show the effectiveness of the base models, we transfer them to a wide range of individual academic benchmarks, using a simple unified transfer recipe with few hyper-parameters.
And to showcase the versatility beyond academic tasks, we also provide a ``mix'' transfer checkpoint, which transfers to a subset of these tasks at the same time, along with detailed captioning and long question-answering data.
While this is not instruction tuning, it is a step in that direction.

We also transfer PaliGemma to tasks which take multiple images as input.
NLVR2 is one such task, which asks one question about two images, and requires looking at both to give the correct answer.
Other such tasks are standard short-video understanding tasks subsampled to 16 frames.
In all these cases, we follow PaLI-3 and encode each image separately, then concatenate the image tokens without any special separator or embedding tokens.
Thus, 16 frames at 224px resolution result in $N_\text{img}=4096$ image tokens, the same amount as a single image at 896px resolution.

For all transfers, we perform fine-tuning of all the model parameters.
The hyper-parameters we modify per-task are the following, in decreasing order of importance:
\begin{itemize}
\item Resolution (\ie/ checkpoint): {\bf 224}, 448, 896.
\item Epochs: {\bf 1, 3, 10}, 30, 100.
\item Learning-rate: 3e-5, {\bf 1e-5}, 3e-6.
\item Label-smoothing: {\bf 0.0}, 0.1, 0.3.
\item Dropout in the LLM: {\bf 0.0}, 0.1, 0.3.
\item Weight decay: {\bf 0.0} or 0.1 $\times$ learning-rate.
\item Freeze ViT: {\bf false}, true.
\item Beam-search may benefit captioning.
\end{itemize}
The above are typical values we suggest exploring, with the recommended initial attempt value in bold.
We provide the best setting for each individual task in Appendix~\ref{sec:app:hparams}.
We study the sensitivity to transfer hyper-parameters in Section~\ref{sec:abl:simple}, and the ``transferability'' in general in Section~\ref{sec:abl:transferability}, showing that good results can be achieved with the aforementioned initial attempt values.

\subsubsection{Pretraining task mixture}\label{sec:pretrain_mix}

Just like for previous PaLI models, the pretraining (Stage1 and Stage2) is designed to result in a model that transfers well, not necessarily a model that is usable out of the box (``0 shot'').
The intuition here is that we want a mix of tasks which force the model to acquire a broad range of ``skills''.
We prefix each task with its unique prefix to avoid conflicting learning signals across skills~\cite{beyer2023litdecoder}.
At transfer time (Stage3), the model then merely needs to recognize which skill is useful for the task, and rewire itself to use that while following the output syntax and vocabulary of the task.
In our experience, these can all be done relatively quickly and based on few examples (Section~\ref{sec:abl:transfer_lowdata}).
We do not use any of our transfer datasets during pretraining, and furthermore remove all near-duplicates of their images from the pretraining datasets~\cite{bit}. 

Largely following previous PaLI works, these are the pretraining tasks:

{\tt caption \{lang\}}\\
We include the simple captioning objective on various datasets, including WebLI in over 100 languages, and CC3M-35L. Previous PaLIs use an encoder-decoder language model with the SplitCap objective, however for PaliGemma with the decoder-only language model, plain captioning is a more informative and simpler objective.

{\tt ocr}\\
Concatenation (in raster order) of all text on the image transcribed by a public OCR system. Potentially skipping random snippets of OCR in order to fit sequence length without biasing recognition towards the beginning of raster order.

{\tt answer en \{question\}}\\
Generated VQA on CC3M-35L following~\cite{Changpinyo} with questions in 35 languages but English answers.
Additionally, English-only object-centric questions on OpenImages following~\cite{Piergiovanni}:\\
listing: {\tt What objects are in the image?},\\
presence: {\tt Is \{thing\} in the image?},\\
multi-object presence: {\tt Which of \{thing\}, \{thing\}... are in the image?},\\
and newly, counting: {\tt How many \{thing\}?}.

{\tt question \{lang\} \{English answer\}}\\
Generated VQG on CC3M-35L following~\cite{Changpinyo} generating questions in 35 languages, for a given English answer.

{\tt detect \{thing\} ; \{thing\} ; ...}\\
Multi-object detection similar to Pix2Seq~\cite{pix2seq} on generated open-world data via pseudo-labeling as described in OWL-ViTv2~\cite{owlvitv2}.

{\tt segment \{thing\} ; \{thing\} ; ...}\\
Multi-object instance segmentation as in PaLI-3~\cite{pali3} on generated open-world data similar to OWL-ViTv2~\cite{owlvitv2} and SAM~\cite{sam}.

{\tt caption <ymin><xmin><ymax><xmax>}\\
Grounded captioning of what is in the box, following LocCa~\cite{locca}. The box is indicated by the same location tokens as used in detection and segmentation: normalized image coordinates binned to 1024 tokens.

Notably distinct from the widely used LLaVa's GPT-4 generated instruction following data, none of PaliGemma's pretraining tasks is the output of a larger commercial VLM.

Finally, we believe that it is important to detect and remove all images in our pretraining datasets which are near-duplicates of images in the transfer tasks we evaluate in this report~\cite{bit}, as well as a few more popular computer vision benchmarks.
Doing so, we more accurately capture PaliGemma's capability to transfer to new tasks.

\subsubsection{Other pretraining details}\label{sec:pretrain_details}

\begin{figure}[t]
  \centering
  \includegraphics[width=\columnwidth]{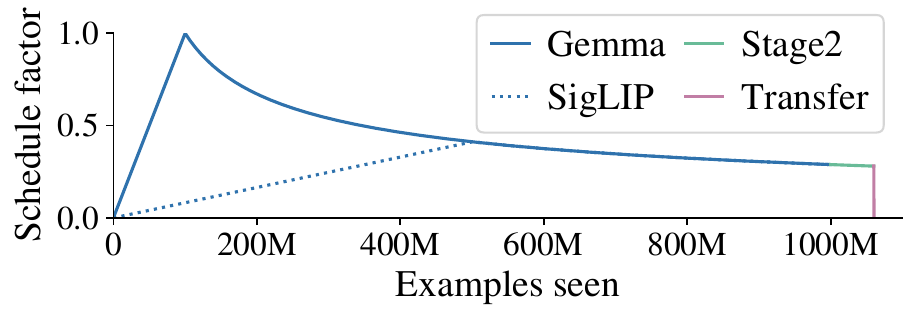}
  \caption{Learning-rate schedule across stages.}
  \label{fig:schedule}
\end{figure}

Throughout pretraining, we use an "infinite" learning-rate schedule following~\cite{scalingvit}, which provides a straightforward way of chaining several stages without decaying the learning-rate between them.
Figure~\ref{fig:schedule} shows the full schedule: pretraining is one continuous rsqrt curve for all stages.
The transfer can then act as a cooldown, fully annealing the learning rate.
We recommend transferring with a simple setup that tunes the full model using a cosine learning-rate schedule with a short linear warm-up and decaying to zero.
This is not well represented by Figure~\ref{fig:schedule} due to its comparatively short duration.

The model was entirely trained in the open-source {\tt big\_vision} codebase~\cite{big_vision} on Cloud TPUv5e~\cite{tpucloud}. However, some of the pretraining datasets remain private.
During training, we partition data, as well as model parameters and optimizer state (Zero-DP style~\cite{zero}) across all available devices using JAX~\cite{jax2018github} with GSPMD~\cite{gspmd}. This fully-sharded data-parallel (FSDP~\cite{fsdp}) sharding strategy is achieved by constructing global arrays and annotating the sharding accordingly, with the XLA compiler~\cite{xla} taking care of the concrete implementation of the computation and communication between devices. 
We measured a model FLOPS utilization (MFU) of 55\%, resulting in 5189 tokens/second/device.  
Model parameters and optimizer state are kept in float32 to guarantee stable training, but we verified that inference works just as well with bfloat16 model parameters.

One training run of the final PaliGemma model using TPUv5e-256 takes slightly less than 3 days for Stage1 and 15h for each Stage2. Stage1 sees slightly less than 350\,B tokens, and both Stage2 combined about 90\,B tokens. Transfers take between 20min and 10h on TPUv3-32, depending on the task.

In order to avoid model brittleness to different image processing details in different frameworks, we randomize the image preprocessing details such as resize method, JPEG encoding, and apply very slight {\tt inception\_crop}.

\section{Results}
\label{sec:results}

\newcommand{\hrrow}{\rowcolor{lightgray!20}}
\newcommand{\zsrow}{$\llcorner$}
\newcommand{\variance}[1]{{\tiny \textcolor{gray}{±#1}}}

\begin{table}[!th]
    \centering \scriptsize
    \caption{Results (1 random run of 5) obtained with PaliGemma.
    Tasks marked with $\llcorner$ indicate zero-shot evaluation of the transferred model above. Where numbers depend on server submissions we report standard deviation from validation splits. Highlighted rows indicate resolution sensitive tasks. Per-task details and hyper-parameters are in Appendix~\ref{sec:app:tasks} and \ref{sec:app:hparams}.
    }
    \label{tbl:main_results}
    \vspace{1em}


\begin{tabular}{lccc}
\toprule
Task & 224px & 448px & 896px \\
\midrule
\multicolumn{4}{c}{\textbf{Image captioning}} \\
COCOcap & 141.9 \variance{0.3} & 144.6 \variance{0.5} & - \\
\zsrow NoCaps & 121.7 \variance{0.3} & 123.6 \variance{0.7} & - \\
COCO-35L (en) & 139.2 \variance{0.4} & 141.2 \variance{0.6} & - \\
COCO-35L (avg34) & 113.7 \variance{0.3} & 115.8 \variance{0.1} & - \\
\zsrow XM3600 (en) & \phantom{0}78.0 \variance{0.8} & \phantom{0}80.0 \variance{0.3} & - \\
\zsrow XM3600 (avg35) & \phantom{0}41.9 \variance{0.0} & \phantom{0}42.4 \variance{0.1} & - \\
Screen2Words & 117.6 \variance{0.7} & 119.6 \variance{0.7} & - \\
\hrrow TextCaps & 127.5 \variance{1.0} & 153.9 \variance{0.5} & - \\
\hrrow SciCap & 162.3 \variance{0.7} & 181.5 \variance{0.6} & - \\
\hrrow WidgetCap & 136.1 \variance{1.4} & 148.4 \variance{0.3} & - \\
\midrule
\multicolumn{4}{c}{\textbf{Visual question answering}} \\
VQAv2 & \phantom{0}83.2 \variance{0.4} & \phantom{0}85.6 \variance{0.2} & - \\
\zsrow MMVP & \phantom{0}47.3 \phantom{\variance{0.0}} & \phantom{0}45.3 \phantom{\variance{0.0}} & - \\
\zsrow POPE & \phantom{0}86.0 \phantom{\variance{0.0}} & \phantom{0}87.0 \phantom{\variance{0.0}} & - \\
\zsrow Objaverse Multiview & \phantom{0}62.7 \phantom{\variance{0.0}} & \phantom{0}62.8 \phantom{\variance{0.0}} & - \\
OKVQA & \phantom{0}63.5 \variance{0.3} & \phantom{0}63.2 \variance{0.2} & - \\
AOKVQA-MC & \phantom{0}76.4 \variance{0.4} & \phantom{0}76.9 \variance{0.3} & - \\
AOKVQA-DA & \phantom{0}61.9 \variance{0.6} & \phantom{0}63.2 \variance{0.5} & - \\
GQA & \phantom{0}65.6 \variance{0.4} & \phantom{0}67.0 \variance{0.3} & - \\
\zsrow xGQA (avg7) & \phantom{0}57.3 \variance{0.2} & \phantom{0}57.9 \variance{0.5} & - \\
NLVR2 & \phantom{0}90.0 \variance{0.2} & \phantom{0}88.9 \variance{0.3} & - \\
\zsrow MARVL (avg5) & \phantom{0}80.6 \variance{0.3} & \phantom{0}76.8 \variance{0.3} & - \\
AI2D & \phantom{0}72.1 \variance{0.7} & \phantom{0}73.3 \variance{0.7} & - \\
ScienceQA & \phantom{0}95.4 \variance{0.3} & \phantom{0}95.9 \variance{0.2} & - \\
RSVQA-lr & \phantom{0}92.6 \variance{0.3} & \phantom{0}93.1 \variance{0.7} & - \\
RSVQA-hr (test) & \phantom{0}92.6 \variance{0.1} & \phantom{0}92.8 \variance{0.1} & - \\
RSVQA-hr (test2) & \phantom{0}90.6 \variance{0.1} & \phantom{0}90.5 \variance{0.2} & - \\
\hrrow ChartQA (human) & \phantom{0}40.0 \variance{0.7} & \phantom{0}54.2 \variance{1.1} & - \\
\hrrow ChartQA (aug) & \phantom{0}74.2 \variance{0.2} & \phantom{0}88.5 \variance{0.3} & - \\
VizWizVQA & \phantom{0}73.7 \variance{0.2} & \phantom{0}75.5 \variance{0.5} & - \\
TallyQA (simple) & \phantom{0}81.7 \variance{0.2} & \phantom{0}84.9 \variance{0.1} & - \\
TallyQA (complex) & \phantom{0}69.6 \variance{0.1} & \phantom{0}72.3 \variance{0.2} & - \\
CountBenchQA & \phantom{0}81.9 \variance{1.6} & \phantom{0}83.1 \variance{2.1} & - \\
OCR-VQA & \phantom{0}72.3 \variance{0.4} & \phantom{0}74.6 \variance{0.4} & \phantom{0}74.9 \\
\hrrow TextVQA & \phantom{0}55.5 \variance{0.2} & \phantom{0}73.2 \variance{0.2} & \phantom{0}76.5 \\
\hrrow DocVQA & \phantom{0}43.7 \variance{0.5} & \phantom{0}78.0 \variance{0.3} & \phantom{0}84.8 \\
\hrrow InfoVQA & \phantom{0}28.5 \variance{0.4} & \phantom{0}40.5 \variance{0.3} & \phantom{0}47.8 \\
\hrrow ST-VQA & \phantom{0}63.3 \variance{0.5} & \phantom{0}81.8 \variance{0.3} & \phantom{0}84.4 \\
\midrule
\multicolumn{4}{c}{\textbf{Image segmentation}} \\
RefCOCO (testA) & \phantom{0}75.7 \variance{0.1} & \phantom{0}77.9 \variance{0.1} & \phantom{0}78.7 \\
RefCOCO (testB) & \phantom{0}70.7 \variance{0.2} & \phantom{0}72.4 \variance{0.2} & \phantom{0}73.9 \\
RefCOCO+ (testA) & \phantom{0}71.9 \variance{0.1} & \phantom{0}74.2 \variance{0.2} & \phantom{0}76.1 \\
RefCOCO+ (testB) & \phantom{0}64.5 \variance{0.6} & \phantom{0}64.5 \variance{0.2} & \phantom{0}66.9 \\
RefCOCOg (test) & \phantom{0}68.2 \variance{0.1} & \phantom{0}71.0 \variance{0.1} & \phantom{0}72.7 \\
\midrule
\multicolumn{4}{c}{\textbf{Video input}} \\
ActivityNet-QA & \phantom{0}50.8 \variance{0.4} & - \phantom{\variance{0.0}} & - \\
ActivityNet-CAP & \phantom{0}34.6 \variance{0.6} & - \phantom{\variance{0.0}} & - \\
MSRVTT-QA & \phantom{0}50.1 \variance{0.1} & - \phantom{\variance{0.0}} & - \\
MSRVTT-CAP & \phantom{0}70.5 \variance{0.9} & - \phantom{\variance{0.0}} & - \\
MSVD-QA & \phantom{0}60.2 \variance{0.3} & - \phantom{\variance{0.0}} & - \\
VATEX & \phantom{0}79.7 \variance{0.4} & - \phantom{\variance{0.0}} & - \\
\bottomrule
\end{tabular}
\vspace{-2em}  

\end{table}

In order to verify the transferability of PaliGemma to a wide variety of tasks, we transfer the pretrained models on more than 30 academic benchmarks via fine-tuning.
Importantly, none of these tasks or datasets are part of the pretraining data mixture, and their images are explicitly removed from the web-scale pretraining data.
Results are presented in Table~\ref{tbl:main_results}.

To select the hyper-parameters for transfer, we first sweep the parameters mentioned in Section~\ref{sec:transfer}, starting from the recommended value.
We do not necessarily perform the full cross-product, and we sometimes extend or supersample the range, if it seems promising.
Importantly, we make any such decisions and hyper-parameter choices based on the transfer task's validation split, and if none is provided, we hold out a small ``minival'' set from the training data.
Once we found good hyper-parameter values for a task, we re-train using the full training and validation data, and report final test numbers.
Details on tasks, metrics, data splits are in Appendix~\ref{sec:app:tasks} and final hyper-parameters in Appendix~\ref{sec:app:hparams}.
In Section~\ref{sec:abl:simple} we show that a single recommended value for each hyper-parameter without any exploration works almost as well on most tasks.

For all but video tasks, we report results on at least two resolutions to provide an impression of which tasks benefit from increased resolution.
We provide many resolution-related ablations in Section~\ref{sec:abl:res}.

Notably, we have not found any significant benefit from data augmentation.
We simply resize the input images to a square fixed resolution, even for tasks such as RefCOCO segmentation (more on that in Section~\ref{sec:abl:res} and Appendix~\ref{sec:app:refcoco}).

\section{Ablations}\label{sec:abl}

We conduct diverse ablations to gain deeper understanding of what matters for training and transferring VLMs.
Unless noted otherwise, all ablations are run with the same setup as the main models, except for making the Stage1 pretraining 10x shorter (\ie/ 100\,M examples seen), and transfer results are reported on validation sets instead of withheld test-sets.
For each experiment, we present only the salient result summary in the main text, but we provide a full per-task breakdown of results in the Appendix.

\subsection{Multimodal pretraining duration}\label{sec:abl:duration}

\begin{figure}[t]
  \centering
  \includegraphics[width=\columnwidth]{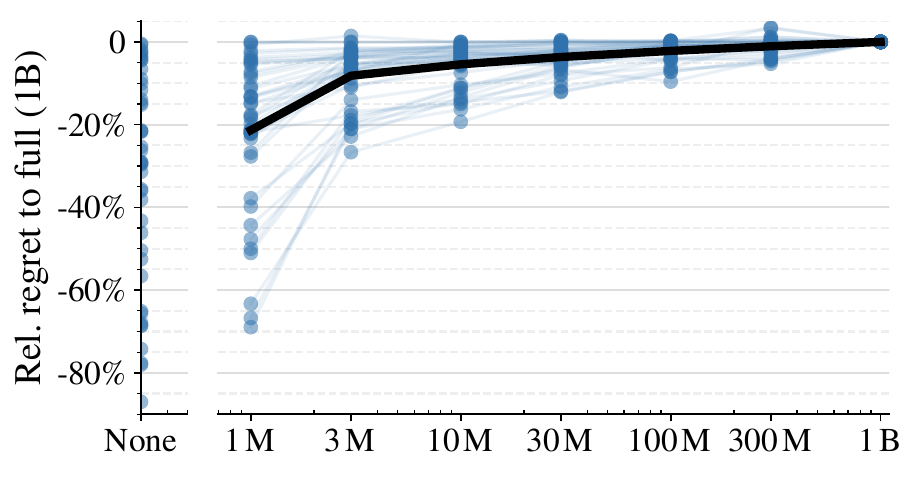}
  \caption{Relative regret of transfers when varying the amount of pretraining during Stage 1. Per task plot in appendix~\ref{sec:app:pt_duration}.}
  \label{fig:s1_duration}
\end{figure}

With its 1\,B examples seen, our multimodal pretraining (Stage1) is on the longer side, similar to BLIP-2~\cite{blip2}, InternVL~\cite{chen2023internvl}, QwenVL~\cite{qwen-vl}, Idefics2~\cite{idefics2} (all around 1\,B), but unlike ShareGPT4-v~\cite{chen2023sharegpt4v}, Mini-Gemini~\cite{li2024minigemini}, LLaVa~\cite{llava1} and its derivatives (around 1\,M).

To the best of our knowledge, the benefits from longer pretraining have not been studied in isolation.
We run pretrainings of various shorter durations, all the way down to completely skipping Stage1, and show the impact in Figure~\ref{fig:s1_duration}, with a complete break-down across tasks in the Appendix~\ref{sec:app:pt_duration}.
For the case of skipping Stage1, we use the best transfer result when sweeping over three learning-rates for each task.

The result shows that shorter training generally hurts, and skipping Stage1 entirely is the worst setting.
Some task are affected significantly, while others only deteriorate a little, highlighting the need for a broad and diverse set of evaluation tasks.
The $100\,$M pretraining duration appears to be a good trade-off for ablations: it is 10x shorter while not significantly hurting any task.

\subsection{Causal masking and learning objective}\label{sec:abl:objective}

\begin{figure}[t]
  \centering
  \includegraphics[width=\columnwidth]{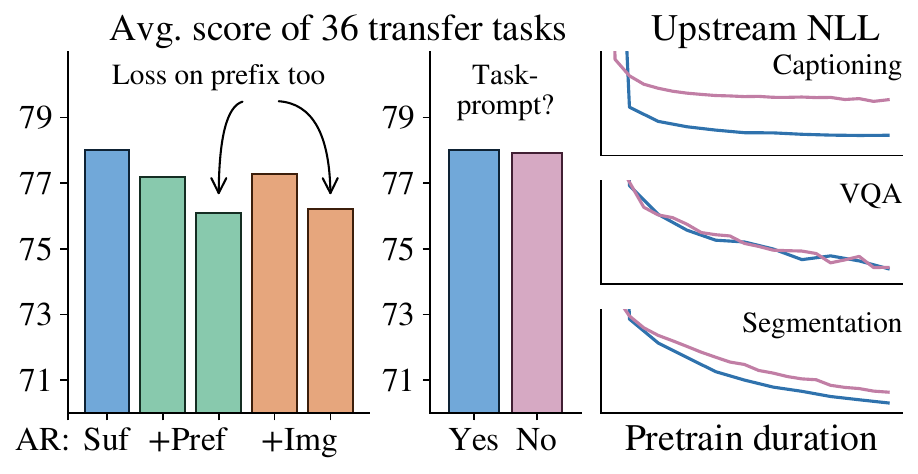}
  \caption{Learning setup for Stage1.
  Left: Where to apply the auto-regressive mask and loss. Middle and right: whether to include a task indicator.}
  \label{fig:s1_training}
\end{figure}

We ablate several of our key choices in the pretraining learning objective in Figure~\ref{fig:s1_training}, full per-task breakdown in Appendix~\ref{sec:app:learning}.

First, we investigate design choices for auto-regressive masking.
PaliGemma uses a prefix-LM strategy which allows full (bi-directional) attention on the ``input'' part of the data, \ie/ the {\tt image} and {\tt prefix} tokens, see also Figure~\ref{fig:attention_mask}.
The motivation is that it allows more tokens to actively participate in the ``thinking'' process from the start, as the image tokens now can attend to the prefix tokens which represent the query.
This is empirically confirmed in Figure~\ref{fig:s1_training} (left), where the green bars also include the auto-regressive masking on the {\tt prefix} tokens, and the orange bars further extend the auto-regressive mask to the {\tt image} tokens.
Both sets of bars work, but perform clearly worse than PaliGemma's prefix-LM setting represented by the blue bar.

Second, we apply the next-token-prediction loss on the {\tt suffix} (the output) only.
In principle, it can also be applied on the {\tt prefix} tokens, once they are auto-regressively masked.
This could provide more learning signal to the model by asking it to ``guess the question'' for example.
Again, Figure~\ref{fig:s1_training} shows that while it works, doing so clearly reduces average performance.

Finally, we eased the multi-task learning by using task-prefixes~\cite{beyer2023litdecoder}.
Figure~\ref{fig:s1_training} (middle) shows that after transfers, this choice of pretraining has no noticeable effect.
However, it would be incorrect to conclude that it has no effect on the model's training.
Indeed, pretraining validation perplexities on three representative tasks of pretraining are shown in Figure~\ref{fig:s1_training} (right): when the {\tt prefix} makes the task obvious, such as in VQA, a task-prefix has no effect.
For tasks where the {\tt prefix} does not perfectly disambiguate the exact task, however, the model (expectedly) does become noticeably more uncertain in its predictions without task-prefix.

Overall, the prefix-LM with task-prefix and supervision only on the {\tt suffix} tokens is an effective VLM pretraining objective.

\subsection{New token initialization}

\begin{figure}[t]
  \centering
  \includegraphics[width=\columnwidth]{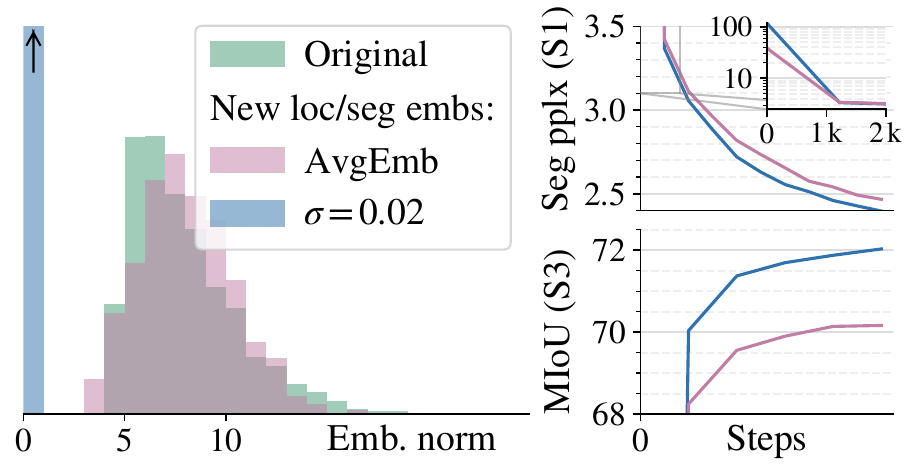}
  \caption{Initialization of new tokens matters. Left: distribution of embedding norms at init. Right: Top: initial loss is better with AvgEmb, but this changes after a few thousand steps. Bottom: the model pretrained with $\sigma=0.02$ inits transfers better to a task heavily using the new tokens.}
  \label{fig:newtok_init}
\end{figure}

We add new tokens to Gemma's vocabulary to support PaliGemma's ability to perform more structured computer vision tasks.
We add 1024 location tokens ({\tt <loc0000>} to {\tt <loc1023>}), which correspond to binned normalized image coordinates and are used in detection, referring expression, and grounded captioning tasks.
We also add 128 VQVAE~\cite{vqvae} tokenized single-object mask tokens~\cite{ning2023allin,pali3} ({\tt <seg000>} to {\tt <seg127>}) to support referring expression segmentation.

This poses the question of how to initialize the embeddings of these new tokens, given all other vocabulary tokens have already been trained as part of Gemma's pretraining.
One option is to use a standard small Gaussian noise initialization ($\sigma=0.02$, blue in Fig~\ref{fig:newtok_init}). 
However, \cite{hewitt2021initializing} argues for matching the average of the pretrained embeddings plus small noise (AvgEmb, mauve in Fig~\ref{fig:newtok_init}).
We compare these strategies in Figure~\ref{fig:newtok_init} and see that, while the AvgEmb strategy significantly improves initial perplexity of tasks using the new tokens (top-right zoom-in, step 0), this gain vanishes after a thousand steps of training.
The standard initialization strategy not only results in significantly better Stage1 perplexities at the end of pretraining (top-right, full plot), but also results in significantly better transfer of the model to tasks using these tokens, here RefCOCO segmentation MIoU (bottom right).

\subsection{To freeze or not to freeze?}\label{sec:abl:freeze}

The current common wisdom in VLMs~\cite{pali,palix,pali3,idefics,llava1,karamcheti2024prismaticvlm,blip2,lin2024moellava,kosmos} is to keep the image encoder and sometimes the LLM frozen during multimodal pretraining (our Stage1).
However, inspired by the positive results from CapPa~\cite{cappa} and LocCa~\cite{locca} which show that pretraining an image encoder using captioning objectives essentially solves contrastive's blind spot~\cite{sugarcrepe} to relation and localization, we pretrained PaliGemma with no frozen parts.
We now ablate the effect of freezing or tuning various parts of the model during Stage1 in Figure~\ref{fig:s1_freeze}, full per-task breakdown in Appendix~\ref{sec:app:freeze}.
Similar to concurrent works~\cite{cambrian1,mm1}, we find not freezing any part of the model is indeed advantageous.
First, after transfers, there is no difference to keeping the image encoder frozen (left, TT and TF).
Second, however, the validation perplexity (hence, predictability) of tasks requiring spatial understanding (right, green) is significantly improved.

Further, we show that all other options that include freezing the language model~\cite{tsimpoukelli21frozen} are significantly worse. 
Finally, resetting (and training, R) any part of the model hurts performance dramatically, confirming that Stage0 (\ie/ leveraging pre-trained components) is indeed crucial for attaining good results.

\begin{figure}[t]
  \centering
  \includegraphics[width=\columnwidth]{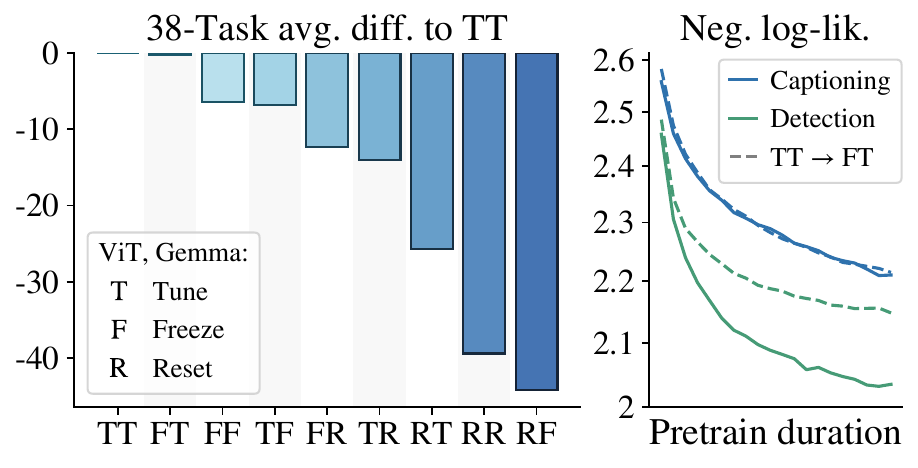}
  \caption{Training setup for Stage1.
  Left: The more is frozen or reset, the more performance deteriorates.
  Right: The effect of freezing ViT is most visible in some pretraining perplexities.}
  \label{fig:s1_freeze}
\end{figure}

\subsection{Connector design}\label{sec:abl:connector}

Throughout our experiments we use a linear connector to map SigLiP output embeddings to the inputs of Gemma. Given that an MLP connector~\cite{llava15} is a popular choice in the VLM literature, we also ablate this choice.

We consider two connector choices: a linear connector and an MLP (1 hidden layer, with GeLU non-linearity). We also consider two Stage1 pretraining settings: tune all weights (TT), or freeze everything but the connector (FF).

When tuning all weights, average transfer score is nearly identical for linear vs MLP, achieving $77.2$ and $77.1$ points respectively.
In the ``all-frozen'' scenario, linear vs MLP achieve $70.7$ vs $69.7$.
Surprisingly, we observe a small performance deterioration with the MLP connector.

Overall, we conclude that in our case, the linear connector seems preferable to the MLP connector.

\subsection{Image encoder: with or without?}\label{sec:abl:enc}

\begin{figure}[t]
  \centering
  \includegraphics[width=\columnwidth]{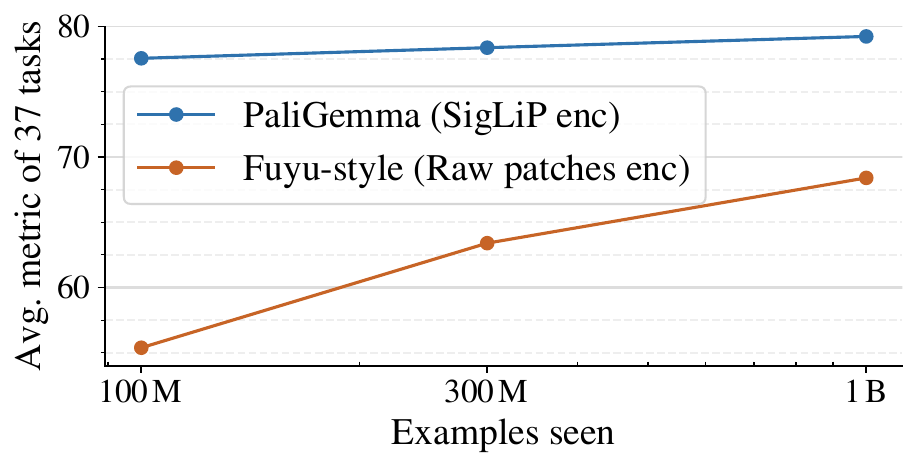}
  \caption{Not using a SigLiP image encoder at all and instead passing a linear projection of raw RGB patches to Gemma works, but is significantly less sample-efficient.}
  \label{fig:fuyu}
  \vspace{-1em}
\end{figure}

Most VLMs follow the setup of having an image encoder, such as CLIP/SigLIP (most works) or VQGAN (the Chameleon line of work~\cite{cm3,cm3_scaling,cm3leon,chameleon}), to turn the image into soft tokens before passing them to the LLM.
We are aware of only two works that attempt to simplify this overall setup by removing the image encoder entirely and passing raw image patches into a decoder-only LLM, namely Fuyu~\cite{fuyu} and EVE~\cite{eve}.
Unfortunately, the former provides no training details or ablations.
The latter, which is a concurrent work, provides some details about training and various ablations, but with mixed results.

Removing the SigLIP encoder in PaliGemma results in a model of the same unified decoder-only architecture.
We run our Stage1 and transfers in this setting.
Since the architecture significantly changed, we also re-tune the learning-rate of Stage1.
Figure~\ref{fig:fuyu} (per-task breakdown in Appendix~\ref{sec:app:enc}) shows that while this architecture still significantly lags behind, the scaling with pretraining duration seems potentially promising.
This is especially noteworthy considering that PaliGemma's SigLIP encoder has seen 40\,B image-text pairs during its Stage0 pre-training, while the Fuyu-style model sees images for the first time in the Stage1 pre-training shown here, and only sees up to 1\,B of them in our experiment.
This ablation confirms that such decoder-only VLMs might be a promising future direction towards simpler multimodal models, although they currently still suffer in training efficiency due to not being able to reuse vision components.

\subsection{Image resolution}\label{sec:abl:res}

Image resolution in VLMs is an important topic that has recently received increased attention~\cite{mm1,idefics2,cambrian1,ultrahdintern,wu2023vstar}.
PaliGemma uses a very simple approach to deal with resolution: Stage1 is pretrained at relatively low and economical 224px resolution, and short Stage2 then ``upcycles'' this checkpoint to higher resolutions (448px and 896px).
Hence, the final PaliGemma model comes with three different checkpoints for three different resolutions.

In this section we justify our approach and the necessity for providing three separate checkpoints, and compare it to a recently popular ``windowing'' approach. 
For the ablation studies, we consider resolutions 224px and 448px, and restrict evaluation to single-image tasks where resolution has significant impact on performance according to Table~\ref{tbl:main_results}, which we call ``Resolution-sensitive'' tasks.

\subsubsection{Resolution or sequence length?}\label{sec:abl:res_or_seqlen}

\begin{figure}[t]
  \centering
  \includegraphics[width=\columnwidth]{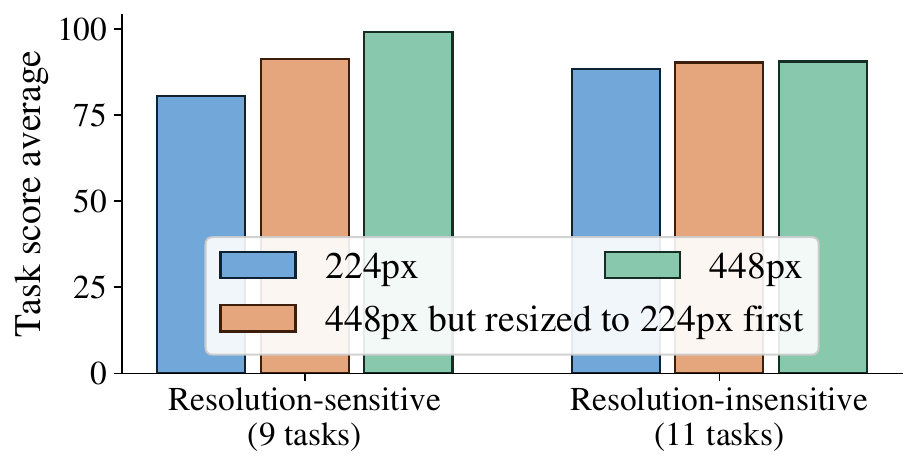}
  \caption{Increasing resolution has two effects: increased information content of the input image, and increased model capacity via sequence length. For tasks that benefit from increased resolution, both of these effects contribute roughly equally to the overall gain.}
  \label{fig:res_via_224}
\end{figure}

We generally see either neutral or improved performance across tasks when increasing the resolution of the input image (see Table~\ref{tbl:main_results}).
However, it is unclear whether this improvement comes from the fact that the image has higher resolution and therefore more information, or whether it is thanks to the resulting longer sequence length and thus increased model FLOPs and capacity.
We disentangle these by running Stage2 and transfers at 448\,px resolution, \emph{but downscaling each image to 224\,px before resizing it to 448\,px}.
Thus, the model gets to see the information content of a low-res (224\,px) image but with the model capacity of the high-res (448\,px) setting.

The result in Fig~\ref{fig:res_via_224} shows that, for those tasks for which resolution has an effect at all, the reason for the improved performance is split roughly equally between these two causes.
This is true for every individual task and not just an effect of averaging, see Appendix~\ref{sec:app:res_or_seqlen}.

\subsubsection{Need resolution-specific checkpoints?}\label{sec:abl:res_ckpt}

\begin{figure}[t]
  \centering
  \includegraphics[width=\columnwidth]{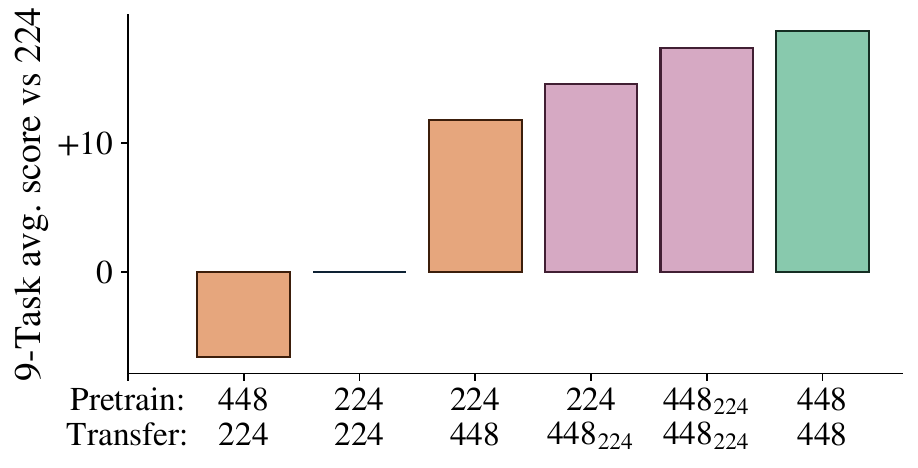}
  \caption{Different ways of increasing resolution.
  While resolution during transfers works (orange), it is not the best setup.
  Using ``windows'' works better (mauve, $s_w$ means image resolution $s$ with window-size $w$),
  but simply running Stage2 at increased resolution with no tricks works best, justifying the need to provide multiple checkpoints. Per-task breakdown in Appendix~\ref{sec:app:res_window}.}
  \label{fig:res_variants}
\end{figure}

PaliGemma provides one checkpoint per resolution.
But is this really needed? Could we not provide just a single, maybe high-resolution checkpoint, and adapt it as needed during transfers?

Figure~\ref{fig:res_variants} shows clearly that providing either only a 224\,px or only a 448\,px checkpoint would not be sufficient:
Transferring the 448\,px checkpoint at 224\,px resolution (first bar, orange) leads to significantly worse results than using the 224\,px checkpoint (second bar, zero), even though the latter had slightly less pretraining.
Similarly, transferring the 224\,px checkpoint at resolution 448\,px (third bar, orange), while improving the results significantly, still lags far behind transferring the checkpoint whose native resolution is 448\,px (last bar, green).

Thus, in the absence of flexible-resolution modeling tricks such as FlexiViT~\cite{beyer2023flexivit} or NaViT~\cite{dehghani2024navit}, we recommend running extended pretraining for increasing resolution (Stage2) and providing separate checkpoints for all supported resolutions.

\subsubsection{To resize or to window?}\label{sec:abl:windows}

Another recently common way of increasing input resolution is by ``windowing'' the models~\cite{visheratin2024curse,wu2023vstar,zhang2024ferretv2}, \ie/ applying the same model on windows of the model's native resolution from the higher-resolution image.
We evaluate the simplest variant of this alternative: the 448\,px resolution image is cut into four pieces, each one passed through the SigLIP image encoder separately.
All four sets of 256 image embedding tokens are then concatenated and passed to Gemma.
This is also related to using windowed attention in a ViT taking the full image~\cite{vitdet}.
We experimented with adding extra ``window ID'' position embeddings to indicate which window tokens come from, but this did not significantly change any of the results.

The mauve bars in Figure~\ref{fig:res_variants} correspond to windowing settings.
While overall the performance is worse than that of a native 448px model, the windowing approach can be a promising way of transferring a model to a higher resolution \emph{when no higher-resolution checkpoint is made available}, and when running a Stage2-like continued pretraining is not feasible.

Windowing might still seem preferable for speed reasons.
However, we only observed at most a 5\% speedup in training across various setups from windowing.
This is explained by Gemma being significantly larger than ViT-So400m, and the Gemma part of the model is unaffected by windowing.

\subsubsection{Stage2 mixture re-weighting}\label{sec:abl:s2mix}

Finally, the pretraining mixture changes between Stage1 and Stage2.
While previous PaLI versions change the mixture makeup, for PaliGemma we always use the full set of tasks, only changing their weighting, sampling ``resolution-related'' tasks (OCR, detection, segmentation) more frequently at higher resolutions.
As an ablation, we run a Stage2 training with the same mixture ratios as Stage1.
After transfers, this checkpoint is significantly worse on only three tasks (DocVQA, ChartQA, XM3600), but otherwise within per-task variance (Full results in Appendix~\ref{sec:app:s2_reweight}).

Thus, while changing the mixing ratio helped a little, it seems that when the intent is to train a base model for fine-tuning, the precise mixture ratios might not be as important as when training a model intended for sampling zero-shot, where it would significantly affect sampling frequencies.

\section{Transferability}\label{sec:abl:transferability}

To show that PaliGemma is suitable for transfer under different scenarios, we run experiments that quantify its repeatability, sensitivity to hyper-parameters, and number of transfer examples.

\subsection{Repeatability (variance)}\label{sec:abl:repeats}
\begin{figure}
  \centering
  \includegraphics[width=\columnwidth]{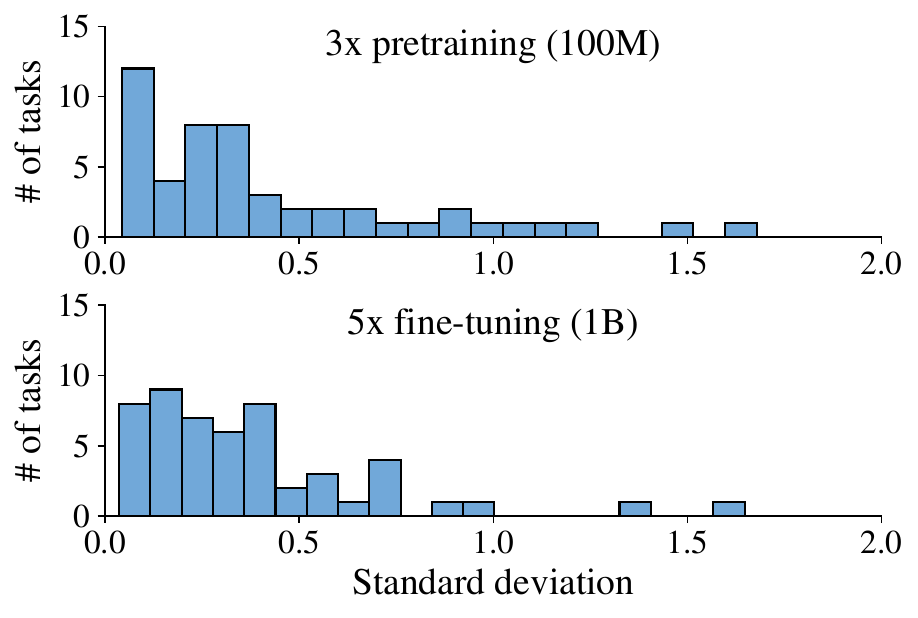}
  \caption{For most tasks, the standard deviation of final metric values between re-runs with different seeds is below $0.5$. This is true both for Stage1 (top, single transfers of three 100M runs), as well as for transfers (bottom, values from Table~\ref{tbl:main_results}).}
  \label{fig:variance_100m}
\end{figure}

In the main results (Table~\ref{tbl:main_results}), we show the standard deviation across 5 transfer reruns using the best hyper-parameter (listed in \ref{sec:app:hparams}).
It is generally very small across most tasks, meaning transfer from the pretrained checkpoint is highly repeatable.
In Figure~\ref{fig:variance_100m} we further show that the standard deviation of transferring from three reruns of Stage1 falls within the same range, meaning pretraining itself is also highly repeatable.

\subsection{Transfer hyper-parameter sensitivity}\label{sec:abl:simple}
\newcommand{\taskval}[2]{#1}
\newcommand{\taskvaldetail}[2]{#1: {\footnotesize #2}}
\begin{table}[t]
\centering
\caption{Relative regret of using the suggested hyper-parameter setting for all tasks instead of performing hyper-parameter search. Per task plot in Appendix~ \ref{sec:app:transfer_simple}.}
\label{tbl:simple_hparams}
\vspace{-2mm}

\begin{tabular}{p{2.4cm}p{0.3cm}L{4cm}}
\toprule
\textbf{Rel. Regret} & \textbf{\#} & \textbf{Tasks} \\

\midrule $[\text{None},2.5\%)$ & 37 & \taskval{All other tasks}{} \\
\midrule $[2.5\%,5.0\%)$ & 2 & \taskvaldetail{ChartQA~(human)}{3.2\%}\newline \taskvaldetail{RefCOCO~(val)}{4.8\%} \\
\midrule $[5.0\%,10.0\%)$ & 2 & \taskvaldetail{RefCOCOg~(val)}{5.8\%}\newline \taskvaldetail{ScienceQA}{6.7\%} \\
\midrule $[10.0\%,100\%]$ & 2 & \taskvaldetail{RefCOCO+~(val)}{10.7\%}\newline \taskvaldetail{SciCap}{60.5\%} \\

\bottomrule
\end{tabular}

\end{table}

However one question a reader may have is whether the choice of transfer hyper-parameter is important.
To ablate that we run all tasks at 224px and under a single and simple hyper-parameter setup, which was also highlighted in bold in Section~\ref{sec:transfer}: lr=1e-5, bs=256, no dropout, no label smoothing, no weight decay, and not freezing anything.
The only task dependent parameter is the number of epochs for which we use each task's best one but cap it at 10.

The results (Table~\ref{tbl:simple_hparams}) show that this simplified and single setup works very well for the majority of the tasks. The main exceptions we found were for tasks like RefCOCO and SciCap tasks which seem to benefit significantly from increasing the number of epochs while enabling label smoothing and dropout.

\subsection{Transfer with limited examples}\label{sec:abl:transfer_lowdata}

\begin{figure}[t]
  \centering
  \includegraphics[width=\columnwidth]{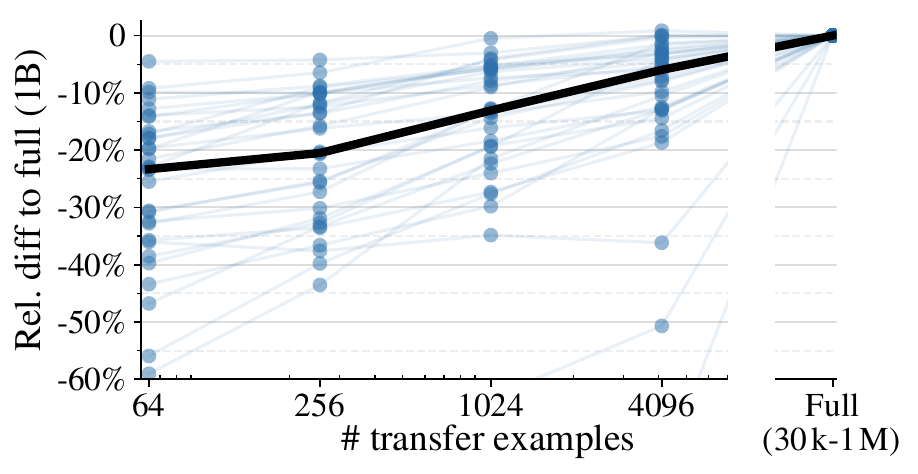}
  \caption{Relative regret of using a limited number of transfer examples. Thick line represents the median. Per task plot in appendix~\ref{sec:app:transfer_lowdata}.}
  \label{fig:lowex_transfer}
\end{figure}

To analyze how many examples are needed to make PaliGemma solve a new task, we finetune PaliGemma with limited number of examples (64, 256, 1024, 4096).
We sweep transfer with varying learning rates, epochs and batch size and report the best number without separate minival, to indicate the potential.

We run every setting with 5 different seeds, which also affect which examples are used.
We found this important, as finetuning with limited examples exhibits high variance for some tasks (\eg/ RefCOCO mIOU varied within 10\%-30\%).
As a note, this variance also occurs when repeating with the same examples, but different batch order.
Importantly, seed selection is not overfitting to the metric as the selected model performs equally well in the validation and test splits.
But it does allows us to draw conclusions without needing to solve the open problem of making few-example fine-tuning stable.

Overall, when comparing the best runs of each hyper-parameter and seed with the results obtained with the full dataset, Figure~\ref{fig:lowex_transfer} shows that it is not necessary to have a transfer dataset in the order of 10\,k examples.
The majority of the tasks can reach within 10\% of the full-data score when using 4\,k examples and 20\% when using only 256 examples. In many cases the score with 64 transfer examples are good enough to prototype using PaliGemma for a new application.

\section{Noteworthy tidbits}\label{sec:tasks}

We also briefly discuss several small but interesting discoveries or we made, providing more details on each in the Appendix.

{\bf Plain resize to square for segmentation.} We found simple resize to square ($224\times224$) to work just as well as popular aspect-ratio preserving zoom and crop augmentations. (Appendix~\ref{sec:app:refcoco})

{\bf Counting: introducing CountBenchQA.}
Due to skewed number distribution and varying image quality~\cite{kaji2024evaluating}, we found TallyQA lacking in its ability to assess current VLM's ability to count.
This is why we introduce CountBenchQA, a VLM-ready version of the CountBench~\cite{paiss2023countclip} dataset. (Appendix~\ref{sec:app:countbenchqa}).

{\bf Issues in published WidgetCaps numbers.} We found issues in at least three previous work's evaluation of WidgetCaps, rendering numerical comparisons invalid. (Appendix~\ref{sec:app:widgetcaps_eval}).

{\bf Image annotations work as well as prompts.} Marking the widget to be captioned with a red box in the image gives the same results as indicating it with {\tt <loc>} tokens in the prompt. (Appendix~\ref{sec:app:widgetcaps_box})

{\bf RoPE interpolation unnecessary for upscaling.} For Stage2, we tried interpolating the RoPE position indices for the image tokens to preserve their semantics from Stage1. However, we saw no benefit from doing so.

{\bf Zero-shot generalization to 3D renders.} Although never explicitly trained for it, PaliGemma generalizes surprisingly well to 3D renders form Objaverse without fine-tuning. (Appendix~\ref{sec:app:objaverse})

{\bf Our MMVP result is SOTA by a large margin.} PaliGemma at 224px achieves 47.3\% paired accuracy, while GPT4-V and Gemini achieve 38.7\% and 40.7\%, respectively, and all other models including LLaVa perform below chance.

\section{Conclusion}

PaliGemma is a new, small, open base VLM that shines when transferred to a broad range of tasks.
Our results show that VLMs on the ``smaller'' side can provide state-of-the-art performance across a wide variety of benchmarks.
We also hope that providing the base model without instruction tuning serves as a useful starting point for further research in instruction tuning, specific applications, and encourages clearer separation of base models and fine-tunes in VLM research.

\bibliographystyle{abbrvnat}
\nobibliography*
\bibliography{document}

\begin{thebibliography}{139}
\providecommand{\natexlab}[1]{#1}
\providecommand{\url}[1]{\texttt{#1}}
\expandafter\ifx\csname urlstyle\endcsname\relax
  \providecommand{\doi}[1]{doi: #1}\else
  \providecommand{\doi}{doi: \begingroup \urlstyle{rm}\Url}\fi

\bibitem[Acharya et~al.(2019)Acharya, Kafle, and Kanan]{acharya2019tallyqa}
M.~Acharya, K.~Kafle, and C.~Kanan.
\newblock Tallyqa: Answering complex counting questions.
\newblock In \emph{AAAI}, 2019.

\bibitem[Aghajanyan et~al.(2022)Aghajanyan, Huang, Ross, Karpukhin, Xu, Goyal,
  Okhonko, Joshi, Ghosh, Lewis, and Zettlemoyer]{cm3}
A.~Aghajanyan, B.~Huang, C.~Ross, V.~Karpukhin, H.~Xu, N.~Goyal, D.~Okhonko,
  M.~Joshi, G.~Ghosh, M.~Lewis, and L.~Zettlemoyer.
\newblock {CM3:} {A} causal masked multimodal model of the internet.
\newblock \emph{CoRR}, abs/2201.07520, 2022.
\newblock URL \url{https://arxiv.org/abs/2201.07520}.

\bibitem[Aghajanyan et~al.(2023)Aghajanyan, Yu, Conneau, Hsu, Hambardzumyan,
  Zhang, Roller, Goyal, Levy, and Zettlemoyer]{cm3_scaling}
A.~Aghajanyan, L.~Yu, A.~Conneau, W.~Hsu, K.~Hambardzumyan, S.~Zhang,
  S.~Roller, N.~Goyal, O.~Levy, and L.~Zettlemoyer.
\newblock Scaling laws for generative mixed-modal language models.
\newblock In A.~Krause, E.~Brunskill, K.~Cho, B.~Engelhardt, S.~Sabato, and
  J.~Scarlett, editors, \emph{International Conference on Machine Learning,
  {ICML} 2023, 23-29 July 2023, Honolulu, Hawaii, {USA}}, volume 202 of
  \emph{Proceedings of Machine Learning Research}, pages 265--279. {PMLR},
  2023.
\newblock URL \url{https://proceedings.mlr.press/v202/aghajanyan23a.html}.

\bibitem[Agrawal et~al.(2019)Agrawal, Desai, Wang, Chen, Jain, Johnson, Batra,
  Parikh, Lee, and Anderson]{agrawal2019nocaps}
H.~Agrawal, K.~Desai, Y.~Wang, X.~Chen, R.~Jain, M.~Johnson, D.~Batra,
  D.~Parikh, S.~Lee, and P.~Anderson.
\newblock nocaps: novel object captioning at scale.
\newblock In \emph{Proceedings of the IEEE International Conference on Computer
  Vision}, pages 8948--8957, 2019.

\bibitem[Alabdulmohsin et~al.(2023)Alabdulmohsin, Zhai, Kolesnikov, and
  Beyer]{sovit}
I.~Alabdulmohsin, X.~Zhai, A.~Kolesnikov, and L.~Beyer.
\newblock Getting vit in shape: Scaling laws for compute-optimal model design.
\newblock In \emph{NeurIPS}, 2023.

\bibitem[Alayrac et~al.(2022)Alayrac, Donahue, Luc, Miech, Barr, Hasson, Lenc,
  Mensch, Millican, Reynolds, Ring, Rutherford, Cabi, Han, Gong, Samangooei,
  Monteiro, Menick, Borgeaud, Brock, Nematzadeh, Sharifzadeh, Binkowski,
  Barreira, Vinyals, Zisserman, and Simonyan]{alayrac2022flamingo}
J.-B. Alayrac, J.~Donahue, P.~Luc, A.~Miech, I.~Barr, Y.~Hasson, K.~Lenc,
  A.~Mensch, K.~Millican, M.~Reynolds, R.~Ring, E.~Rutherford, S.~Cabi, T.~Han,
  Z.~Gong, S.~Samangooei, M.~Monteiro, J.~Menick, S.~Borgeaud, A.~Brock,
  A.~Nematzadeh, S.~Sharifzadeh, M.~Binkowski, R.~Barreira, O.~Vinyals,
  A.~Zisserman, and K.~Simonyan.
\newblock Flamingo: a visual language model for few-shot learning, 2022.
\newblock URL \url{https://arxiv.org/abs/2204.14198}.

\bibitem[Anil et~al.(2023)Anil, Borgeaud, Wu, Alayrac, Yu, Soricut, Schalkwyk,
  Dai, Hauth, Millican, Silver, Petrov, Johnson, Antonoglou, Schrittwieser,
  Glaese, Chen, Pitler, Lillicrap, Lazaridou, Firat, Molloy, Isard, Barham,
  Hennigan, Lee, Viola, Reynolds, Xu, Doherty, Collins, Meyer, Rutherford,
  Moreira, Ayoub, Goel, Tucker, Piqueras, Krikun, Barr, Savinov, Danihelka,
  Roelofs, White, Andreassen, von Glehn, Yagati, Kazemi, Gonzalez, Khalman,
  Sygnowski, and et~al.]{gemini}
R.~Anil, S.~Borgeaud, Y.~Wu, J.~Alayrac, J.~Yu, R.~Soricut, J.~Schalkwyk, A.~M.
  Dai, A.~Hauth, K.~Millican, D.~Silver, S.~Petrov, M.~Johnson, I.~Antonoglou,
  J.~Schrittwieser, A.~Glaese, J.~Chen, E.~Pitler, T.~P. Lillicrap,
  A.~Lazaridou, O.~Firat, J.~Molloy, M.~Isard, P.~R. Barham, T.~Hennigan,
  B.~Lee, F.~Viola, M.~Reynolds, Y.~Xu, R.~Doherty, E.~Collins, C.~Meyer,
  E.~Rutherford, E.~Moreira, K.~Ayoub, M.~Goel, G.~Tucker, E.~Piqueras,
  M.~Krikun, I.~Barr, N.~Savinov, I.~Danihelka, B.~Roelofs, A.~White,
  A.~Andreassen, T.~von Glehn, L.~Yagati, M.~Kazemi, L.~Gonzalez, M.~Khalman,
  J.~Sygnowski, and et~al.
\newblock Gemini: {A} family of highly capable multimodal models.
\newblock \emph{CoRR}, abs/2312.11805, 2023.
\newblock \doi{10.48550/ARXIV.2312.11805}.
\newblock URL \url{https://doi.org/10.48550/arXiv.2312.11805}.

\bibitem[Aniruddha~Kembhavi(2016)]{kembhavi2016eccv-AI2D}
E.~K. M. S. H. H. A.~F. Aniruddha~Kembhavi, Mike~Salvato.
\newblock A diagram is worth a dozen images.
\newblock In \emph{European Conference on Computer Vision (ECCV)}, 2016.
\newblock URL \url{https://api.semanticscholar.org/CorpusID:2682274}.

\bibitem[Baechler et~al.(2024)Baechler, Sunkara, Wang, Zubach, Mansoor, Etter,
  Cărbune, Lin, Chen, and Sharma]{screenai}
G.~Baechler, S.~Sunkara, M.~Wang, F.~Zubach, H.~Mansoor, V.~Etter, V.~Cărbune,
  J.~Lin, J.~Chen, and A.~Sharma.
\newblock Screenai: A vision-language model for ui and infographics
  understanding, 2024.
\newblock URL \url{https://arxiv.org/abs/2402.04615}.

\bibitem[Bai et~al.(2023)Bai, Bai, Yang, Wang, Tan, Wang, Lin, Zhou, and
  Zhou]{qwen-vl}
J.~Bai, S.~Bai, S.~Yang, S.~Wang, S.~Tan, P.~Wang, J.~Lin, C.~Zhou, and
  J.~Zhou.
\newblock Qwen-vl: {A} versatile vision-language model for understanding,
  localization, text reading, and beyond.
\newblock \emph{CoRR}, abs/2308.12966, 2023.
\newblock \doi{10.48550/ARXIV.2308.12966}.
\newblock URL \url{https://doi.org/10.48550/arXiv.2308.12966}.

\bibitem[Bavishi et~al.(2023)Bavishi, Elsen, Hawthorne, Nye, Odena, Somani, and
  Ta\c{s}\i{}rlar]{fuyu}
R.~Bavishi, E.~Elsen, C.~Hawthorne, M.~Nye, A.~Odena, A.~Somani, and
  S.~Ta\c{s}\i{}rlar.
\newblock Introducing our multimodal models, 2023.
\newblock URL \url{https://www.adept.ai/blog/fuyu-8b}.

\bibitem[Beyer et~al.(2022)Beyer, Zhai, and Kolesnikov]{big_vision}
L.~Beyer, X.~Zhai, and A.~Kolesnikov.
\newblock Big vision.
\newblock \url{https://github.com/google-research/big_vision}, 2022.

\bibitem[Beyer et~al.(2023{\natexlab{a}})Beyer, Izmailov, Kolesnikov, Caron,
  Kornblith, Zhai, Minderer, Tschannen, Alabdulmohsin, and
  Pavetic]{beyer2023flexivit}
L.~Beyer, P.~Izmailov, A.~Kolesnikov, M.~Caron, S.~Kornblith, X.~Zhai,
  M.~Minderer, M.~Tschannen, I.~Alabdulmohsin, and F.~Pavetic.
\newblock Flexivit: One model for all patch sizes.
\newblock In \emph{Proceedings of the IEEE/CVF Conference on Computer Vision
  and Pattern Recognition}, pages 14496--14506, 2023{\natexlab{a}}.

\bibitem[Beyer et~al.(2023{\natexlab{b}})Beyer, Wan, Madan, Pavetic, Steiner,
  Kolesnikov, Pinto, Bugliarello, Wang, Yu, et~al.]{beyer2023litdecoder}
L.~Beyer, B.~Wan, G.~Madan, F.~Pavetic, A.~Steiner, A.~Kolesnikov, A.~S. Pinto,
  E.~Bugliarello, X.~Wang, Q.~Yu, et~al.
\newblock A study of autoregressive decoders for multi-tasking in computer
  vision.
\newblock \emph{arXiv preprint arXiv:2303.17376}, 2023{\natexlab{b}}.

\bibitem[Biten et~al.(2019)Biten, Tito, Mafla, Gomez, Rusinol, Jawahar,
  Valveny, and Karatzas]{Biten_2019-STVQA}
A.~F. Biten, R.~Tito, A.~Mafla, L.~Gomez, M.~Rusinol, C.~Jawahar, E.~Valveny,
  and D.~Karatzas.
\newblock Scene text visual question answering.
\newblock In \emph{2019 IEEE/CVF International Conference on Computer Vision
  (ICCV)}. IEEE, Oct. 2019.
\newblock \doi{10.1109/iccv.2019.00439}.
\newblock URL \url{http://dx.doi.org/10.1109/ICCV.2019.00439}.

\bibitem[Bradbury et~al.(2018)Bradbury, Frostig, Hawkins, Johnson, Leary,
  Maclaurin, Necula, Paszke, Vander{P}las, Wanderman-{M}ilne, and
  Zhang]{jax2018github}
J.~Bradbury, R.~Frostig, P.~Hawkins, M.~J. Johnson, C.~Leary, D.~Maclaurin,
  G.~Necula, A.~Paszke, J.~Vander{P}las, S.~Wanderman-{M}ilne, and Q.~Zhang.
\newblock {JAX}: composable transformations of {P}ython+{N}um{P}y programs,
  2018.
\newblock URL \url{http://github.com/google/jax}.

\bibitem[Bugliarello et~al.(2022)Bugliarello, Liu, Pfeiffer, Reddy, Elliott,
  Ponti, and Vuli{'c}]{bugliarello-etal-2022-iglue}
E.~Bugliarello, F.~Liu, J.~Pfeiffer, S.~Reddy, D.~Elliott, E.~M. Ponti, and
  I.~Vuli{'c}.
\newblock {IGLUE}: {A} benchmark for transfer learning across modalities,
  tasks, and languages.
\newblock In \emph{Proceedings of the 39th International Conference on Machine
  Learning}, volume 162 of \emph{Proceedings of Machine Learning Research},
  page 2370–2392, Balitmore, MA, July 2022. PMLR.
\newblock URL \url{https://proceedings.mlr.press/v162/bugliarello22a.html}.

\bibitem[Cha et~al.(2023)Cha, Kang, Mun, and Roh]{Honeybee}
J.~Cha, W.~Kang, J.~Mun, and B.~Roh.
\newblock Honeybee: Locality-enhanced projector for multimodal {LLM}.
\newblock \emph{CoRR}, abs/2312.06742, 2023.
\newblock \doi{10.48550/ARXIV.2312.06742}.
\newblock URL \url{https://doi.org/10.48550/arXiv.2312.06742}.

\bibitem[Changpinyo et~al.(2022)Changpinyo, Kukliansy, Szpektor, Chen, Ding,
  and Soricut]{Changpinyo}
S.~Changpinyo, D.~Kukliansy, I.~Szpektor, X.~Chen, N.~Ding, and R.~Soricut.
\newblock All you may need for {VQA} are image captions.
\newblock In M.~Carpuat, M.-C. de~Marneffe, and I.~V. Meza~Ruiz, editors,
  \emph{Proceedings of the 2022 Conference of the North American Chapter of the
  Association for Computational Linguistics: Human Language Technologies},
  pages 1947--1963, Seattle, United States, July 2022. Association for
  Computational Linguistics.
\newblock \doi{10.18653/v1/2022.naacl-main.142}.
\newblock URL \url{https://aclanthology.org/2022.naacl-main.142}.

\bibitem[Chen and Dolan(2011)]{msvd}
D.~L. Chen and W.~B. Dolan.
\newblock Collecting highly parallel data for paraphrase evaluation.
\newblock In D.~Lin, Y.~Matsumoto, and R.~Mihalcea, editors, \emph{The 49th
  Annual Meeting of the Association for Computational Linguistics: Human
  Language Technologies, Proceedings of the Conference, 19-24 June, 2011,
  Portland, Oregon, {USA}}, pages 190--200. The Association for Computer
  Linguistics, 2011.
\newblock URL \url{https://aclanthology.org/P11-1020/}.

\bibitem[Chen et~al.(2023{\natexlab{a}})Chen, Li, Dong, Zhang, He, Wang, Zhao,
  and Lin]{chen2023sharegpt4v}
L.~Chen, J.~Li, X.~Dong, P.~Zhang, C.~He, J.~Wang, F.~Zhao, and D.~Lin.
\newblock Sharegpt4v: Improving large multi-modal models with better captions.
\newblock \emph{arXiv preprint arXiv:2311.12793}, 2023{\natexlab{a}}.

\bibitem[Chen et~al.(2022{\natexlab{a}})Chen, Saxena, Li, Fleet, and
  Hinton]{pix2seq}
T.~Chen, S.~Saxena, L.~Li, D.~J. Fleet, and G.~E. Hinton.
\newblock Pix2seq: {A} language modeling framework for object detection.
\newblock In \emph{ICLR}, 2022{\natexlab{a}}.

\bibitem[Chen et~al.(2022{\natexlab{b}})Chen, Wang, Changpinyo, Piergiovanni,
  Padlewski, Salz, Goodman, Grycner, Mustafa, Beyer, Kolesnikov, Puigcerver,
  Ding, Rong, Akbari, Mishra, Xue, Thapliyal, Bradbury, Kuo, Seyedhosseini,
  Jia, Ayan, Riquelme, Steiner, Angelova, Zhai, Houlsby, and Soricut]{pali}
X.~Chen, X.~Wang, S.~Changpinyo, A.~J. Piergiovanni, P.~Padlewski, D.~Salz,
  S.~Goodman, A.~Grycner, B.~Mustafa, L.~Beyer, A.~Kolesnikov, J.~Puigcerver,
  N.~Ding, K.~Rong, H.~Akbari, G.~Mishra, L.~Xue, A.~Thapliyal, J.~Bradbury,
  W.~Kuo, M.~Seyedhosseini, C.~Jia, B.~K. Ayan, C.~Riquelme, A.~Steiner,
  A.~Angelova, X.~Zhai, N.~Houlsby, and R.~Soricut.
\newblock {PaLI}: {A} jointly-scaled multilingual language-image model.
\newblock \emph{CoRR}, arXiv:2209.06794, 2022{\natexlab{b}}.

\bibitem[Chen et~al.(2023{\natexlab{b}})Chen, Djolonga, Padlewski, Mustafa,
  Changpinyo, Wu, Ruiz, Goodman, Wang, Tay, Shakeri, Dehghani, Salz, Lucic,
  Tschannen, Nagrani, Hu, Joshi, Pang, Montgomery, Pietrzyk, Ritter,
  Piergiovanni, Minderer, Pavetic, Waters, Li, Alabdulmohsin, Beyer, Amelot,
  Lee, Steiner, Li, Keysers, Arnab, Xu, Rong, Kolesnikov, Seyedhosseini,
  Angelova, Zhai, Houlsby, and Soricut]{palix}
X.~Chen, J.~Djolonga, P.~Padlewski, B.~Mustafa, S.~Changpinyo, J.~Wu, C.~R.
  Ruiz, S.~Goodman, X.~Wang, Y.~Tay, S.~Shakeri, M.~Dehghani, D.~Salz,
  M.~Lucic, M.~Tschannen, A.~Nagrani, H.~Hu, M.~Joshi, B.~Pang, C.~Montgomery,
  P.~Pietrzyk, M.~Ritter, A.~J. Piergiovanni, M.~Minderer, F.~Pavetic,
  A.~Waters, G.~Li, I.~Alabdulmohsin, L.~Beyer, J.~Amelot, K.~Lee, A.~P.
  Steiner, Y.~Li, D.~Keysers, A.~Arnab, Y.~Xu, K.~Rong, A.~Kolesnikov,
  M.~Seyedhosseini, A.~Angelova, X.~Zhai, N.~Houlsby, and R.~Soricut.
\newblock {PaLI-X}: On scaling up a multilingual vision and language model.
\newblock \emph{CoRR}, arXiv:2305.18565, 2023{\natexlab{b}}.

\bibitem[Chen et~al.(2023{\natexlab{c}})Chen, Wang, Beyer, Kolesnikov, Wu,
  Voigtlaender, Mustafa, Goodman, Alabdulmohsin, Padlewski, Salz, Xiong,
  Vlasic, Pavetic, Rong, Yu, Keysers, Zhai, and Soricut]{pali3}
X.~Chen, X.~Wang, L.~Beyer, A.~Kolesnikov, J.~Wu, P.~Voigtlaender, B.~Mustafa,
  S.~Goodman, I.~Alabdulmohsin, P.~Padlewski, D.~Salz, X.~Xiong, D.~Vlasic,
  F.~Pavetic, K.~Rong, T.~Yu, D.~Keysers, X.~Zhai, and R.~Soricut.
\newblock {PaLI-3} vision language models: Smaller, faster, stronger.
\newblock \emph{CoRR}, arXiv:2310.09199, 2023{\natexlab{c}}.

\bibitem[Chen et~al.(2023{\natexlab{d}})Chen, Wu, Wang, Su, Chen, Xing, Zhong,
  Zhang, Zhu, Lu, Li, Luo, Lu, Qiao, and Dai]{chen2023internvl}
Z.~Chen, J.~Wu, W.~Wang, W.~Su, G.~Chen, S.~Xing, M.~Zhong, Q.~Zhang, X.~Zhu,
  L.~Lu, B.~Li, P.~Luo, T.~Lu, Y.~Qiao, and J.~Dai.
\newblock Internvl: Scaling up vision foundation models and aligning for
  generic visual-linguistic tasks.
\newblock \emph{arXiv preprint arXiv:2312.14238}, 2023{\natexlab{d}}.

\bibitem[Cho et~al.(2021)Cho, Lei, Tan, and Bansal]{cho21unifying}
J.~Cho, J.~Lei, H.~Tan, and M.~Bansal.
\newblock Unifying vision-and-language tasks via text generation.
\newblock In M.~Meila and T.~Zhang, editors, \emph{Proceedings of the 38th
  International Conference on Machine Learning}, volume 139 of
  \emph{Proceedings of Machine Learning Research}, pages 1931--1942. PMLR,
  18--24 Jul 2021.
\newblock URL \url{https://proceedings.mlr.press/v139/cho21a.html}.

\bibitem[Chowdhery et~al.(2023)Chowdhery, Narang, Devlin, Bosma, Mishra,
  Roberts, Barham, Chung, Sutton, Gehrmann, Schuh, Shi, Tsvyashchenko, Maynez,
  Rao, Barnes, Tay, Shazeer, Prabhakaran, Reif, Du, Hutchinson, Pope, Bradbury,
  Austin, Isard, Gur{-}Ari, Yin, Duke, Levskaya, Ghemawat, Dev, Michalewski,
  Garcia, Misra, Robinson, Fedus, Zhou, Ippolito, Luan, Lim, Zoph, Spiridonov,
  Sepassi, Dohan, Agrawal, Omernick, Dai, Pillai, Pellat, Lewkowycz, Moreira,
  Child, Polozov, Lee, Zhou, Wang, Saeta, Diaz, Firat, Catasta, Wei,
  Meier{-}Hellstern, Eck, Dean, Petrov, and Fiedel]{palm}
A.~Chowdhery, S.~Narang, J.~Devlin, M.~Bosma, G.~Mishra, A.~Roberts, P.~Barham,
  H.~W. Chung, C.~Sutton, S.~Gehrmann, P.~Schuh, K.~Shi, S.~Tsvyashchenko,
  J.~Maynez, A.~Rao, P.~Barnes, Y.~Tay, N.~Shazeer, V.~Prabhakaran, E.~Reif,
  N.~Du, B.~Hutchinson, R.~Pope, J.~Bradbury, J.~Austin, M.~Isard,
  G.~Gur{-}Ari, P.~Yin, T.~Duke, A.~Levskaya, S.~Ghemawat, S.~Dev,
  H.~Michalewski, X.~Garcia, V.~Misra, K.~Robinson, L.~Fedus, D.~Zhou,
  D.~Ippolito, D.~Luan, H.~Lim, B.~Zoph, A.~Spiridonov, R.~Sepassi, D.~Dohan,
  S.~Agrawal, M.~Omernick, A.~M. Dai, T.~S. Pillai, M.~Pellat, A.~Lewkowycz,
  E.~Moreira, R.~Child, O.~Polozov, K.~Lee, Z.~Zhou, X.~Wang, B.~Saeta,
  M.~Diaz, O.~Firat, M.~Catasta, J.~Wei, K.~Meier{-}Hellstern, D.~Eck, J.~Dean,
  S.~Petrov, and N.~Fiedel.
\newblock Palm: Scaling language modeling with pathways.
\newblock \emph{J. Mach. Learn. Res.}, 24:\penalty0 240:1--240:113, 2023.
\newblock URL \url{http://jmlr.org/papers/v24/22-1144.html}.

\bibitem[Dehghani et~al.(2023)Dehghani, Djolonga, Mustafa, Padlewski, Heek,
  Gilmer, Steiner, Caron, Geirhos, Alabdulmohsin, Jenatton, Beyer, Tschannen,
  Arnab, Wang, Ruiz, Minderer, Puigcerver, Evci, Kumar, van Steenkiste,
  Elsayed, Mahendran, Yu, Oliver, Huot, Bastings, Collier, Gritsenko, Birodkar,
  Vasconcelos, Tay, Mensink, Kolesnikov, Pavetic, Tran, Kipf, Lucic, Zhai,
  Keysers, Harmsen, and Houlsby]{vit22b}
M.~Dehghani, J.~Djolonga, B.~Mustafa, P.~Padlewski, J.~Heek, J.~Gilmer, A.~P.
  Steiner, M.~Caron, R.~Geirhos, I.~Alabdulmohsin, R.~Jenatton, L.~Beyer,
  M.~Tschannen, A.~Arnab, X.~Wang, C.~R. Ruiz, M.~Minderer, J.~Puigcerver,
  U.~Evci, M.~Kumar, S.~van Steenkiste, G.~F. Elsayed, A.~Mahendran, F.~Yu,
  A.~Oliver, F.~Huot, J.~Bastings, M.~Collier, A.~A. Gritsenko, V.~Birodkar,
  C.~N. Vasconcelos, Y.~Tay, T.~Mensink, A.~Kolesnikov, F.~Pavetic, D.~Tran,
  T.~Kipf, M.~Lucic, X.~Zhai, D.~Keysers, J.~J. Harmsen, and N.~Houlsby.
\newblock Scaling vision transformers to 22 billion parameters.
\newblock In A.~Krause, E.~Brunskill, K.~Cho, B.~Engelhardt, S.~Sabato, and
  J.~Scarlett, editors, \emph{International Conference on Machine Learning,
  {ICML} 2023, 23-29 July 2023, Honolulu, Hawaii, {USA}}, volume 202 of
  \emph{Proceedings of Machine Learning Research}, pages 7480--7512. {PMLR},
  2023.
\newblock URL \url{https://proceedings.mlr.press/v202/dehghani23a.html}.

\bibitem[Dehghani et~al.(2024)Dehghani, Mustafa, Djolonga, Heek, Minderer,
  Caron, Steiner, Puigcerver, Geirhos, Alabdulmohsin,
  et~al.]{dehghani2024navit}
M.~Dehghani, B.~Mustafa, J.~Djolonga, J.~Heek, M.~Minderer, M.~Caron,
  A.~Steiner, J.~Puigcerver, R.~Geirhos, I.~M. Alabdulmohsin, et~al.
\newblock Patch n’pack: Navit, a vision transformer for any aspect ratio and
  resolution.
\newblock \emph{Advances in Neural Information Processing Systems}, 36, 2024.

\bibitem[Deitke et~al.(2023)Deitke, Schwenk, Salvador, Weihs, Michel,
  VanderBilt, Schmidt, Ehsani, Kembhavi, and Farhadi]{deitke2023objaverse}
M.~Deitke, D.~Schwenk, J.~Salvador, L.~Weihs, O.~Michel, E.~VanderBilt,
  L.~Schmidt, K.~Ehsani, A.~Kembhavi, and A.~Farhadi.
\newblock Objaverse: {A} universe of annotated 3d objects.
\newblock In \emph{{IEEE/CVF} Conference on Computer Vision and Pattern
  Recognition, {CVPR} 2023, Vancouver, BC, Canada, June 17-24, 2023}, pages
  13142--13153. {IEEE}, 2023.
\newblock \doi{10.1109/CVPR52729.2023.01263}.
\newblock URL \url{https://doi.org/10.1109/CVPR52729.2023.01263}.

\bibitem[Desai and Johnson(2021)]{desai2021virtex}
K.~Desai and J.~Johnson.
\newblock Virtex: Learning visual representations from textual annotations.
\newblock In \emph{Proceedings of the IEEE/CVF conference on computer vision
  and pattern recognition}, pages 11162--11173, 2021.

\bibitem[Devlin et~al.(2019)Devlin, Chang, Lee, and Toutanova]{bert}
J.~Devlin, M.~Chang, K.~Lee, and K.~Toutanova.
\newblock {BERT:} pre-training of deep bidirectional transformers for language
  understanding.
\newblock In J.~Burstein, C.~Doran, and T.~Solorio, editors, \emph{Proceedings
  of the 2019 Conference of the North American Chapter of the Association for
  Computational Linguistics: Human Language Technologies, {NAACL-HLT} 2019,
  Minneapolis, MN, USA, June 2-7, 2019, Volume 1 (Long and Short Papers)},
  pages 4171--4186. Association for Computational Linguistics, 2019.
\newblock \doi{10.18653/V1/N19-1423}.
\newblock URL \url{https://doi.org/10.18653/v1/n19-1423}.

\bibitem[Diao et~al.(2024)Diao, Cui, Li, Wang, Lu, and Wang]{eve}
H.~Diao, Y.~Cui, X.~Li, Y.~Wang, H.~Lu, and X.~Wang.
\newblock Unveiling encoder-free vision-language models, 2024.
\newblock URL \url{https://arxiv.org/abs/2406.11832}.

\bibitem[Dong et~al.(2024)Dong, Zhang, Zang, Cao, Wang, Ouyang, Zhang, Duan,
  Zhang, Li, Yan, Gao, Chen, Zhang, Li, Li, Wang, Chen, He, Zhang, Dai, Qiao,
  Lin, and Wang]{ultrahdintern}
X.~Dong, P.~Zhang, Y.~Zang, Y.~Cao, B.~Wang, L.~Ouyang, S.~Zhang, H.~Duan,
  W.~Zhang, Y.~Li, H.~Yan, Y.~Gao, Z.~Chen, X.~Zhang, W.~Li, J.~Li, W.~Wang,
  K.~Chen, C.~He, X.~Zhang, J.~Dai, Y.~Qiao, D.~Lin, and J.~Wang.
\newblock Internlm-xcomposer2-4khd: A pioneering large vision-language model
  handling resolutions from 336 pixels to 4k hd, 2024.
\newblock URL \url{https://arxiv.org/abs/2404.06512}.

\bibitem[Driess et~al.(2023)Driess, Xia, Sajjadi, Lynch, Chowdhery, Ichter,
  Wahid, Tompson, Vuong, Yu, Huang, Chebotar, Sermanet, Duckworth, Levine,
  Vanhoucke, Hausman, Toussaint, Greff, Zeng, Mordatch, and Florence]{palm_e}
D.~Driess, F.~Xia, M.~S.~M. Sajjadi, C.~Lynch, A.~Chowdhery, B.~Ichter,
  A.~Wahid, J.~Tompson, Q.~Vuong, T.~Yu, W.~Huang, Y.~Chebotar, P.~Sermanet,
  D.~Duckworth, S.~Levine, V.~Vanhoucke, K.~Hausman, M.~Toussaint, K.~Greff,
  A.~Zeng, I.~Mordatch, and P.~Florence.
\newblock {PaLM-E}: An embodied multimodal language model.
\newblock In A.~Krause, E.~Brunskill, K.~Cho, B.~Engelhardt, S.~Sabato, and
  J.~Scarlett, editors, \emph{International Conference on Machine Learning,
  {ICML} 2023, 23-29 July 2023, Honolulu, Hawaii, {USA}}, volume 202 of
  \emph{Proceedings of Machine Learning Research}, pages 8469--8488. {PMLR},
  2023.
\newblock URL \url{https://proceedings.mlr.press/v202/driess23a.html}.

\bibitem[Geigle et~al.(2024)Geigle, Timofte, and Glavaš]{geigle2024african}
G.~Geigle, R.~Timofte, and G.~Glavaš.
\newblock African or european swallow? benchmarking large vision-language
  models for fine-grained object classification, 2024.
\newblock URL \url{https://arxiv.org/abs/2406.14496}.

\bibitem[{Google Cloud}(20xx)]{tpucloud}
{Google Cloud}.
\newblock Introduction to {C}loud {TPU}.
\newblock \url{https://cloud.google.com/tpu/docs/intro-to-tpu}, 20xx.
\newblock Accessed: 2024-07-04.

\bibitem[Goyal et~al.(2017)Goyal, Khot, Summers{-}Stay, Batra, and
  Parikh]{balanced_vqa_v2}
Y.~Goyal, T.~Khot, D.~Summers{-}Stay, D.~Batra, and D.~Parikh.
\newblock Making the {V} in {VQA} matter: Elevating the role of image
  understanding in {V}isual {Q}uestion {A}nswering.
\newblock In \emph{Computer Vision and Pattern Recognition (CVPR)}, 2017.

\bibitem[Gurari et~al.(2018)Gurari, Li, Stangl, Guo, Lin, Grauman, Luo, and
  Bigham]{gurari2018vizwiz}
D.~Gurari, Q.~Li, A.~J. Stangl, A.~Guo, C.~Lin, K.~Grauman, J.~Luo, and J.~P.
  Bigham.
\newblock Vizwiz grand challenge: Answering visual questions from blind people.
\newblock In \emph{Proceedings of the IEEE conference on computer vision and
  pattern recognition}, pages 3608--3617, 2018.

\bibitem[He et~al.(2024)He, Liu, Wu, Yuan, Wang, Huang, and Zhao]{bunny}
M.~He, Y.~Liu, B.~Wu, J.~Yuan, Y.~Wang, T.~Huang, and B.~Zhao.
\newblock Efficient multimodal learning from data-centric perspective.
\newblock \emph{CoRR}, abs/2402.11530, 2024.
\newblock \doi{10.48550/ARXIV.2402.11530}.
\newblock URL \url{https://doi.org/10.48550/arXiv.2402.11530}.

\bibitem[Hewitt(2021)]{hewitt2021initializing}
J.~Hewitt.
\newblock Initializing new word embeddings for pretrained language models.
\newblock \url{https:/nlp.stanford.edu/~johnhew//vocab-expansion.html}, 2021.

\bibitem[Hsieh et~al.(2024)Hsieh, Zhang, Ma, Kembhavi, and Krishna]{sugarcrepe}
C.-Y. Hsieh, J.~Zhang, Z.~Ma, A.~Kembhavi, and R.~Krishna.
\newblock Sugarcrepe: fixing hackable benchmarks for vision-language
  compositionality.
\newblock In \emph{Proceedings of the 37th International Conference on Neural
  Information Processing Systems}, NIPS '23, Red Hook, NY, USA, 2024. Curran
  Associates Inc.

\bibitem[Hsu et~al.(2021)Hsu, Giles, and Huang]{hsu2021scicap}
T.-Y. Hsu, C.~L. Giles, and T.-H. Huang.
\newblock Scicap: Generating captions for scientific figures.
\newblock \emph{arXiv preprint arXiv:2110.11624}, 2021.

\bibitem[Huang et~al.(2023)Huang, Dong, Wang, Hao, Singhal, Ma, Lv, Cui,
  Mohammed, Patra, Liu, Aggarwal, Chi, Bjorck, Chaudhary, Som, Song, and
  Wei]{kosmos}
S.~Huang, L.~Dong, W.~Wang, Y.~Hao, S.~Singhal, S.~Ma, T.~Lv, L.~Cui, O.~K.
  Mohammed, B.~Patra, Q.~Liu, K.~Aggarwal, Z.~Chi, N.~J.~B. Bjorck,
  V.~Chaudhary, S.~Som, X.~Song, and F.~Wei.
\newblock Language is not all you need: Aligning perception with language
  models.
\newblock In A.~Oh, T.~Naumann, A.~Globerson, K.~Saenko, M.~Hardt, and
  S.~Levine, editors, \emph{Advances in Neural Information Processing Systems
  36: Annual Conference on Neural Information Processing Systems 2023, NeurIPS
  2023, New Orleans, LA, USA, December 10 - 16, 2023}, 2023.

\bibitem[Huang et~al.(2024)Huang, Meng, Liu, Su, Collier, and
  Lu]{huang2024sparkles}
Y.~Huang, Z.~Meng, F.~Liu, Y.~Su, N.~Collier, and Y.~Lu.
\newblock Sparkles: Unlocking chats across multiple images for multimodal
  instruction-following models, 2024.
\newblock URL \url{https://openreview.net/forum?id=oq5EF8parZ}.

\bibitem[Hudson and Manning(2019)]{DBLP:journals/corr/abs-2306-14610-GQA}
D.~Hudson and C.~Manning.
\newblock Gqa: A new dataset for real-world visual reasoning and compositional
  question answering.
\newblock \emph{Computer Vision and Pattern Recognition (CVPR)},
  abs/1902.09506, 2019.
\newblock \doi{10.48550/arXiv.1902.09506}.
\newblock URL \url{https://doi.org/10.48550/arXiv.1902.09506}.

\bibitem[Jaegle et~al.(2021)Jaegle, Gimeno, Brock, Vinyals, Zisserman, and
  Carreira]{perceiver}
A.~Jaegle, F.~Gimeno, A.~Brock, O.~Vinyals, A.~Zisserman, and J.~Carreira.
\newblock Perceiver: General perception with iterative attention.
\newblock In M.~Meila and T.~Zhang, editors, \emph{Proceedings of the 38th
  International Conference on Machine Learning, {ICML} 2021, 18-24 July 2021,
  Virtual Event}, volume 139 of \emph{Proceedings of Machine Learning
  Research}, pages 4651--4664. {PMLR}, 2021.
\newblock URL \url{http://proceedings.mlr.press/v139/jaegle21a.html}.

\bibitem[Jia et~al.(2021)Jia, Yang, Xia, Chen, Parekh, Pham, Le, Sung, Li, and
  Duerig]{align}
C.~Jia, Y.~Yang, Y.~Xia, Y.~Chen, Z.~Parekh, H.~Pham, Q.~V. Le, Y.~Sung, Z.~Li,
  and T.~Duerig.
\newblock Scaling up visual and vision-language representation learning with
  noisy text supervision.
\newblock In M.~Meila and T.~Zhang, editors, \emph{Proceedings of the 38th
  International Conference on Machine Learning, {ICML} 2021, 18-24 July 2021,
  Virtual Event}, volume 139 of \emph{Proceedings of Machine Learning
  Research}, pages 4904--4916. {PMLR}, 2021.
\newblock URL \url{http://proceedings.mlr.press/v139/jia21b.html}.

\bibitem[Kabra et~al.(2024)Kabra, Matthey, Lerchner, and
  Mitra]{kabra2024objaverse}
R.~Kabra, L.~Matthey, A.~Lerchner, and N.~J. Mitra.
\newblock Leveraging vlm-based pipelines to annotate 3d objects.
\newblock In \emph{Proceedings of the 41st International Conference on Machine
  Learning}, volume 235 of \emph{Proceedings of Machine Learning Research}.
  PMLR, 2024.

\bibitem[Kajić et~al.(2024)Kajić, Wiles, Albuquerque, Bauer, Wang,
  Pont-Tuset, and Nematzadeh]{kaji2024evaluating}
I.~Kajić, O.~Wiles, I.~Albuquerque, M.~Bauer, S.~Wang, J.~Pont-Tuset, and
  A.~Nematzadeh.
\newblock Evaluating numerical reasoning in text-to-image models, 2024.

\bibitem[Karamcheti et~al.(2024)Karamcheti, Nair, Balakrishna, Liang, Kollar,
  and Sadigh]{karamcheti2024prismaticvlm}
S.~Karamcheti, S.~Nair, A.~Balakrishna, P.~Liang, T.~Kollar, and D.~Sadigh.
\newblock Prismatic vlms: Investigating the design space of
  visually-conditioned language models, 2024.
\newblock URL \url{https://arxiv.org/abs/2402.07865}.

\bibitem[Kazemzadeh et~al.(2014)Kazemzadeh, Ordonez, Matten, and
  Berg]{kazemzadeh2014referit}
S.~Kazemzadeh, V.~Ordonez, M.~Matten, and T.~Berg.
\newblock {R}efer{I}t{G}ame: Referring to objects in photographs of natural
  scenes.
\newblock In A.~Moschitti, B.~Pang, and W.~Daelemans, editors,
  \emph{Proceedings of the 2014 Conference on Empirical Methods in Natural
  Language Processing ({EMNLP})}, pages 787--798, Doha, Qatar, Oct. 2014.
  Association for Computational Linguistics.
\newblock \doi{10.3115/v1/D14-1086}.
\newblock URL \url{https://aclanthology.org/D14-1086}.

\bibitem[Kirillov et~al.(2023)Kirillov, Mintun, Ravi, Mao, Rolland, Gustafson,
  Xiao, Whitehead, Berg, Lo, Dollar, and Girshick]{sam}
A.~Kirillov, E.~Mintun, N.~Ravi, H.~Mao, C.~Rolland, L.~Gustafson, T.~Xiao,
  S.~Whitehead, A.~C. Berg, W.-Y. Lo, P.~Dollar, and R.~Girshick.
\newblock Segment anything.
\newblock In \emph{Proceedings of the IEEE/CVF International Conference on
  Computer Vision (ICCV)}, pages 4015--4026, October 2023.

\bibitem[Kolesnikov et~al.(2020)Kolesnikov, Beyer, Zhai, Puigcerver, Yung,
  Gelly, and Houlsby]{bit}
A.~Kolesnikov, L.~Beyer, X.~Zhai, J.~Puigcerver, J.~Yung, S.~Gelly, and
  N.~Houlsby.
\newblock Big transfer ({BiT}): General visual representation learning.
\newblock In \emph{European Conference on Computer Vision (ECCV)}, 2020.

\bibitem[Kolesnikov et~al.(2022)Kolesnikov, Pinto, Beyer, Zhai, Harmsen, and
  Houlsby]{uvim}
A.~Kolesnikov, A.~S. Pinto, L.~Beyer, X.~Zhai, J.~Harmsen, and N.~Houlsby.
\newblock Uvim: {A} unified modeling approach for vision with learned guiding
  codes.
\newblock In S.~Koyejo, S.~Mohamed, A.~Agarwal, D.~Belgrave, K.~Cho, and A.~Oh,
  editors, \emph{Advances in Neural Information Processing Systems 35: Annual
  Conference on Neural Information Processing Systems 2022, NeurIPS 2022, New
  Orleans, LA, USA, November 28 - December 9, 2022}, 2022.

\bibitem[Krishna et~al.(2017)Krishna, Hata, Ren, Fei-Fei, and
  Carlos~Niebles]{krishna2017activitynetcap}
R.~Krishna, K.~Hata, F.~Ren, L.~Fei-Fei, and J.~Carlos~Niebles.
\newblock Dense-captioning events in videos.
\newblock In \emph{Proceedings of the IEEE international conference on computer
  vision}, pages 706--715, 2017.

\bibitem[Kudo and Richardson(2018)]{kudo-richardson-2018-sentencepiece}
T.~Kudo and J.~Richardson.
\newblock {S}entence{P}iece: A simple and language independent subword
  tokenizer and detokenizer for neural text processing.
\newblock In E.~Blanco and W.~Lu, editors, \emph{Proceedings of the 2018
  Conference on Empirical Methods in Natural Language Processing: System
  Demonstrations}, pages 66--71, Brussels, Belgium, Nov. 2018. Association for
  Computational Linguistics.
\newblock \doi{10.18653/v1/D18-2012}.
\newblock URL \url{https://aclanthology.org/D18-2012}.

\bibitem[Lauren{\c{c}}on et~al.(2024)Lauren{\c{c}}on, Tronchon, Cord, and
  Sanh]{idefics2}
H.~Lauren{\c{c}}on, L.~Tronchon, M.~Cord, and V.~Sanh.
\newblock What matters when building vision-language models?
\newblock \emph{CoRR}, abs/2405.02246, 2024.
\newblock \doi{10.48550/ARXIV.2405.02246}.
\newblock URL \url{https://doi.org/10.48550/arXiv.2405.02246}.

\bibitem[Laurençon et~al.(2023)Laurençon, Saulnier, Tronchon, Bekman, Singh,
  Lozhkov, Wang, Karamcheti, Rush, Kiela, Cord, and Sanh]{idefics}
H.~Laurençon, L.~Saulnier, L.~Tronchon, S.~Bekman, A.~Singh, A.~Lozhkov,
  T.~Wang, S.~Karamcheti, A.~M. Rush, D.~Kiela, M.~Cord, and V.~Sanh.
\newblock Obelics: An open web-scale filtered dataset of interleaved image-text
  documents, 2023.
\newblock URL \url{https://arxiv.org/abs/2306.16527}.

\bibitem[Li et~al.(2022{\natexlab{a}})Li, Li, Xiong, and Hoi]{blip}
J.~Li, D.~Li, C.~Xiong, and S.~C.~H. Hoi.
\newblock {BLIP:} bootstrapping language-image pre-training for unified
  vision-language understanding and generation.
\newblock In K.~Chaudhuri, S.~Jegelka, L.~Song, C.~Szepesv{\'{a}}ri, G.~Niu,
  and S.~Sabato, editors, \emph{International Conference on Machine Learning,
  {ICML} 2022, 17-23 July 2022, Baltimore, Maryland, {USA}}, volume 162 of
  \emph{Proceedings of Machine Learning Research}, pages 12888--12900. {PMLR},
  2022{\natexlab{a}}.
\newblock URL \url{https://proceedings.mlr.press/v162/li22n.html}.

\bibitem[Li et~al.(2023)Li, Li, Savarese, and Hoi]{blip2}
J.~Li, D.~Li, S.~Savarese, and S.~C.~H. Hoi.
\newblock {BLIP-2:} bootstrapping language-image pre-training with frozen image
  encoders and large language models.
\newblock In A.~Krause, E.~Brunskill, K.~Cho, B.~Engelhardt, S.~Sabato, and
  J.~Scarlett, editors, \emph{International Conference on Machine Learning,
  {ICML} 2023, 23-29 July 2023, Honolulu, Hawaii, {USA}}, volume 202 of
  \emph{Proceedings of Machine Learning Research}, pages 19730--19742. {PMLR},
  2023.
\newblock URL \url{https://proceedings.mlr.press/v202/li23q.html}.

\bibitem[Li et~al.(2020)Li, Li, He, Zheng, Li, and Guan]{Li2020WidgetCap}
Y.~Li, G.~Li, L.~He, J.~Zheng, H.~Li, and Z.~Guan.
\newblock Widget captioning: Generating natural language description for
  mobileuser interface elements.
\newblock In \emph{Conference on Empirical Methods in Natural Language
  Processing}, 2020.

\bibitem[Li et~al.(2022{\natexlab{b}})Li, Mao, Girshick, and He]{vitdet}
Y.~Li, H.~Mao, R.~Girshick, and K.~He.
\newblock Exploring plain vision transformer backbones for object detection,
  2022{\natexlab{b}}.
\newblock URL \url{https://arxiv.org/abs/2203.16527}.

\bibitem[Li et~al.(2024)Li, Zhang, Wang, Zhong, Chen, Chu, Liu, and
  Jia]{li2024minigemini}
Y.~Li, Y.~Zhang, C.~Wang, Z.~Zhong, Y.~Chen, R.~Chu, S.~Liu, and J.~Jia.
\newblock Mini-gemini: Mining the potential of multi-modality vision language
  models.
\newblock \emph{arXiv preprint arXiv:2403.18814}, 2024.

\bibitem[Lin et~al.(2024)Lin, Tang, Ye, Cui, Zhu, Jin, Huang, Zhang, Pang,
  Ning, and Yuan]{lin2024moellava}
B.~Lin, Z.~Tang, Y.~Ye, J.~Cui, B.~Zhu, P.~Jin, J.~Huang, J.~Zhang, Y.~Pang,
  M.~Ning, and L.~Yuan.
\newblock Moe-llava: Mixture of experts for large vision-language models, 2024.
\newblock URL \url{https://arxiv.org/abs/2401.15947}.

\bibitem[Lin et~al.(2014)Lin, Maire, Belongie, Bourdev, Girshick, Hays, Perona,
  Ramanan, Doll{'{a} }r, and Zitnick]{coco2014}
T.~Lin, M.~Maire, S.~J. Belongie, L.~D. Bourdev, R.~B. Girshick, J.~Hays,
  P.~Perona, D.~Ramanan, P.~Doll{'{a} }r, and C.~L. Zitnick.
\newblock Microsoft {COCO:} common objects in context.
\newblock \emph{CoRR}, abs/1405.0312, 2014.
\newblock URL \url{http://arxiv.org/abs/1405.0312}.

\bibitem[Liu et~al.(2021)Liu, Bugliarello, Ponti, Reddy, Collier, and
  Elliott]{liu-etal-2021-visually}
F.~Liu, E.~Bugliarello, E.~M. Ponti, S.~Reddy, N.~Collier, and D.~Elliott.
\newblock Visually grounded reasoning across languages and cultures.
\newblock In M.-F. Moens, X.~Huang, L.~Specia, and S.~W.-t. Yih, editors,
  \emph{Proceedings of the 2021 Conference on Empirical Methods in Natural
  Language Processing}, pages 10467--10485, Online and Punta Cana, Dominican
  Republic, Nov. 2021. Association for Computational Linguistics.
\newblock \doi{10.18653/v1/2021.emnlp-main.818}.
\newblock URL \url{https://aclanthology.org/2021.emnlp-main.818}.

\bibitem[Liu et~al.(2023{\natexlab{a}})Liu, Li, Li, and Lee]{llava15}
H.~Liu, C.~Li, Y.~Li, and Y.~J. Lee.
\newblock Improved baselines with visual instruction tuning,
  2023{\natexlab{a}}.

\bibitem[Liu et~al.(2023{\natexlab{b}})Liu, Li, Wu, and Lee]{llava1}
H.~Liu, C.~Li, Q.~Wu, and Y.~J. Lee.
\newblock Visual instruction tuning.
\newblock In \emph{NeurIPS}, 2023{\natexlab{b}}.

\bibitem[Lobry et~al.(2020)Lobry, Marcos, Murray, and Tuia]{Lobry_2020-RSVQA}
S.~Lobry, D.~Marcos, J.~Murray, and D.~Tuia.
\newblock Rsvqa: Visual question answering for remote sensing data.
\newblock \emph{IEEE Transactions on Geoscience and Remote Sensing},
  58\penalty0 (12):\penalty0 8555--8566, Dec. 2020.
\newblock ISSN 1558-0644.
\newblock \doi{10.1109/tgrs.2020.2988782}.
\newblock URL \url{http://dx.doi.org/10.1109/TGRS.2020.2988782}.

\bibitem[Lu et~al.(2023{\natexlab{a}})Lu, Clark, Lee, Zhang, Khosla, Marten,
  Hoiem, and Kembhavi]{unifiedio2}
J.~Lu, C.~Clark, S.~Lee, Z.~Zhang, S.~Khosla, R.~Marten, D.~Hoiem, and
  A.~Kembhavi.
\newblock Unified-io 2: Scaling autoregressive multimodal models with vision,
  language, audio, and action.
\newblock \emph{CoRR}, abs/2312.17172, 2023{\natexlab{a}}.
\newblock \doi{10.48550/ARXIV.2312.17172}.
\newblock URL \url{https://doi.org/10.48550/arXiv.2312.17172}.

\bibitem[Lu et~al.(2023{\natexlab{b}})Lu, Clark, Zellers, Mottaghi, and
  Kembhavi]{unifiedio}
J.~Lu, C.~Clark, R.~Zellers, R.~Mottaghi, and A.~Kembhavi.
\newblock {UNIFIED-IO:} {A} unified model for vision, language, and multi-modal
  tasks.
\newblock In \emph{ICLR}, 2023{\natexlab{b}}.

\bibitem[Lu et~al.(2022)Lu, Mishra, Xia, Qiu, Chang, Zhu, Tafjord, Clark, and
  Kalyan]{lu2022learn-scienceqa}
P.~Lu, S.~Mishra, T.~Xia, L.~Qiu, K.-W. Chang, S.-C. Zhu, O.~Tafjord, P.~Clark,
  and A.~Kalyan.
\newblock Learn to explain: Multimodal reasoning via thought chains for science
  question answering.
\newblock In \emph{The 36th Conference on Neural Information Processing Systems
  (NeurIPS)}, 2022.

\bibitem[Luo et~al.(2023)Luo, Rockwell, Lee, and Johnson]{luo2023cap3d}
T.~Luo, C.~Rockwell, H.~Lee, and J.~Johnson.
\newblock Scalable 3d captioning with pretrained models, 2023.
\newblock URL \url{https://arxiv.org/abs/2306.07279}.

\bibitem[Mao et~al.(2016)Mao, Huang, Toshev, Camburu, Yuille, and
  Murphy]{refcocog}
J.~Mao, J.~Huang, A.~Toshev, O.~Camburu, A.~L. Yuille, and K.~Murphy.
\newblock Generation and comprehension of unambiguous object descriptions.
\newblock In \emph{Proceedings of the IEEE Conference on Computer Vision and
  Pattern Recognition (CVPR)}, June 2016.

\bibitem[Marino et~al.(2019)Marino, Rastegari, Farhadi, and Mottaghi]{okvqa}
K.~Marino, M.~Rastegari, A.~Farhadi, and R.~Mottaghi.
\newblock Ok-vqa: A visual question answering benchmark requiring external
  knowledge.
\newblock In \emph{Conference on Computer Vision and Pattern Recognition
  (CVPR)}, 2019.

\bibitem[Masry et~al.(2022)Masry, Do, Tan, Joty, and
  Hoque]{masry-etal-2022-chartqa}
A.~Masry, X.~L. Do, J.~Q. Tan, S.~Joty, and E.~Hoque.
\newblock {C}hart{QA}: A benchmark for question answering about charts with
  visual and logical reasoning.
\newblock In S.~Muresan, P.~Nakov, and A.~Villavicencio, editors,
  \emph{Findings of the Association for Computational Linguistics: ACL 2022},
  pages 2263--2279, Dublin, Ireland, May 2022. Association for Computational
  Linguistics.
\newblock \doi{10.18653/v1/2022.findings-acl.177}.
\newblock URL \url{https://aclanthology.org/2022.findings-acl.177}.

\bibitem[Mathew et~al.(2020)Mathew, Karatzas, Manmatha, and
  Jawahar]{DBLP:journals/corr/abs-2007-00398-DOCVQA}
M.~Mathew, D.~Karatzas, R.~Manmatha, and C.~V. Jawahar.
\newblock Docvqa: {A} dataset for {VQA} on document images.
\newblock \emph{CoRR}, abs/2007.00398, 2020.
\newblock URL \url{https://arxiv.org/abs/2007.00398}.

\bibitem[Mathew et~al.(2022)Mathew, Bagal, Tito, Karatzas, Valveny, and
  Jawahar]{Mathew_2022-InfoVQA}
M.~Mathew, V.~Bagal, R.~Tito, D.~Karatzas, E.~Valveny, and C.~V. Jawahar.
\newblock Infographicvqa.
\newblock In \emph{2022 IEEE/CVF Winter Conference on Applications of Computer
  Vision (WACV)}. IEEE, Jan. 2022.
\newblock \doi{10.1109/wacv51458.2022.00264}.
\newblock URL \url{http://dx.doi.org/10.1109/WACV51458.2022.00264}.

\bibitem[McKinzie et~al.(2024)McKinzie, Gan, Fauconnier, Dodge, Zhang, Dufter,
  Shah, Du, Peng, Weers, Belyi, Zhang, Singh, Kang, Jain, H{\`{e}}, Schwarzer,
  Gunter, Kong, Zhang, Wang, Wang, Du, Lei, Wiseman, Yin, Lee, Wang, Pang,
  Grasch, Toshev, and Yang]{mm1}
B.~McKinzie, Z.~Gan, J.~Fauconnier, S.~Dodge, B.~Zhang, P.~Dufter, D.~Shah,
  X.~Du, F.~Peng, F.~Weers, A.~Belyi, H.~Zhang, K.~Singh, D.~Kang, A.~Jain,
  H.~H{\`{e}}, M.~Schwarzer, T.~Gunter, X.~Kong, A.~Zhang, J.~Wang, C.~Wang,
  N.~Du, T.~Lei, S.~Wiseman, G.~Yin, M.~Lee, Z.~Wang, R.~Pang, P.~Grasch,
  A.~Toshev, and Y.~Yang.
\newblock {MM1:} methods, analysis {\&} insights from multimodal {LLM}
  pre-training.
\newblock \emph{CoRR}, abs/2403.09611, 2024.
\newblock \doi{10.48550/ARXIV.2403.09611}.
\newblock URL \url{https://doi.org/10.48550/arXiv.2403.09611}.

\bibitem[Mesnard et~al.(2024)Mesnard, Hardin, Dadashi, Bhupatiraju, Pathak,
  Sifre, Rivi{\`{e}}re, Kale, Love, Tafti, Hussenot, Chowdhery, Roberts, Barua,
  Botev, Castro{-}Ros, Slone, H{\'{e}}liou, Tacchetti, Bulanova, Paterson,
  Tsai, Shahriari, Lan, Choquette{-}Choo, Crepy, Cer, Ippolito, Reid,
  Buchatskaya, Ni, Noland, Yan, Tucker, Muraru, Rozhdestvenskiy, Michalewski,
  Tenney, Grishchenko, Austin, Keeling, Labanowski, Lespiau, Stanway, Brennan,
  Chen, Ferret, Chiu, and et~al.]{gemma}
T.~Mesnard, C.~Hardin, R.~Dadashi, S.~Bhupatiraju, S.~Pathak, L.~Sifre,
  M.~Rivi{\`{e}}re, M.~S. Kale, J.~Love, P.~Tafti, L.~Hussenot, A.~Chowdhery,
  A.~Roberts, A.~Barua, A.~Botev, A.~Castro{-}Ros, A.~Slone, A.~H{\'{e}}liou,
  A.~Tacchetti, A.~Bulanova, A.~Paterson, B.~Tsai, B.~Shahriari, C.~L. Lan,
  C.~A. Choquette{-}Choo, C.~Crepy, D.~Cer, D.~Ippolito, D.~Reid,
  E.~Buchatskaya, E.~Ni, E.~Noland, G.~Yan, G.~Tucker, G.~Muraru,
  G.~Rozhdestvenskiy, H.~Michalewski, I.~Tenney, I.~Grishchenko, J.~Austin,
  J.~Keeling, J.~Labanowski, J.~Lespiau, J.~Stanway, J.~Brennan, J.~Chen,
  J.~Ferret, J.~Chiu, and et~al.
\newblock Gemma: Open models based on gemini research and technology.
\newblock \emph{CoRR}, abs/2403.08295, 2024.
\newblock \doi{10.48550/ARXIV.2403.08295}.
\newblock URL \url{https://doi.org/10.48550/arXiv.2403.08295}.

\bibitem[Minderer et~al.(2023)Minderer, Gritsenko, and Houlsby]{owlvitv2}
M.~Minderer, A.~A. Gritsenko, and N.~Houlsby.
\newblock Scaling open-vocabulary object detection.
\newblock In \emph{Thirty-seventh Conference on Neural Information Processing
  Systems}, 2023.
\newblock URL \url{https://openreview.net/forum?id=mQPNcBWjGc}.

\bibitem[Mishra et~al.(2019)Mishra, Shekhar, Singh, and
  Chakraborty]{mishraICDAR19-ocrvqa}
A.~Mishra, S.~Shekhar, A.~K. Singh, and A.~Chakraborty.
\newblock Ocr-vqa: Visual question answering by reading text in images.
\newblock In \emph{ICDAR}, 2019.

\bibitem[Nguyen et~al.(2024)Nguyen, Wallingford, Santy, Ma, Oh, Schmidt, Koh,
  and Krishna]{nguyen2024multilingual}
T.~Nguyen, M.~Wallingford, S.~Santy, W.~Ma, S.~Oh, L.~Schmidt, P.~W. Koh, and
  R.~Krishna.
\newblock Multilingual diversity improves vision-language representations.
\newblock \emph{CoRR}, abs/2405.16915, 2024.
\newblock \doi{10.48550/ARXIV.2405.16915}.
\newblock URL \url{https://doi.org/10.48550/arXiv.2405.16915}.

\bibitem[Ning et~al.(2023)Ning, Li, Zhang, Geng, Dai, He, and
  Hu]{ning2023allin}
J.~Ning, C.~Li, Z.~Zhang, Z.~Geng, Q.~Dai, K.~He, and H.~Hu.
\newblock All in tokens: Unifying output space of visual tasks via soft token.
\newblock \emph{arXiv preprint arXiv:2301.02229}, 2023.

\bibitem[OpenAI(2023)]{openai2024gpt4}
OpenAI.
\newblock Gpt-4 technical report, 2023.
\newblock URL \url{https://arxiv.org/abs/2303.08774}.

\bibitem[Paiss et~al.(2023)Paiss, Ephrat, Tov, Zada, Mosseri, Irani, and
  Dekel]{paiss2023countclip}
R.~Paiss, A.~Ephrat, O.~Tov, S.~Zada, I.~Mosseri, M.~Irani, and T.~Dekel.
\newblock Teaching clip to count to ten.
\newblock \emph{arXiv preprint arXiv:2302.12066}, 2023.

\bibitem[Peng et~al.(2022)Peng, Dong, Bao, Ye, and Wei]{beit2}
Z.~Peng, L.~Dong, H.~Bao, Q.~Ye, and F.~Wei.
\newblock Beit v2: Masked image modeling with vector-quantized visual
  tokenizers.
\newblock \emph{CoRR}, abs/2208.06366, 2022.
\newblock \doi{10.48550/ARXIV.2208.06366}.
\newblock URL \url{https://doi.org/10.48550/arXiv.2208.06366}.

\bibitem[Pfeiffer et~al.(2022)Pfeiffer, Geigle, Kamath, Steitz, Roth,
  Vuli{\'c}, and Gurevych]{pfeiffer-etal-2022-xgqa}
J.~Pfeiffer, G.~Geigle, A.~Kamath, J.-M. Steitz, S.~Roth, I.~Vuli{\'c}, and
  I.~Gurevych.
\newblock x{GQA}: Cross-lingual visual question answering.
\newblock In \emph{Findings of the Association for Computational Linguistics:
  ACL 2022}, pages 2497--2511, Dublin, Ireland, May 2022. Association for
  Computational Linguistics.
\newblock \doi{10.18653/v1/2022.findings-acl.196}.
\newblock URL \url{https://aclanthology.org/2022.findings-acl.196}.

\bibitem[Piergiovanni et~al.(2022)Piergiovanni, Kuo, and
  Angelova]{Piergiovanni}
A.~Piergiovanni, W.~Kuo, and A.~Angelova.
\newblock Pre-training image-language transformers for open-vocabulary tasks,
  2022.
\newblock URL \url{https://arxiv.org/abs/2209.04372}.

\bibitem[Pouget et~al.(2024)Pouget, Beyer, Bugliarello, Wang, Steiner, Zhai,
  and Alabdulmohsin]{nofilter}
A.~Pouget, L.~Beyer, E.~Bugliarello, X.~Wang, A.~P. Steiner, X.~Zhai, and
  I.~Alabdulmohsin.
\newblock No filter: Cultural and socioeconomic diversity in contrastive
  vision-language models.
\newblock \emph{CoRR}, abs/2405.13777, 2024.
\newblock \doi{10.48550/ARXIV.2405.13777}.
\newblock URL \url{https://doi.org/10.48550/arXiv.2405.13777}.

\bibitem[Qiu et~al.(2022)Qiu, Onea{\textcommabelow{t}}{\u{a}}, Bugliarello,
  Frank, and Elliott]{qiu-etal-2022-multilingual}
C.~Qiu, D.~Onea{\textcommabelow{t}}{\u{a}}, E.~Bugliarello, S.~Frank, and
  D.~Elliott.
\newblock Multilingual multimodal learning with machine translated text.
\newblock In Y.~Goldberg, Z.~Kozareva, and Y.~Zhang, editors, \emph{Findings of
  the Association for Computational Linguistics: EMNLP 2022}, pages 4178--4193,
  Abu Dhabi, United Arab Emirates, Dec. 2022. Association for Computational
  Linguistics.
\newblock \doi{10.18653/v1/2022.findings-emnlp.308}.
\newblock URL \url{https://aclanthology.org/2022.findings-emnlp.308}.

\bibitem[Radford et~al.(2021)Radford, Kim, Hallacy, Ramesh, Goh, Agarwal,
  Sastry, Askell, Mishkin, Clark, Krueger, and Sutskever]{clip}
A.~Radford, J.~W. Kim, C.~Hallacy, A.~Ramesh, G.~Goh, S.~Agarwal, G.~Sastry,
  A.~Askell, P.~Mishkin, J.~Clark, G.~Krueger, and I.~Sutskever.
\newblock Learning transferable visual models from natural language
  supervision.
\newblock In \emph{International Conference on Machine Learning, {ICML}}, 2021.

\bibitem[Raffel et~al.(2020)Raffel, Shazeer, Roberts, Lee, Narang, Matena,
  Zhou, Li, and Liu]{t5}
C.~Raffel, N.~Shazeer, A.~Roberts, K.~Lee, S.~Narang, M.~Matena, Y.~Zhou,
  W.~Li, and P.~J. Liu.
\newblock Exploring the limits of transfer learning with a unified text-to-text
  transformer.
\newblock \emph{J. Mach. Learn. Res.}, 21:\penalty0 140:1--140:67, 2020.
\newblock URL \url{http://jmlr.org/papers/v21/20-074.html}.

\bibitem[Rajbhandari et~al.(2020)Rajbhandari, Rasley, Ruwase, and He]{zero}
S.~Rajbhandari, J.~Rasley, O.~Ruwase, and Y.~He.
\newblock {ZeRO}: memory optimizations toward training trillion parameter
  models.
\newblock In C.~Cuicchi, I.~Qualters, and W.~T. Kramer, editors,
  \emph{Proceedings of the International Conference for High Performance
  Computing, Networking, Storage and Analysis, {SC} 2020, Virtual Event /
  Atlanta, Georgia, USA, November 9-19, 2020}, page~20. {IEEE/ACM}, 2020.
\newblock \doi{10.1109/SC41405.2020.00024}.
\newblock URL \url{https://doi.org/10.1109/SC41405.2020.00024}.

\bibitem[Sabne(2020)]{xla}
A.~Sabne.
\newblock {XLA} : Compiling machine learning for peak performance, 2020.

\bibitem[Schwenk et~al.(2022)Schwenk, Khandelwal, Clark, Marino, and
  Mottaghi]{AOKVQA}
D.~Schwenk, A.~Khandelwal, C.~Clark, K.~Marino, and R.~Mottaghi.
\newblock A-okvqa: A benchmark for visual question answering using world
  knowledge.
\newblock \emph{arXiv}, 2022.

\bibitem[Shoeybi et~al.(2019)Shoeybi, Patwary, Puri, LeGresley, Casper, and
  Catanzaro]{megatron}
M.~Shoeybi, M.~Patwary, R.~Puri, P.~LeGresley, J.~Casper, and B.~Catanzaro.
\newblock Megatron-{LM}: Training multi-billion parameter language models using
  model parallelism.
\newblock \emph{CoRR}, abs/1909.08053, 2019.
\newblock URL \url{http://arxiv.org/abs/1909.08053}.

\bibitem[Sidorov et~al.(2020)Sidorov, Hu, Rohrbach, and
  Singh]{sidorov2019textcaps}
O.~Sidorov, R.~Hu, M.~Rohrbach, and A.~Singh.
\newblock Textcaps: {A} dataset for image captioning with reading
  comprehension.
\newblock In A.~Vedaldi, H.~Bischof, T.~Brox, and J.~Frahm, editors,
  \emph{Computer Vision - {ECCV} 2020 - 16th European Conference, Glasgow, UK,
  August 23-28, 2020, Proceedings, Part {II}}, volume 12347 of \emph{Lecture
  Notes in Computer Science}, pages 742--758. Springer, 2020.
\newblock \doi{10.1007/978-3-030-58536-5\_44}.
\newblock URL \url{https://doi.org/10.1007/978-3-030-58536-5\_44}.

\bibitem[Singh et~al.(2019)Singh, Natarjan, Shah, Jiang, Chen, Parikh, and
  Rohrbach]{singh2019towards-textvqa}
A.~Singh, V.~Natarjan, M.~Shah, Y.~Jiang, X.~Chen, D.~Parikh, and M.~Rohrbach.
\newblock Towards vqa models that can read.
\newblock In \emph{Proceedings of the IEEE Conference on Computer Vision and
  Pattern Recognition}, pages 8317--8326, 2019.

\bibitem[Suhr et~al.(2019)Suhr, Zhou, Zhang, Zhang, Bai, and
  Artzi]{suhr-etal-2019-corpus-nlvr2}
A.~Suhr, S.~Zhou, A.~Zhang, I.~Zhang, H.~Bai, and Y.~Artzi.
\newblock A corpus for reasoning about natural language grounded in
  photographs.
\newblock In A.~Korhonen, D.~Traum, and L.~Marquez, editors, \emph{Proceedings
  of the 57th Annual Meeting of the Association for Computational Linguistics},
  pages 6418--6428, Florence, Italy, July 2019. Association for Computational
  Linguistics.
\newblock \doi{10.18653/v1/P19-1644}.
\newblock URL \url{https://aclanthology.org/P19-1644}.

\bibitem[Sun et~al.(2023)Sun, Cui, Zhang, Zhang, Yu, Luo, Wang, Rao, Liu,
  Huang, and Wang]{emu2}
Q.~Sun, Y.~Cui, X.~Zhang, F.~Zhang, Q.~Yu, Z.~Luo, Y.~Wang, Y.~Rao, J.~Liu,
  T.~Huang, and X.~Wang.
\newblock Generative multimodal models are in-context learners.
\newblock \emph{CoRR}, abs/2312.13286, 2023.
\newblock \doi{10.48550/ARXIV.2312.13286}.
\newblock URL \url{https://doi.org/10.48550/arXiv.2312.13286}.

\bibitem[Tay et~al.(2023)Tay, Dehghani, Tran, Garcia, Wei, Wang, Chung, Bahri,
  Schuster, Zheng, Zhou, Houlsby, and Metzler]{ul_2}
Y.~Tay, M.~Dehghani, V.~Q. Tran, X.~Garcia, J.~Wei, X.~Wang, H.~W. Chung,
  D.~Bahri, T.~Schuster, H.~S. Zheng, D.~Zhou, N.~Houlsby, and D.~Metzler.
\newblock {UL2:} unifying language learning paradigms.
\newblock In \emph{The Eleventh International Conference on Learning
  Representations, {ICLR} 2023, Kigali, Rwanda, May 1-5, 2023}. OpenReview.net,
  2023.
\newblock URL \url{https://openreview.net/pdf?id=6ruVLB727MC}.

\bibitem[Team(2024)]{chameleon}
C.~Team.
\newblock Chameleon: Mixed-modal early-fusion foundation models.
\newblock \emph{CoRR}, abs/2405.09818, 2024.
\newblock \doi{10.48550/ARXIV.2405.09818}.
\newblock URL \url{https://doi.org/10.48550/arXiv.2405.09818}.

\bibitem[Thapliyal et~al.(2022)Thapliyal, Pont~Tuset, Chen, and
  Soricut]{thapliyal-etal-2022-crossmodal-COCO35L-XM3600}
A.~V. Thapliyal, J.~Pont~Tuset, X.~Chen, and R.~Soricut.
\newblock Crossmodal-3600: A massively multilingual multimodal evaluation
  dataset.
\newblock In Y.~Goldberg, Z.~Kozareva, and Y.~Zhang, editors, \emph{Proceedings
  of the 2022 Conference on Empirical Methods in Natural Language Processing},
  pages 715--729, Abu Dhabi, United Arab Emirates, Dec. 2022. Association for
  Computational Linguistics.
\newblock \doi{10.18653/v1/2022.emnlp-main.45}.
\newblock URL \url{https://aclanthology.org/2022.emnlp-main.45}.

\bibitem[Tong et~al.(2024{\natexlab{a}})Tong, Brown, Wu, Woo, Middepogu, Akula,
  Yang, Yang, Iyer, Pan, Wang, Fergus, LeCun, and Xie]{cambrian1}
S.~Tong, E.~Brown, P.~Wu, S.~Woo, M.~Middepogu, S.~C. Akula, J.~Yang, S.~Yang,
  A.~Iyer, X.~Pan, A.~Wang, R.~Fergus, Y.~LeCun, and S.~Xie.
\newblock {Cambrian-1: A Fully Open, Vision-Centric Exploration of Multimodal
  LLMs}.
\newblock \emph{arXiv preprint arXiv:2406.16860}, 2024{\natexlab{a}}.

\bibitem[Tong et~al.(2024{\natexlab{b}})Tong, Liu, Zhai, Ma, LeCun, and
  Xie]{tong2024eyes}
S.~Tong, Z.~Liu, Y.~Zhai, Y.~Ma, Y.~LeCun, and S.~Xie.
\newblock Eyes wide shut? exploring the visual shortcomings of multimodal llms,
  2024{\natexlab{b}}.

\bibitem[Touvron et~al.(2022)Touvron, Vedaldi, Douze, and
  Jégou]{touvron2022fixres}
H.~Touvron, A.~Vedaldi, M.~Douze, and H.~Jégou.
\newblock Fixing the train-test resolution discrepancy, 2022.
\newblock URL \url{https://arxiv.org/abs/1906.06423}.

\bibitem[Tschannen et~al.(2023)Tschannen, Kumar, Steiner, Zhai, Houlsby, and
  Beyer]{cappa}
M.~Tschannen, M.~Kumar, A.~Steiner, X.~Zhai, N.~Houlsby, and L.~Beyer.
\newblock Image captioners are scalable vision learners too.
\newblock \emph{NeurIPS}, 2023.

\bibitem[Tsimpoukelli et~al.(2021)Tsimpoukelli, Menick, Cabi, Eslami, Vinyals,
  and Hill]{tsimpoukelli21frozen}
M.~Tsimpoukelli, J.~Menick, S.~Cabi, S.~M.~A. Eslami, O.~Vinyals, and F.~Hill.
\newblock Multimodal few-shot learning with frozen language models.
\newblock \emph{CoRR}, abs/2106.13884, 2021.
\newblock URL \url{https://arxiv.org/abs/2106.13884}.

\bibitem[van~den Oord et~al.(2017)van~den Oord, Vinyals, and
  Kavukcuoglu]{vqvae}
A.~van~den Oord, O.~Vinyals, and K.~Kavukcuoglu.
\newblock Neural discrete representation learning.
\newblock In I.~Guyon, U.~von Luxburg, S.~Bengio, H.~M. Wallach, R.~Fergus,
  S.~V.~N. Vishwanathan, and R.~Garnett, editors, \emph{Advances in Neural
  Information Processing Systems 30: Annual Conference on Neural Information
  Processing Systems 2017, December 4-9, 2017, Long Beach, CA, {USA}}, pages
  6306--6315, 2017.

\bibitem[Vikhyat(2024)]{moondream}
Vikhyat.
\newblock Moondream.
\newblock \url{https://github.com/vikhyat/moondream}, 2024.
\newblock Accessed: 2024-07-04.

\bibitem[Visheratin(2024)]{visheratin2024curse}
A.~Visheratin.
\newblock Breaking resolution curse of vision-language models.
\newblock \url{https://huggingface.co/blog/visheratin/vlm-resolution-curse},
  2024.

\bibitem[Wan et~al.(2024)Wan, Tschannen, Xian, Pavetic, Alabdulmohsin, Wang,
  Pinto, Steiner, Beyer, and Zhai]{locca}
B.~Wan, M.~Tschannen, Y.~Xian, F.~Pavetic, I.~Alabdulmohsin, X.~Wang, A.~S.
  Pinto, A.~Steiner, L.~Beyer, and X.~Zhai.
\newblock Locca: Visual pretraining with location-aware captioners.
\newblock \emph{CoRR}, abs/2403.19596, 2024.
\newblock \doi{10.48550/ARXIV.2403.19596}.
\newblock URL \url{https://doi.org/10.48550/arXiv.2403.19596}.

\bibitem[Wang et~al.(2021)Wang, Li, Zhou, Chen, Grossman, and
  Li]{wang2021screen2words}
B.~Wang, G.~Li, X.~Zhou, Z.~Chen, T.~Grossman, and Y.~Li.
\newblock Screen2words: Automatic mobile ui summarization with multimodal
  learning.
\newblock In \emph{The 34th Annual ACM Symposium on User Interface Software and
  Technology}, pages 498--510, 2021.

\bibitem[Wang et~al.(2022{\natexlab{a}})Wang, Bao, Dong, Bjorck, Peng, Liu,
  Aggarwal, Mohammed, Singhal, Som, and Wei]{beit3}
W.~Wang, H.~Bao, L.~Dong, J.~Bjorck, Z.~Peng, Q.~Liu, K.~Aggarwal, O.~K.
  Mohammed, S.~Singhal, S.~Som, and F.~Wei.
\newblock Image as a foreign language: Beit pretraining for all vision and
  vision-language tasks.
\newblock \emph{CoRR}, abs/2208.10442, 2022{\natexlab{a}}.
\newblock \doi{10.48550/ARXIV.2208.10442}.
\newblock URL \url{https://doi.org/10.48550/arXiv.2208.10442}.

\bibitem[Wang et~al.(2023)Wang, Lv, Yu, Hong, Qi, Wang, Ji, Yang, Zhao, Song,
  Xu, Xu, Li, Dong, Ding, and Tang]{cogvlm}
W.~Wang, Q.~Lv, W.~Yu, W.~Hong, J.~Qi, Y.~Wang, J.~Ji, Z.~Yang, L.~Zhao,
  X.~Song, J.~Xu, B.~Xu, J.~Li, Y.~Dong, M.~Ding, and J.~Tang.
\newblock Cogvlm: Visual expert for pretrained language models.
\newblock \emph{CoRR}, abs/2311.03079, 2023.
\newblock \doi{10.48550/ARXIV.2311.03079}.
\newblock URL \url{https://doi.org/10.48550/arXiv.2311.03079}.

\bibitem[Wang et~al.(2019)Wang, Wu, Chen, Li, Wang, and Wang]{wang2019vatex}
X.~Wang, J.~Wu, J.~Chen, L.~Li, Y.-F. Wang, and W.~Y. Wang.
\newblock Vatex: A large-scale, high-quality multilingual dataset for
  video-and-language research.
\newblock In \emph{The IEEE International Conference on Computer Vision
  (ICCV)}, October 2019.

\bibitem[Wang et~al.(2022{\natexlab{b}})Wang, Yu, Yu, Dai, Tsvetkov, and
  Cao]{wang2021simvlm}
Z.~Wang, J.~Yu, A.~W. Yu, Z.~Dai, Y.~Tsvetkov, and Y.~Cao.
\newblock Simvlm: Simple visual language model pretraining with weak
  supervision, 2022{\natexlab{b}}.
\newblock URL \url{https://arxiv.org/abs/2108.10904}.

\bibitem[Wu and Xie(2023)]{wu2023vstar}
P.~Wu and S.~Xie.
\newblock V*: Guided visual search as a core mechanism in multimodal llms,
  2023.
\newblock URL \url{https://arxiv.org/abs/2312.14135}.

\bibitem[Xiao et~al.(2023)Xiao, Wu, Xu, Dai, Hu, Lu, Zeng, Liu, and
  Yuan]{florence2}
B.~Xiao, H.~Wu, W.~Xu, X.~Dai, H.~Hu, Y.~Lu, M.~Zeng, C.~Liu, and L.~Yuan.
\newblock Florence-2: Advancing a unified representation for a variety of
  vision tasks.
\newblock \emph{CoRR}, abs/2311.06242, 2023.
\newblock \doi{10.48550/ARXIV.2311.06242}.
\newblock URL \url{https://doi.org/10.48550/arXiv.2311.06242}.

\bibitem[Xu et~al.(2017)Xu, Zhao, Xiao, Wu, Zhang, He, and Zhuang]{xu2017video}
D.~Xu, Z.~Zhao, J.~Xiao, F.~Wu, H.~Zhang, X.~He, and Y.~Zhuang.
\newblock Video question answering via gradually refined attention over
  appearance and motion.
\newblock In \emph{ACM Multimedia}, 2017.

\bibitem[Xu et~al.(2016)Xu, Mei, Yao, and Rui]{msrvtt}
J.~Xu, T.~Mei, T.~Yao, and Y.~Rui.
\newblock Msr-vtt: A large video description dataset for bridging video and
  language.
\newblock In \emph{2016 IEEE Conference on Computer Vision and Pattern
  Recognition (CVPR)}, pages 5288--5296, 2016.
\newblock \doi{10.1109/CVPR.2016.571}.

\bibitem[Xu et~al.(2021)Xu, Lee, Chen, Hechtman, Huang, Joshi, Krikun,
  Lepikhin, Ly, Maggioni, Pang, Shazeer, Wang, Wang, Wu, and Chen]{gspmd}
Y.~Xu, H.~Lee, D.~Chen, B.~A. Hechtman, Y.~Huang, R.~Joshi, M.~Krikun,
  D.~Lepikhin, A.~Ly, M.~Maggioni, R.~Pang, N.~Shazeer, S.~Wang, T.~Wang,
  Y.~Wu, and Z.~Chen.
\newblock {GSPMD:} general and scalable parallelization for {ML} computation
  graphs.
\newblock \emph{CoRR}, abs/2105.04663, 2021.
\newblock URL \url{https://arxiv.org/abs/2105.04663}.

\bibitem[Xue et~al.(2020)Xue, Constant, Roberts, Kale, Al-Rfou, Siddhant,
  Barua, and Raffel]{xue2020mt5}
L.~Xue, N.~Constant, A.~Roberts, M.~Kale, R.~Al-Rfou, A.~Siddhant, A.~Barua,
  and C.~Raffel.
\newblock mt5: A massively multilingual pre-trained text-to-text transformer.
\newblock \emph{arXiv preprint arXiv:2010.11934}, 2020.

\bibitem[Yifan~Li and Wen(2023)]{Li23pope}
K.~Z. J. W. W. X.~Z. Yifan~Li, Yifan~Du and J.-R. Wen.
\newblock Evaluating object hallucination in large vision-language models.
\newblock In \emph{The 2023 Conference on Empirical Methods in Natural Language
  Processing}, 2023.
\newblock URL \url{https://openreview.net/forum?id=xozJw0kZXF}.

\bibitem[Yu et~al.(2016)Yu, Poirson, Yang, Berg, and
  Berg]{DBLP:conf/eccv/YuPYBB16-refcoco}
L.~Yu, P.~Poirson, S.~Yang, A.~C. Berg, and T.~L. Berg.
\newblock Modeling context in referring expressions.
\newblock In B.~Leibe, J.~Matas, N.~Sebe, and M.~Welling, editors,
  \emph{Computer Vision - {ECCV} 2016 - 14th European Conference, Amsterdam,
  The Netherlands, October 11-14, 2016, Proceedings, Part {II}}, volume 9906 of
  \emph{Lecture Notes in Computer Science}, pages 69--85. Springer, 2016.
\newblock \doi{10.1007/978-3-319-46475-6\_5}.
\newblock URL \url{https://doi.org/10.1007/978-3-319-46475-6\_5}.

\bibitem[Yu et~al.(2023)Yu, Shi, Pasunuru, Muller, Golovneva, Wang, Babu, Tang,
  Karrer, Sheynin, Ross, Polyak, Howes, Sharma, Xu, Tamoyan, Ashual, Singer,
  Li, Zhang, James, Ghosh, Taigman, Fazel{-}Zarandi, Celikyilmaz, Zettlemoyer,
  and Aghajanyan]{cm3leon}
L.~Yu, B.~Shi, R.~Pasunuru, B.~Muller, O.~Golovneva, T.~Wang, A.~Babu, B.~Tang,
  B.~Karrer, S.~Sheynin, C.~Ross, A.~Polyak, R.~Howes, V.~Sharma, P.~Xu,
  H.~Tamoyan, O.~Ashual, U.~Singer, S.~Li, S.~Zhang, R.~James, G.~Ghosh,
  Y.~Taigman, M.~Fazel{-}Zarandi, A.~Celikyilmaz, L.~Zettlemoyer, and
  A.~Aghajanyan.
\newblock Scaling autoregressive multi-modal models: Pretraining and
  instruction tuning.
\newblock \emph{CoRR}, abs/2309.02591, 2023.
\newblock \doi{10.48550/ARXIV.2309.02591}.
\newblock URL \url{https://doi.org/10.48550/arXiv.2309.02591}.

\bibitem[Yu et~al.(2019)Yu, Xu, Yu, Yu, Zhao, Zhuang, and
  Tao]{yu2019activitynet}
Z.~Yu, D.~Xu, J.~Yu, T.~Yu, Z.~Zhao, Y.~Zhuang, and D.~Tao.
\newblock Activitynet-qa: {A} dataset for understanding complex web videos via
  question answering.
\newblock In \emph{The Thirty-Third {AAAI} Conference on Artificial
  Intelligence, {AAAI} 2019, The Thirty-First Innovative Applications of
  Artificial Intelligence Conference, {IAAI} 2019, The Ninth {AAAI} Symposium
  on Educational Advances in Artificial Intelligence, {EAAI} 2019, Honolulu,
  Hawaii, USA, January 27 - February 1, 2019}, pages 9127--9134. {AAAI} Press,
  2019.
\newblock \doi{10.1609/AAAI.V33I01.33019127}.
\newblock URL \url{https://doi.org/10.1609/aaai.v33i01.33019127}.

\bibitem[Zhai et~al.(2022{\natexlab{a}})Zhai, Kolesnikov, Houlsby, and
  Beyer]{scalingvit}
X.~Zhai, A.~Kolesnikov, N.~Houlsby, and L.~Beyer.
\newblock Scaling vision transformers.
\newblock \emph{Computer Vision and Pattern Recognition (CVPR)},
  2022{\natexlab{a}}.

\bibitem[Zhai et~al.(2022{\natexlab{b}})Zhai, Wang, Mustafa, Steiner, Keysers,
  Kolesnikov, and Beyer]{lit}
X.~Zhai, X.~Wang, B.~Mustafa, A.~Steiner, D.~Keysers, A.~Kolesnikov, and
  L.~Beyer.
\newblock Lit: Zero-shot transfer with locked-image text tuning.
\newblock In \emph{Computer Vision and Pattern Recognition (CVPR)},
  2022{\natexlab{b}}.

\bibitem[Zhai et~al.(2023)Zhai, Mustafa, Kolesnikov, and Beyer]{siglip}
X.~Zhai, B.~Mustafa, A.~Kolesnikov, and L.~Beyer.
\newblock Sigmoid loss for language image pre-training.
\newblock In \emph{ICCV}, pages 11941--11952, 2023.

\bibitem[Zhang et~al.(2024{\natexlab{a}})Zhang, You, Dufter, Zhang, Chen, Chen,
  Fu, Wang, Chang, Gan, and Yang]{zhang2024ferretv2}
H.~Zhang, H.~You, P.~Dufter, B.~Zhang, C.~Chen, H.-Y. Chen, T.-J. Fu, W.~Y.
  Wang, S.-F. Chang, Z.~Gan, and Y.~Yang.
\newblock Ferret-v2: An improved baseline for referring and grounding with
  large language models, 2024{\natexlab{a}}.
\newblock URL \url{https://arxiv.org/abs/2404.07973}.

\bibitem[Zhang et~al.(2020)Zhang, Jiang, Miura, Manning, and
  Langlotz]{zhang2020contrastive}
Y.~Zhang, H.~Jiang, Y.~Miura, C.~D. Manning, and C.~P. Langlotz.
\newblock Contrastive learning of medical visual representations from paired
  images and text.
\newblock \emph{arXiv preprint arXiv:2010.00747}, 2020.

\bibitem[Zhang et~al.(2024{\natexlab{b}})Zhang, Unell, Wang, Ghosh, Su,
  Schmidt, and Yeung-Levy]{zhang2024vlmbadclf}
Y.~Zhang, A.~Unell, X.~Wang, D.~Ghosh, Y.~Su, L.~Schmidt, and S.~Yeung-Levy.
\newblock Why are visually-grounded language models bad at image
  classification?, 2024{\natexlab{b}}.
\newblock URL \url{https://arxiv.org/abs/2405.18415}.

\bibitem[Zhao et~al.(2023)Zhao, Gu, Varma, Luo, Huang, Xu, Wright, Shojanazeri,
  Ott, Shleifer, Desmaison, Balioglu, Damania, Nguyen, Chauhan, Hao, Mathews,
  and Li]{fsdp}
Y.~Zhao, A.~Gu, R.~Varma, L.~Luo, C.~Huang, M.~Xu, L.~Wright, H.~Shojanazeri,
  M.~Ott, S.~Shleifer, A.~Desmaison, C.~Balioglu, P.~Damania, B.~Nguyen,
  G.~Chauhan, Y.~Hao, A.~Mathews, and S.~Li.
\newblock Pytorch {FSDP:} experiences on scaling fully sharded data parallel.
\newblock \emph{Proc. {VLDB} Endow.}, 16\penalty0 (12):\penalty0 3848--3860,
  2023.
\newblock \doi{10.14778/3611540.3611569}.
\newblock URL \url{https://www.vldb.org/pvldb/vol16/p3848-huang.pdf}.

\bibitem[Zhou et~al.(2020)Zhou, Palangi, Zhang, Hu, Corso, and
  Gao]{zhou20unified}
L.~Zhou, H.~Palangi, L.~Zhang, H.~Hu, J.~J. Corso, and J.~Gao.
\newblock Unified vision-language pre-training for image captioning and {VQA}.
\newblock In \emph{The Thirty-Fourth {AAAI} Conference on Artificial
  Intelligence, {AAAI} 2020, The Thirty-Second Innovative Applications of
  Artificial Intelligence Conference, {IAAI} 2020, The Tenth {AAAI} Symposium
  on Educational Advances in Artificial Intelligence, {EAAI} 2020, New York,
  NY, USA, February 7-12, 2020}, pages 13041--13049. {AAAI} Press, 2020.
\newblock \doi{10.1609/AAAI.V34I07.7005}.
\newblock URL \url{https://doi.org/10.1609/aaai.v34i07.7005}.

\bibitem[Zou et~al.(2023)Zou, Dou, Yang, Gan, Li, Li, Dai, Behl, Wang, Yuan,
  Peng, Wang, Lee, and Gao]{xdecoder}
X.~Zou, Z.~Dou, J.~Yang, Z.~Gan, L.~Li, C.~Li, X.~Dai, H.~Behl, J.~Wang,
  L.~Yuan, N.~Peng, L.~Wang, Y.~J. Lee, and J.~Gao.
\newblock Generalized decoding for pixel, image, and language.
\newblock In \emph{Computer Vision and Pattern Recognition (CVPR)}, pages
  15116--15127, 2023.

\end{thebibliography}

\clearpage

\section*{Author Contributions}

\subsection*{Model development contributors}
\subsubsection*{Core Contributors}
Lucas Beyer \\
Andreas Steiner \\
André Susano Pinto \\
Alexander Kolesnikov \\
Xiao Wang \\
Xiaohua Zhai

\subsubsection*{Contributors}
Daniel Salz \\
Maxim Neumann \\
Ibrahim Alabdulmohsin \\
Michael Tschannen \\
Emanuele Bugliarello \\
Thomas Unterthiner \\
Daniel Keysers \\
Skanda Koppula \\
Fangyu Liu \\
Adam Grycner \\
Alexey Gritsenko \\
Neil Houlsby \\
Manoj Kumar \\
Keran Rong \\
Julian Eisenschlos \\
Rishabh Kabra \\
Matthias Bauer \\
Matko Bošnjak \\
Xi Chen \\
Matthias Minderer \\
Paul Voigtlaender \\
Ioana Bica \\
Ivana Balazevic \\
Joan Puigcerver \\
Pinelopi Papalampidi \\
Olivier Henaff \\
Xi Xiong \\
Radu Soricut \\
Jeremiah Harmsen

\subsubsection*{Leads}
Xiaohua Zhai \\
Lucas Beyer

\subsection*{Model release contributors and \\general support}

\subsubsection*{PM}
Tris Warkentin

\subsubsection*{Go-to-Market} 
Kat Black \\
Luiz Gustavo Martins \\
Glenn Cameron \\
Raj Gundluru \\
Manvinder Singh

\subsubsection*{Kaggle}
Meg Risdal \\
Nilay Chauhan \\
Nate Keating \\
Nesh Devanathan

\subsubsection*{Documentation}
Elisa Bandy \\
Joe Fernandez

\subsubsection*{Ethics and Safety}
Antonia Paterson \\
Jenny Brennan \\
Tom Eccles \\
Pankil Botadra \\
Ben Bariach

\subsubsection*{Vertex AI}
Lav Rai \\
Minwoo Park \\
Dustin Luong \\
Daniel Vlasic \\
Bo Wu \\
Wenming Ye

\subsubsection*{Keras}
Divyashree Sreepathihalli \\
Kiranbir Sodhia \\
Gabriel Rasskin  \\
Matthew Watson \\
Varun Singh

\subsubsection*{Gemma Model}
Alek Andreev \\
Armand Joulin \\
Surya Bhupatiraju \\
Minh Giang

\subsubsection*{Hugging Face Partners}
Arthur Zucker \\
Lucain Pouget \\
Merve Noyan \\
Omar Sanseviero \\
Pablo Montalvo \\
Pedro Cuenca

\subsubsection*{Nvidia Partners}
Maryam Moosaei \\
Fangzhou Mu \\
Santosh Bhavani \\
Anjali Shah \\
Vladislav Kozlov \\
Dong Meng \\
Niaz Syed \\
Chintan Patel \\
Ankit Patel \\
Marta Stepniewska-Dziubinska \\
Anna Warno \\
Xinqi Wang \\
David Tamok \\
Steven Baughman \\
Sandip Bhaskar \\
Jason Dudash \\

\subsubsection*{Executive Sponsors}
Joelle Barral \\
Zoubin Ghahramani

\clearpage
\appendix
\onecolumn  

\section{More related work}\label{sec:app:relwork}

There are generally two ways to build vision-language models (VLMs): the first option is to connect a vision encoder to a large language model while the second option is to use a transformer decoder-only architecture to handle both vision and language modalities.

It is popular to connect frozen image component and language component with lightweight adapters, e.g. linear or MLP projector, resampler~\cite{perceiver, alayrac2022flamingo}, Q-Former~\cite{blip2}. 
Flamingo~\cite{alayrac2022flamingo} uses a perceiver-based~\cite{perceiver} resampler to connect frozen vision and language models.
Idefics2~\cite{idefics2} shows that the perceiver resampler works significantly better than a linear projector.
BLIP2~\cite{blip2} explores using trainable Q-Former~\cite{blip2} to align frozen vision and language models. 
They first train the Q-Former with frozen image model. Then attach the frozen image model and the Q-Former into frozen language model to continue the Q-Former training.
LLaVA~\cite{llava1} opted to train a projection layer between frozen vision backbone and frozen language backbone, with GPT-4 generated small but high quality instruction-following data.
Afterwards, they unfreeze the language backbone and finetune the projection layer and language model together.
LLaVA-1.5~\cite{llava15} extends this to MLP connector.
Bunny~\cite{bunny} opted to use MLP connector for their models.
Honeybee~\cite{Honeybee} introduced locality-enhanced projectors by using convolution and deformable attention, for better spatial understanding.
MM1~\cite{mm1} claimed that convolution adaptor~\cite{Honeybee} performs close to average pooling and attention pooling baselines.
Cambrian-1~\cite{cambrian1} performed a thorough study of vision encoders for VLMs, and proposed spatial vision aggregator to better integrate visual tokens.
CogVLM~\cite{cogvlm} introduced additional trainable visual expert module in the attention and FFN layers of the frozen language model.
This way, the language model is able to process visual tokens and language tokens with different experts, while keeping the same level of performance for text-only tasks.
We performed an ablation study of linear and MLP vision-language connectors for PaliGemma in Section~\ref{sec:abl:connector}.

There are also methods exploring training both the vision and language components, PaliGemma falls into this category.
Many VLM systems ~\cite{pali, palix, pali3, chen2023internvl, qwen-vl} follow a multi-stage training procedure, including a stage to train both vision and language components. 
PaLI line of work~\cite{pali,palix,pali3} gradually scales up the training resolution with different data mixtures in three stages.
Florence2~\cite{florence2} and LocCa~\cite{locca} train vision-centric models by modeling very diverse tasks with a universal language interface.
Unified-IO 2~\cite{unifiedio2} trains a single encoder-decoder multimodal model on an ensemble of 120 datasets.
Kosmos~\cite{kosmos} trains the language model and the last layer of a CLIP vision model.
BLIP~\cite{blip} proposed to dump COCO like pseudo captions on million scale web data, and then use filters to choose from original noisy captions and pseudo captions to improve the quality of vision-language training data.
BEIT3~\cite{beit3} treats image data and language data the same way as discrete tokens. The image data is tokenized by the tokenizer of BEIT2~\cite{beit2}.
Image, text, and image-text pairs are randomly masked and the model is trained to recover the randomly masked tokens.
EMU2~\cite{emu2} operates in the continuous visual embedding space and jointly modeling the visual embeddings and text embeddings with a language decoder. The visual embeddings could later be decoded back to image pixels or video clips.

In the category of decoder-only VLMs,
Fuyu~\cite{fuyu} proposed to use an vision encoder-free architecture, that employs a transformer decoder-only model to process both image and text inputs.
The input image is first patchified and then linearly projected to a shared continuous embedding space as text tokens, so that the decoder-only model could process image and text tokens seamlessly.
CM3~\cite{cm3,cm3_scaling,cm3leon} and Chameleon~\cite{chameleon} proposed to convert images into discrete tokens and then model them together with language in the token space with a shared transformer decoder model.
Our work also shared promising early results of a Fuyu-style decoder-only setup following this line of work, by ablating the SigLIP vision encoder component from PaliGemma in Section~\ref{sec:abl:enc}.

\clearpage

\section{Tasks}
\label{sec:app:tasks}

We provide one training example for each transfer task, exactly as it reaches the model.
The image shown has been resized to $224 \times 224$ pixels in order to convey how much can be recognized at that resolution.

%

\subsection{ActivityNet-CAP: Captioning of short video snippets of activities}
\begin{tabular}{m{4cm}p{1.5cm}p{9cm}}
\multirow{4}{*}{\centering\begin{tabular}{c}
        \frame{\includegraphics[interpolate=false,width=1.7cm]{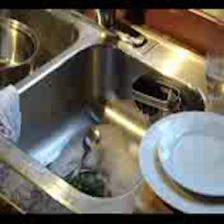}} \\
        \frame{\includegraphics[interpolate=false,width=1.7cm]{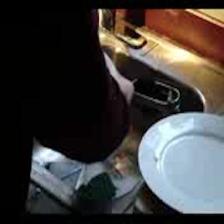}} \\
\end{tabular}}
& Prefix: & \texttt{"caption en\textbackslash n"} \\
& Suffix: & \texttt{"The man washes pots and a cutting board."} \\
& & \\
& Train: & $44775$ examples (train+val). \\
& Metric: & CIDEr (test split). \\
& Reference: & \citet{krishna2017activitynetcap}. \\
\end{tabular}
\vspace{1em}  

\subsection{ActivityNet-QA: Answering questions about short video snippets of activities}
\begin{tabular}{m{4cm}p{1.5cm}p{9cm}}
\multirow{4}{*}{\centering\begin{tabular}{c}
        \frame{\includegraphics[interpolate=false,width=1.7cm]{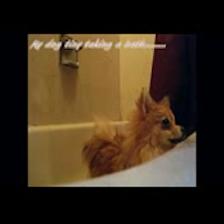}} \\
        \frame{\includegraphics[interpolate=false,width=1.7cm]{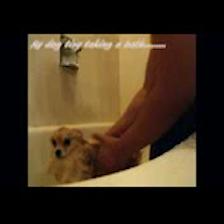}} \\
\end{tabular}}
& Prefix: & \texttt{"answer en what is the person in the video doing\textbackslash n"} \\
& Suffix: & \texttt{"bathe dog"} \\
& & \\
& Train: & $43130$ examples (train+val). \\
& Metric: & Exact match accuracy (test split). \\
& Reference: & \citet{yu2019activitynet}. \\
\end{tabular}
\vspace{1em}  

\subsection{AI2D: VQA on science diagrams}
\begin{tabular}{m{4cm}p{1.5cm}p{9cm}}
\multirow{4}{*}{\includegraphics[interpolate=false,width=3.5cm]{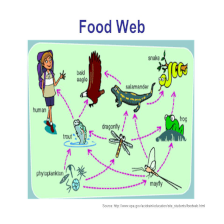}}

& Prefix: & \texttt{"answer en  Mayfly and dragonfly are eaten by ? choose from: Bald eagle \textbackslash t Frog \textbackslash t Phytoplankton \textbackslash t None of the above\textbackslash n"} \\
& Suffix: & \texttt{"Frog"} \\
& & \\
& Train: & $12413$ examples (train split). \\
& Metric: & Exact match accuracy (test split). \\
& Reference: & \citet{kembhavi2016eccv-AI2D} \\
\end{tabular}
\vspace{1em}  

\subsection{AOKVQA-DA: VQA using world knowledge (direct answer)}
\begin{tabular}{m{4cm}p{1.5cm}p{9cm}}
\multirow{4}{*}{\includegraphics[interpolate=false,width=3.5cm]{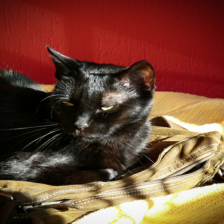}}
& Prefix: & \texttt{"answer en What is a sound this animal makes?\textbackslash n"} \\
& Suffix: & \texttt{"meow"} \\
& & \\
& Train: & $18201$ examples (train+val splits). \\
& Metric: & Exact match accuracy (test server). \\
& Reference: & \citet{AOKVQA} \\
\end{tabular}
\vspace{1em}  

\subsection{AOKVQA-MC: VQA using world knowledge (multiple choice)}
\begin{tabular}{m{4cm}p{1.5cm}p{9cm}}
\multirow{4}{*}{\includegraphics[interpolate=false,width=3.5cm]{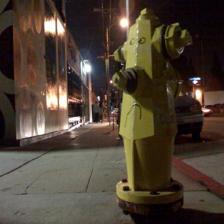}}
& Prefix: & \texttt{"answer en  What has the yellow object been drawn on to resemble? choose from: eagle \textbackslash t face \textbackslash t dog \textbackslash t star\textbackslash n"} \\
& Suffix: & \texttt{"face"} \\
& & \\
& Train: & $18201$ examples (train+val splits). \\
& Metric: & Accuracy (test server). \\
& Reference: & \citet{AOKVQA} \\
\end{tabular}
\vspace{1em}  

\subsection{ChartQA: VQA about charts with logical reasoning}
\begin{tabular}{m{4cm}p{1.5cm}p{9cm}}
\multirow{4}{*}{\includegraphics[interpolate=false,width=3.5cm]{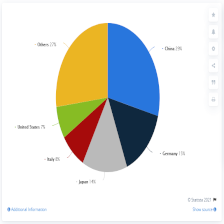}}
& Prefix: & \texttt{"What country accounted for 15 percent of the world\textquotesingle s machine tool production in 2020?\textbackslash n"} \\
& Suffix: & \texttt{"Germany"} \\
& & \\
& Train & $30219$ examples (human+augmented train and validation splits). \\
& Metric: & Relaxed match accuracy. ChartQA-human (human test split), ChartQA-aug (augmented test split). \\
& Reference: & \citet{masry-etal-2022-chartqa}
\end{tabular}
\vspace{1em}  

\subsection{COCOcap: COCO image captioning task}
\begin{tabular}{m{4cm}p{1.5cm}p{9cm}}
\multirow{5}{*}{\includegraphics[interpolate=false,width=3.5cm]{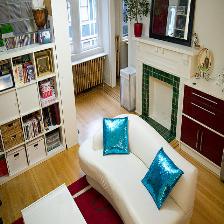}}
& Prefix: & \texttt{"caption en\textbackslash n"} \\
& Suffix: & \texttt{"A full view of a living room with pillows and books. "} \\
& & \\
& Train: & $113287$ images each with 5 captions (train+restval splits). \\
& Metric: & CIDEr score (val split). \\
& Reference: & \citet{coco2014} \\
\end{tabular}
\vspace{1em}  

\subsection{NoCaps: Novel object captioning}
\begin{tabular}{m{4cm}p{1.5cm}p{9cm}}
\multirow{5}{*}{\includegraphics[interpolate=false,width=3.5cm]{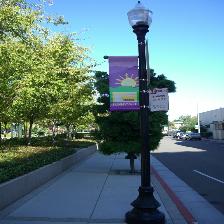}}
& Prefix: & \texttt{"caption en\textbackslash n"} \\
& Suffix: & \texttt{"up close shot of street light with banners and tree\textquotesingle s in the distance"} \\
& & \\
& Train: & Zero-shot evaluation of the model trained for COCOcap. \\
& Metric: & CIDEr score (val split). \\
& Reference: & \citet{agrawal2019nocaps} \\
\end{tabular}
\vspace{1em}  

\subsection{COCO-35L: COCO captions translated in 35 languages}
\begin{tabular}{m{4cm}p{1.5cm}p{9cm}}
\multirow{4}{*}{\includegraphics[interpolate=false,width=3.5cm]{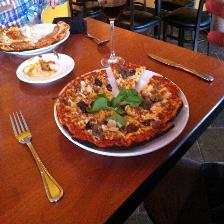}}
& Prefix: & \texttt{"caption no\textbackslash n"} \\
& Suffix: & \texttt{"pizza serveres p\textbackslash u00e5 bordet med gafler og kniver . ."} \\
& & \\
& Train: & $113287$ images each with 5 captions in each of the 35 languages (train split). \\
& Metric: & Mean of CIDEr on each of the 35 languages (dev split). \\
& Reference: & \citet{thapliyal-etal-2022-crossmodal-COCO35L-XM3600} \\
& Note: & The original raw data comes pre-tokenized with sometimes odd punctuation. \\
\end{tabular}
\vspace{1em}  

\subsection{XM3600: caption geographically-diverse images in 36 languages}
\begin{tabular}{m{4cm}p{1.5cm}p{9cm}}
\multirow{4}{*}{\includegraphics[interpolate=false,width=3.5cm]{figures/tasks/xm3600.jpg}}
& Prefix: & \texttt{"caption pt\textbackslash n"} \\
& Suffix: & \texttt{"Silhoueta de um corvo"} \\
& & \\
& Train: & Zero-shot evaluation of the model trained for COCO-35L. \\
& Metric: & Mean of CIDEr on each of the 36 languages (dev split). \\
& Reference: & \citet{thapliyal-etal-2022-crossmodal-COCO35L-XM3600} \\
\end{tabular}
\vspace{1em}  

\subsection{DocVQA: VQA on document images}
\begin{tabular}{m{4cm}p{1.5cm}p{9cm}}
\multirow{4}{*}{\includegraphics[interpolate=false,width=3.5cm]{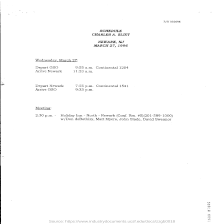}}
& Prefix: & \texttt{"What is the arrival time in GSO?\textbackslash n"} \\
& Suffix: & \texttt{"9:33 p.m."} \\
& & \\
& Train: & $44812$ examples (train+val splits) \\
& Metric: & ANLS (test server). \\
& Reference: & \citet{DBLP:journals/corr/abs-2007-00398-DOCVQA} \\
\end{tabular}
\vspace{1em}  

\subsection{GQA: VQA on image scene graphs}
\begin{tabular}{m{4cm}p{1.5cm}p{9cm}}
\multirow{4}{*}{\includegraphics[interpolate=false,width=3.5cm]{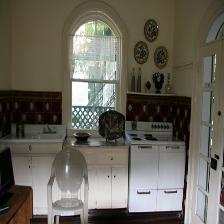}}
& Prefix: & \texttt{"answer en Where in the picture is the plate, in the bottom or in the top?\textbackslash n"} \\
& Suffix: & \texttt{"top"} \\
& & \\
& Train: & $1075062$ examples (train+val balanced splits). \\
& Metric: &  Exact-match accuracy (testdev balanced split). \\
& Reference: & \citet{DBLP:journals/corr/abs-2306-14610-GQA} \\
\end{tabular}
\vspace{1em}  

\subsection{xGQA: Cross-lingual VQA}
\begin{tabular}{m{4cm}p{1.5cm}p{9cm}}
\multirow{5}{*}{\includegraphics[interpolate=false,width=3.5cm]{figures/tasks/xgqa.jpg}}
& Prefix: & \texttt{"answer en Gibt es irgendwelche Schubladen oder Tische?\textbackslash n"} \\
& Suffix: & \texttt{"no"} \\
& & \\
& Train: & Zero-shot evaluation of the model trained for GQA. \\
& Metric: & Exact-match accuracy (test zero-shot splits). \\
& Reference: & \citet{pfeiffer-etal-2022-xgqa} \\
\end{tabular}
\vspace{1em}  

\subsection{InfoVQA: VQA on infographics}
\begin{tabular}{m{4cm}p{1.5cm}p{9cm}}
\multirow{4}{*}{\includegraphics[interpolate=false,width=3.5cm]{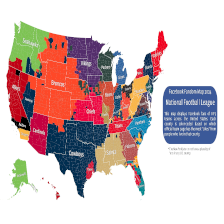}}
& Prefix: & \texttt{"answer en WHich team has most number of FB likes in Tennessee\textbackslash n"} \\
& Suffix: & \texttt{"titans"} \\
& & \\
& Train: & $26747$ examples (train + val). \\
& Metric: & ANLS (test server). \\
& Reference: & \citet{Mathew_2022-InfoVQA} \\
\end{tabular}
\vspace{1em}  

\subsection{MSRVTT-CAP: Open-domain short video captioning}
\begin{tabular}{m{4cm}p{1.5cm}p{9cm}}
\multirow{4}{*}{\centering\begin{tabular}{c}
        \frame{\includegraphics[interpolate=false,width=1.7cm]{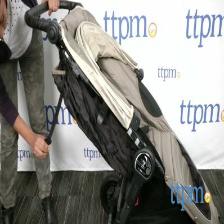}} \\
        \frame{\includegraphics[interpolate=false,width=1.7cm]{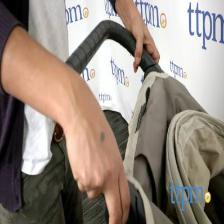}} \\
\end{tabular}}
& Prefix: & \texttt{"caption en\textbackslash n"} \\
& Suffix: & \texttt{"a woman is presenting a baby stroller"} \\
& & \\
& Train: & $4965$ examples (train+valid). \\
& Metric: & CIDEr (test split). \\
& Reference: & \citet{msrvtt} \\
\end{tabular}
\vspace{1em}  

\subsection{MSRVTT-QA: Open-domain sort video question answering}
\begin{tabular}{m{4cm}p{1.5cm}p{9cm}}
\multirow{4}{*}{\centering\begin{tabular}{c}
        \frame{\includegraphics[interpolate=false,width=1.7cm]{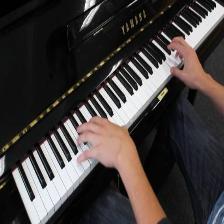}} \\
        \frame{\includegraphics[interpolate=false,width=1.7cm]{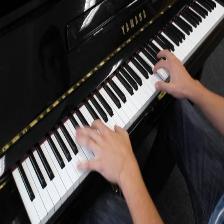}} \\
\end{tabular}}
& Prefix: & \texttt{"answer en what does a person play?\textbackslash n"} \\
& Suffix: & \texttt{"keyboard"} \\
& & \\
& Train: & $121646$ examples (train+valid splits). \\
& Metric: & Exact match accuracy (test split). \\
& Reference: & \citet{xu2017video} \\
\end{tabular}
\vspace{1em}  

\subsection{MSVD-QA: Answering questions about short video segments of events}
\begin{tabular}{m{4cm}p{1.5cm}p{9cm}}
\multirow{4}{*}{\centering\begin{tabular}{c}
        \frame{\includegraphics[interpolate=false,width=1.7cm]{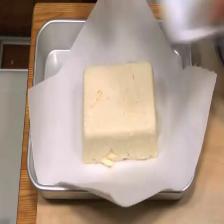}} \\
        \frame{\includegraphics[interpolate=false,width=1.7cm]{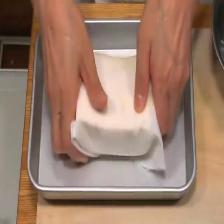}} \\
\end{tabular}}
& Prefix: & \texttt{"answer en what is a woman wrapping?\textbackslash n"} \\
& Suffix: & \texttt{"food"} \\
& & \\
& Train: & $29749$ examples (train+valid splits). \\
& Metric: & Exact match accuracy (test split). \\
& Reference: & \citet{xu2017video} based on \citet{msvd} \\
\end{tabular}
\vspace{1em}  

\subsection{NLVR2: Reasoning about natural language grounded in photographs}
\begin{tabular}{m{4cm}p{1.5cm}p{9cm}}
\multirow{4}{*}{\centering\begin{tabular}{c}
        \frame{\includegraphics[interpolate=false,width=1.7cm]{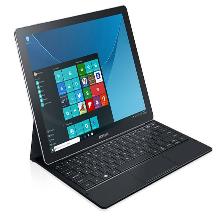}} \\
        \frame{\includegraphics[interpolate=false,width=1.7cm]{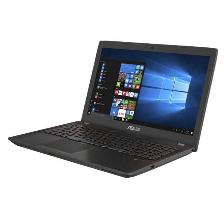}} \\
\end{tabular}}
& Prefix: & \texttt{"answer en Each image shows one opened laptop angled so the screen faces rightward.\textbackslash n"} \\
& Suffix: & \texttt{"False"} \\
& & \\
& Train: & $93355$ examples (train+dev splits). \\
& Metric: & Accuracy (test split). \\
& Reference: & \citet{suhr-etal-2019-corpus-nlvr2} \\
\end{tabular}

\subsection{MaRVL: Reasoning about multilingual language grounded in multicultural photographs}
\begin{tabular}{m{4cm}p{1.5cm}p{9cm}}
\multirow{4}{*}{\centering\begin{tabular}{c}
        \frame{\includegraphics[interpolate=false,width=1.7cm]{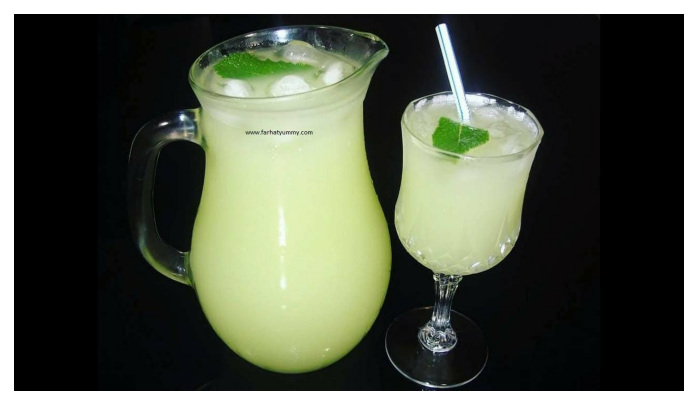}} \\
        \frame{\includegraphics[interpolate=false,width=1.7cm]{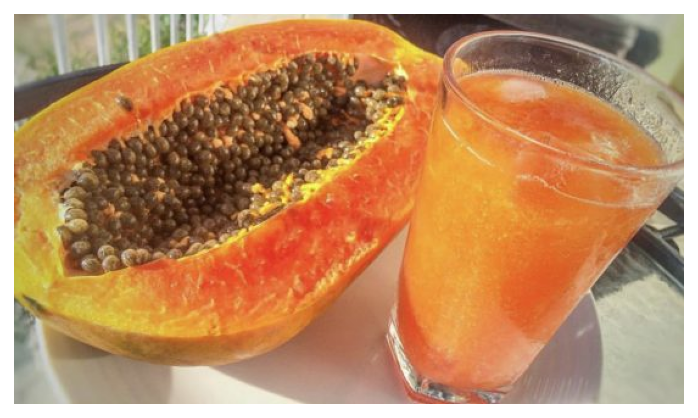}} \\
\end{tabular}}
& Prefix: & \texttt{"answer en Picha moja inaonyesha vyombo vyenye maji ya matunda pamoja na matunda pembeni na picha nyingine inaonyesha vyombo vyenye maji ya matunda peke yake.\textbackslash n"} \\
& Suffix: & \texttt{"True"} \\
& & \\
& Train: & Zero-shot evaluation of the model trained for NLVR2. \\
& Metric: & Accuracy (test splits). \\
& Reference: & \citet{liu-etal-2021-visually} \\
\end{tabular}

\subsection{OCR-VQA: VQA by reading text in images}
\begin{tabular}{m{4cm}p{1.5cm}p{9cm}}
\multirow{4}{*}{\includegraphics[interpolate=false,width=3.5cm]{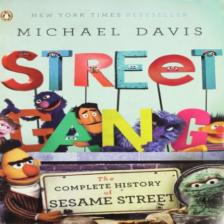}}
& Prefix: & \texttt{"answer en Is this a comedy book?\textbackslash n"} \\
& Suffix: & \texttt{"Yes"} \\
& & \\
& Train: & $901717$ examples (train+val splits). \\
& Metric: & Exact match accuracy (test split). \\
& Reference: & \citet{mishraICDAR19-ocrvqa} \\
\end{tabular}
\vspace{1em}  

\subsection{OKVQA: Outside knowledge VQA}
\begin{tabular}{m{4cm}p{1.5cm}p{9cm}}
\multirow{4}{*}{\includegraphics[interpolate=false,width=3.5cm]{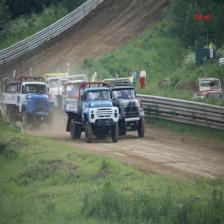}}
& Prefix: & \texttt{"answer en What type of race is this?\textbackslash n"} \\
& Suffix: & \texttt{"truck"} \\
& & \\
& Train: & $9009$ examples (train split). \\
& Metric: & Exact match accuracy (val split). \\
& Reference: & \citet{okvqa} \\
\end{tabular}
\vspace{1em}  

\subsection{RefCOCO\_seg: Referring expression segmentation}
\begin{tabular}{m{4cm}p{1.5cm}p{9cm}}
\multirow{4}{*}{\includegraphics[interpolate=false,width=3.5cm]{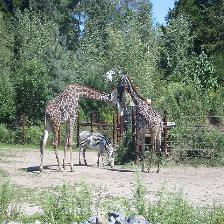}}
& Prefix: & \texttt{"the giraffe on the right standing tall\textbackslash n"} \\
& Suffix: & \texttt{"<loc0347><loc0553><loc0788><loc0749>\linebreak[1]<seg093>\linebreak[1]<seg106>\linebreak[1]<seg114>\linebreak[1]<seg078>\linebreak[1]<seg064>\linebreak[1]<seg012>\linebreak[1]<seg031>\linebreak[1]<seg055>\linebreak[1]<seg050>\linebreak[1]<seg012>\linebreak[1]<seg083>\linebreak[1]<seg118>\linebreak[1]<seg084>\linebreak[1]<seg078>\linebreak[1]<seg127>\linebreak[1]<seg121>"} \\
& & \\
& Train: & $24407$ examples: The combined training sets of RefCOCO, RefCOCO+, and RefCOCOg, \emph{but with all val and test images removed}, see LocCa~\cite{locca} for details. \\
& Metric: & mean intersection over union (mIoU) on all test splits. \\
& Reference: & RefCOCO and RefCOCO+: \citet{DBLP:conf/eccv/YuPYBB16-refcoco,kazemzadeh2014referit} and RefCOCOg: ~\citet{refcocog}. \\
\end{tabular}
\vspace{1em}  

\subsection{RSVQA-hr: VQA for remote sensing (high res)}
\begin{tabular}{m{4cm}p{1.5cm}p{9cm}}
\multirow{4}{*}{\includegraphics[interpolate=false,width=3.5cm]{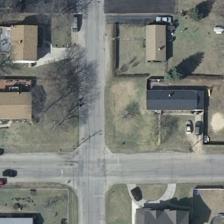}}
& Prefix: & \texttt{"answer en Are there more commercial buildings than roads?\textbackslash n"} \\
& Suffix: & \texttt{"no"} \\
& & \\
& Train: & $432239$ examples (non-numeric train+val splits). \\
& Metric: & Mean accuracy across questions types (reported on test and test2 splits). \\
& Reference: & \citet{Lobry_2020-RSVQA} \\
\end{tabular}
\vspace{1em}  

\subsection{RSVQA-lr: VQA for remote sensing (low res)}
\begin{tabular}{m{4cm}p{1.5cm}p{9cm}}
\multirow{4}{*}{\includegraphics[interpolate=false,width=3.5cm]{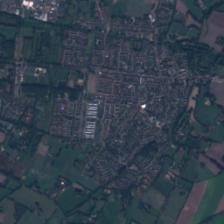}}
& Prefix: & \texttt{"answer en Is the number of buildings equal to the number of water areas in the image?\textbackslash n"} \\
& Suffix: & \texttt{"no"} \\
& & \\
& Train: & $47173$ (non-numeric train+val splits). \\
& Metric: & Average accuracy across questions types (non-numeric test split). \\
& Reference: & \citet{Lobry_2020-RSVQA} \\
\end{tabular}
\vspace{1em}  

\subsection{SciCap: Captions for scientific figures}
\begin{tabular}{m{4cm}p{1.5cm}p{9cm}}
\multirow{4}{*}{\includegraphics[interpolate=false,width=3.5cm]{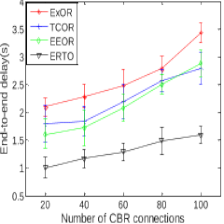}}
& Prefix: & \texttt{"caption en\textbackslash n"} \\
& Suffix: & \texttt{"end-to-end delay under different traffic load ."} \\
& & \\
& Train: & $120188$ examples (first sentence, no subfigure train+val splits). \\
& Metric: & CIDEr (first sentence, no subfigure test split). \\
& Reference: & \citet{hsu2021scicap} \\
\end{tabular}
\vspace{1em}  

\subsection{ScienceQA: Science question answering}
\begin{tabular}{m{4cm}p{1.5cm}p{9cm}}
\multirow{4}{*}{\includegraphics[interpolate=false,width=3.5cm]{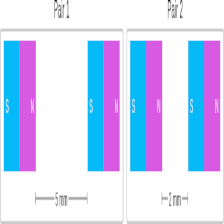}}
& Prefix: & \texttt{"Question: Think about the magnetic force between the magnets in each pair. Which of the following statements is true?\textbackslash nContext: The images below show two pairs of magnets. The magnets in different pairs do not affect each other. All the magnets shown are made of the same material.\textbackslash nOptions: (A) The magnetic force is weaker in Pair 1., (B) The magnetic force is weaker in Pair 2., (C) The strength of the magnetic force is the same in both pairs.\textbackslash nAnswer:\textbackslash n"} \\
& Suffix: & \texttt{"The answer is A."} \\
& & \\
& Train: & $8315$ examples (train+val splits). \\
& Metric: & Exact-match accuracy (test split). \\
& Reference: & \citet{lu2022learn-scienceqa} \\
\end{tabular}
\vspace{1em}  

\subsection{Screen2Words: Mobile UI summarization}
\begin{tabular}{m{4cm}p{1.5cm}p{9cm}}
\multirow{4}{*}{\includegraphics[interpolate=false,width=3.5cm]{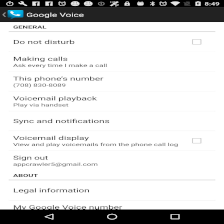}}
& Prefix: & \texttt{"caption en\textbackslash n"} \\
& Suffix: & \texttt{"screen showing general settings page"} \\
& & \\
& Train: & $18107$ (train+dev splits). \\
& Metric: & CIDEr (test split). \\
& Reference: & \citet{wang2021screen2words} \\
\end{tabular}
\vspace{1em}  

\subsection{ST-VQA: Scene text VQA}
\begin{tabular}{m{4cm}p{1.5cm}p{9cm}}
\multirow{4}{*}{\includegraphics[interpolate=false,width=3.5cm]{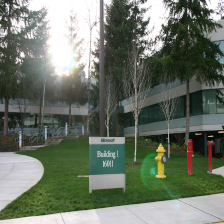}}
& Prefix: & \texttt{"answer en What company is this?\textbackslash n"} \\
& Suffix: & \texttt{"microsoft"} \\
& & \\
& Train: & $26074$ examples (train+val). \\
& Metric: & ANLS (test server). \\
& Reference: & \citet{Biten_2019-STVQA} \\
\end{tabular}
\vspace{1em}  

\subsection{TallyQA: Complex counting questions}
\begin{tabular}{m{4cm}p{1.5cm}p{9cm}}
\multirow{4}{*}{\includegraphics[interpolate=false,width=3.5cm]{figures/tasks/tallyqa.jpg}}
& Prefix: & \texttt{"answer en How many bowls are in the picture?\textbackslash n"} \\
& Suffix: & \texttt{"1"} \\
& & \\
& Train: & $249318$ examples (train split). \\
& Metric: & Accuracy (test split) reported on simple and complex splits. \\
& Reference: & \citet{acharya2019tallyqa} \\
\end{tabular}
\vspace{1em}  

\subsection{CountBenchQA: Evaluate counting in a structured, controlled way}
\begin{tabular}{m{4cm}p{1.5cm}p{9cm}}
\multirow{4}{*}{\includegraphics[interpolate=false,width=3.5cm]{figures/tasks/countbenchqa.jpg}}
& Prefix: & \texttt{"answer en How many sculptures are there in the image?\textbackslash n"} \\
& Suffix: & \texttt{"7"} \\
& & \\
& Train: & Zero-shot evaluation of the model trained for TallyQA. \\
& Metric: & Exact match (full test split). \\
& Reference: & Introduced in this report, based on \citet{paiss2023countclip} \\
\end{tabular}
\vspace{1em}  

\subsection{TextCaps: Image captioning with reading comprehension}
\begin{tabular}{m{4cm}p{1.5cm}p{9cm}}
\multirow{4}{*}{\includegraphics[interpolate=false,width=3.5cm]{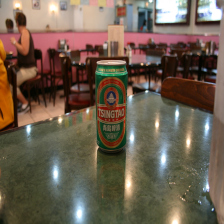}}
& Prefix: & \texttt{"caption en\textbackslash n"} \\
& Suffix: & \texttt{"green and red beer can for tsingtao inside a restaurant."} \\
& & \\
& Train: & $21953$ examples (train split). \\
& Metric: & CIDEr (test split). \\
& Reference: & \citet{sidorov2019textcaps} \\
\end{tabular}
\vspace{1em}  

\subsection{TextVQA: Visual reasoning based on text in images}
\begin{tabular}{m{4cm}p{1.5cm}p{9cm}}
\multirow{4}{*}{\includegraphics[interpolate=false,width=3.5cm,height=3.5cm]{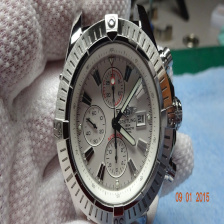}}
& Prefix: & \texttt{"answer en what brand of watch is this?\textbackslash n"} \\
& Suffix: & \texttt{"breitling"} \\
& & \\
& Train: & $39602$ (train+val splits). \\
& Metric: & test server (test-std). \\
& Reference: & \citet{singh2019towards-textvqa} \\
\end{tabular}
\vspace{1em}  

\subsection{VATEX-CAP: Broad video captioning}
\begin{tabular}{m{4cm}p{1.5cm}p{9cm}}
\multirow{4}{*}{\centering\begin{tabular}{c}
        \frame{\includegraphics[interpolate=false,width=1.7cm]{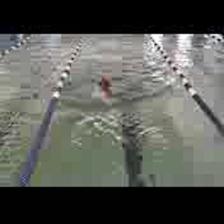}} \\
        \frame{\includegraphics[interpolate=false,width=1.7cm]{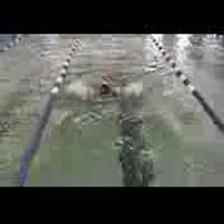}} \\
\end{tabular}}
& Prefix: & \texttt{"caption en\textbackslash n"} \\
& Suffix: & \texttt{"A person swims in a pool and touches the wall while others cheer."} \\
& & \\
& Train: & $24850$ examples (train + valid splits). \\
& Metric: & CIDEr (test split). \\
& Reference: & \citet{wang2019vatex}. \\
\end{tabular}
\vspace{1em}  

\subsection{VizWizVQA: VQA from people who are blind}
\begin{tabular}{m{4cm}p{1.5cm}p{9cm}}
\multirow{4}{*}{\includegraphics[interpolate=false,width=3.5cm]{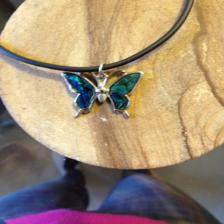}}
& Prefix: & \texttt{"answer en What colors are in the charm of this necklace?\textbackslash n"} \\
& Suffix: & \texttt{"silver green blue"} \\
& & \\
& Train: & $24842$ examples (train+val). \\
& Metric: & test server (test-std). \\
& Reference: & \citet{gurari2018vizwiz} \\
\end{tabular}
\vspace{1em}  

\subsection{VQAV2: Visual Question Answering}
\begin{tabular}{m{4cm}p{1.5cm}p{9cm}}
\multirow{4}{*}{\includegraphics[interpolate=false,width=3.5cm]{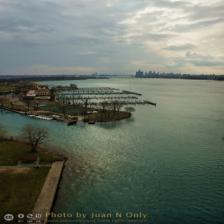}}
& Prefix: & \texttt{"answer en Does this look like a place you\textquotesingle d love to swim?\textbackslash n"} \\
& Suffix: & \texttt{"yes"} \\
& & \\
& Train: & $658111$ examples (train+validation) \\
& Metric: & \texttt{VQAV2}: test server (test-std). \\
& & \texttt{VQAV2 (minival)}: computed on local split of 10k examples of the validation set. \\
& Reference: & \citet{balanced_vqa_v2} \\
\end{tabular}
\vspace{1em}  

\subsection{MMVP: Hard questions about CLIP-blind image pairs}
\begin{tabular}{m{4cm}p{1.5cm}p{9cm}}
\multirow{4}{*}{\includegraphics[interpolate=false,width=3.5cm]{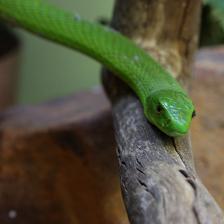}}
& Prefix: & \texttt{"answer en From which angle is this image taken?\textbackslash n"} \\
& Suffix: & \texttt{"Front"} \\
& & \\
& Train: & Zero-shot evaluation of the model trained for VQAv2. \\
& Metric: & Paired accuracy. \\
& Reference: & \citet{tong2024eyes} \\
\end{tabular}
\vspace{1em}  

\subsection{POPE: Object presence as yes/no VQA (object hallucination)}
\begin{tabular}{m{4cm}p{1.5cm}p{9cm}}
\multirow{4}{*}{\includegraphics[interpolate=false,width=3.5cm]{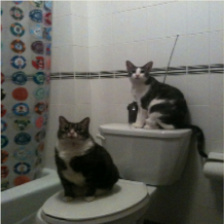}}
& Prefix: & \texttt{"answer en Is there a car in the image?\textbackslash n"} \\
& Suffix: & \texttt{"no"} \\
& & \\
& Train: & Zero-shot evaluation of the model trained for VQAv2. \\
& Metric: & Exact match with ``yes'' or ``no''. The overall score is the average of accuracies on the {\tt random}, {\tt popular}, and {\tt adversarial} splits, which all have the same size. \\
& Reference: & \citet{Li23pope}. \\
& Note: & Car, not cat. \\
\end{tabular}
\vspace{1em}  

\subsection{Objaverse Multiview: view-consistent 3D object annotation}
\begin{tabular}{m{4cm}p{1.5cm}p{9cm}}
\multirow{4}{*}{\includegraphics[interpolate=false,width=3.5cm]{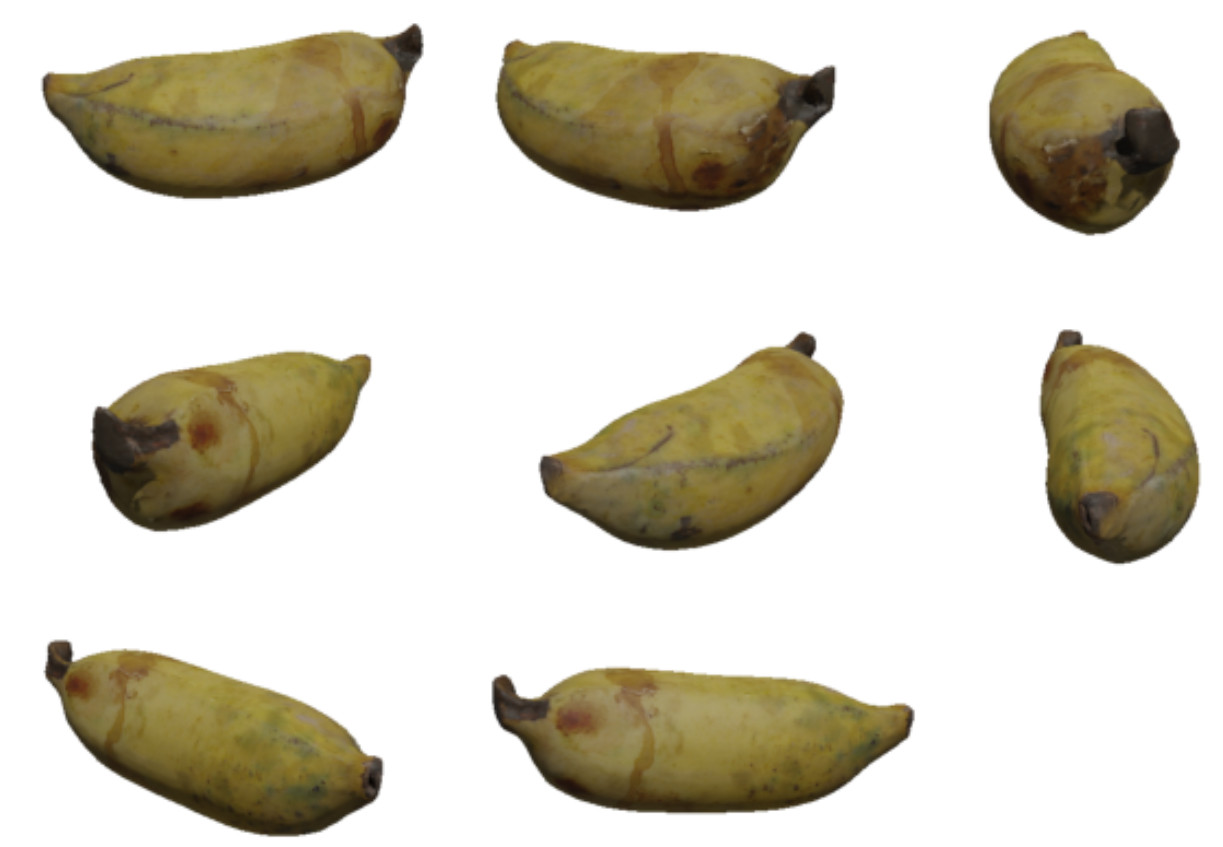}}
& Prefix: & \texttt{"answer en Describe the object in the image?\textbackslash n"} \\
& Suffix: & \texttt{"banana"} \\
& & \\
& Train: & Zero-shot evaluation of the model trained for VQAv2. \\
& Metric: & Cosine similarity according to Universal Sentence Encoder v4. \\
& Reference: & \citet{kabra2024objaverse} \\
& Note: & Each image is seen individually, the final prediction is the score-weighted average. 44k objects, 1100 categories, 8 views rendered per object, 4 prompts per object view. \\
\end{tabular}
\vspace{1em}  

\subsection{WidgetCap: Descriptions for Mobile User Interface Elements}
\begin{tabular}{m{4cm}p{1.5cm}p{9cm}}
\multirow{4}{*}{\includegraphics[interpolate=false,width=3.5cm]{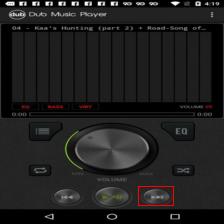}}
& Prefix: & \texttt{"caption en\textbackslash n"} \\
& Suffix: & \texttt{"fast forward"} \\
& & \\
& Train: & $44704$ examples (train+dev splits). We annotate the widget to be captioned by overlaying a red bounding box in the image input. \\
& Metric: & CIDEr (test split). \\
& Reference: & \citet{Li2020WidgetCap} \\
\end{tabular}
\vspace{1em}  

\clearpage

\section{Image augmentations for RefCOCO}\label{sec:app:refcoco}

\begin{figure}
\centering
\begin{minipage}{.48\textwidth}
  \centering
  \includegraphics[width=\linewidth]{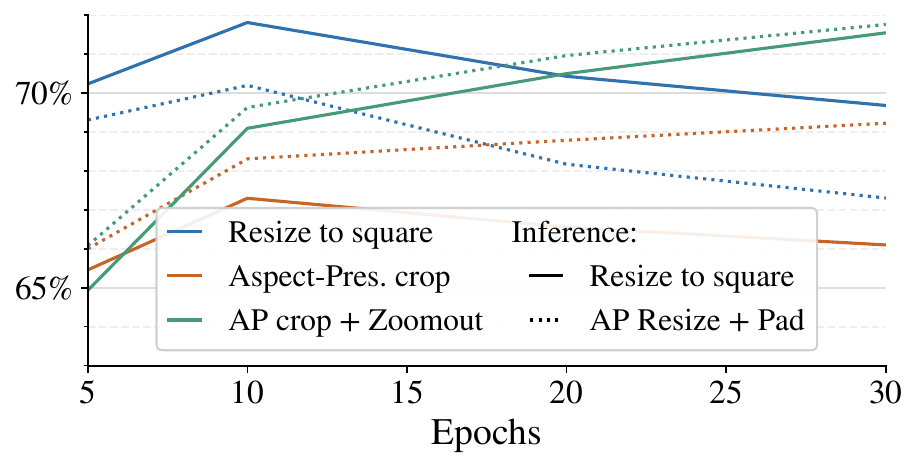}
  \captionof{figure}{Three different ways to get $448\times 448$ images during training: plain resizing, random aspect-ratio preserving crop, or the latter with an additional random ``zoom-out'' augmentation.
  Two options at inference-time: plain resizing or performing an aspect-ratio preserving resize with padding.
  Lines are not training curves, but multiple training runs of 5, 10, 20, and 30 epochs.}
  \label{fig:refcoco_squeeze}
  \vspace{1em}
\end{minipage}%
\hfill
\begin{minipage}{.48\textwidth}
  \centering
  \includegraphics[width=\linewidth]{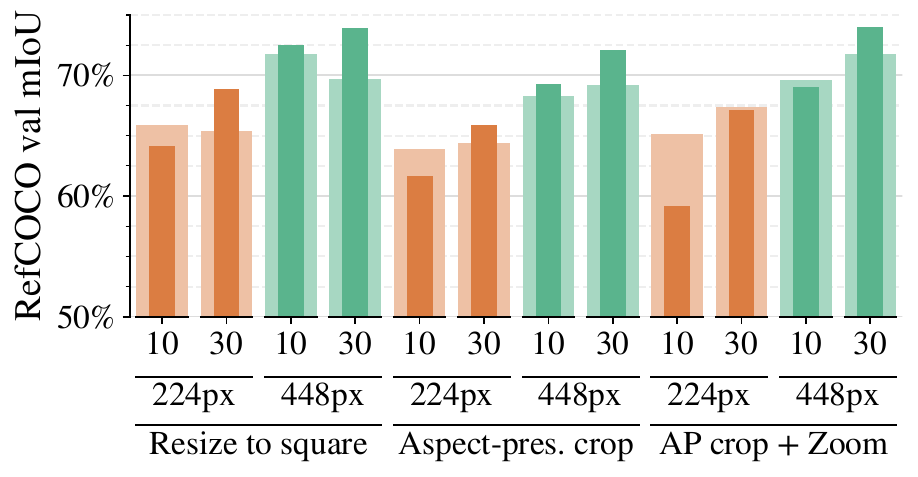}
  \captionof{figure}{Wide, outer bars show performance without label-smoothing and dropout (LSDO), thin inner bars with.
  10 and 30 denote epochs.
  First, LSDO always hurts the 224\,px 10\,ep setting but always helps the 448\,px 30\,ep setting.
  Second, resize to square overfits in longer training, but adding LSDO remediates this and allows such simple method to outperform more complicated image-augmentation based ones.
  }
  \label{fig:refcoco_lsdo}
\end{minipage}
\end{figure}

The current wisdom regarding structured output tasks such as detection and segmentation is that image augmentations are crucial, and especially keeping the original aspect ratio while performing zoom-like augmentations is key.

We found this \emph{not} to be the case, at least in the setting of pretrained VLMs.
After extensive experiments, we found that simply resizing the image to a square $224\times224$ pixels achieves best (and state-of-the-art) performance.

Figure~\ref{fig:refcoco_squeeze} shows that one needs to explore both training-time resizing technique and inference-time strategy simultaneously.
It seems natural that, when performing aspect-ratio preserving resize with padding at inference time, the training augmentations need to reflect that to some degree.
However, one can also simply resize test images to a square at inference time, and then resize the output logits back to the original resolution for evaluation.
Doing so works very well in tandem with simple resizing to square at training time and, as one might expect, does not work well with aspect-preserving resizes during training.

Besides this training-inference mismatch, Figure~\ref{fig:refcoco_squeeze} also shows that the plain resize suffers from overfitting when training for longer than 10 epochs, while cropping and zooming do not, as they also act as image augmentations.
However, overfitting can also be combatted via other, simpler techniques than image processing.

Figure~\ref{fig:refcoco_lsdo} shows that adding label-smoothing and dropout to the training not only completely eliminates the overfitting, but also allows the simple resize strategy to achieve the same or better performance than the more complicated image augmentation ones.
Hence, for our final setting that reaches state-of-the-art results, we stick to simple image resizing with increased training schedule, label-smoothing, and dropout.

\section{Introducing CountBenchQA}\label{sec:app:countbenchqa}


TallyQA is the only dataset commonly used to evaluate counting, an important skill for VLMs.
However, we found that it has two issues~\cite{kaji2024evaluating}:
First, the ground-truth label distribution is highly skewed towards small counts (0, 1, 2).
Second, while the dataset is large, the ground-truth labels themselves are rather noisy, and the validation and test splits have not undergone any further verification or clean-up.

The recently introduced CountBench~\cite{paiss2023countclip} dataset of 540 images fixes both these shortcomings.
Images and captions are taken from the LAION-400M image-text dataset and each caption mentions the main object class present in the image and specifies its count.
The correspondence of images, counts and captions has been manually verified.

There are 540 images in total with 60 images per count in [2, 3, ... 10].
There is no 0 or 1 count, the smallest count is 2.
The data is not originally available in a VQA format but just as (image, caption, count) triplets.
To construct CountbenchQA, we manually annotated the dataset with questions of the form {\tt How many \{object class\} are there in the image?}, where we extracted the {\tt \{object class\}} from the caption manually. 
where the object class was overly specific or difficult to understand we replaced them with simpler versions (\eg/ we changed ``glass star christmas tree decorations'' to "stars", and ``founders of Kappa Sigma'' to ``people''.)
There are only 3 exceptions deviating from the above template (these were needed to ensure correctness of the answer in ambiguous cases) and these are:%
\begin{itemize}[topsep=0pt]
  \item How many people are in the foreground of this image?
  \item How many stories does this cottage have?
  \item How many petals does each flower have in this image?
\end{itemize}%
For some counts, several images look extremely similar.
For 9 items, several images are button/icon symbols with slightly different colours/symbols on them, arranged in a regular grid.

Out of the 540 images, several are not available anymore; the parquet file provided at \url{https://huggingface.co/datasets/nielsr/countbench} contains entries for all 540 examples but only image data for 490 of those examples.
This is why we have two splits, although we computed our metrics using the full split and have not recomputed them since.
Our dataset is available at \url{https://github.com/google-research/big_vision/blob/main/big_vision/datasets/countbenchqa/}.

\clearpage
\section{Issues with published WidgetCaps numbers}\label{sec:app:widgetcaps_eval}

We have found that at least two prior works reporting results on WidgetCaps do so wrongly.
First, previous PaLI numbers (PaLI-X: 153.0, PaLI-3: 159.8) are reported on the validation set while implying they are test-set results.
This makes comparison with test-set numbers such as our 148.4 invalid.
However, a re-run of our transfers in a comparable setting (training on {\tt train}, evaluating on {\tt dev}) achieves a validation set CIDEr of 140.2 at $224\,$px and 155.2 at $448\,$px resolution, placing it firmly between PaLI-X and PaLI-3 at much smaller model size and resolution.

Furthermore, ScreenAI~\cite{screenai} had a mistake in the CIDEr computation for WidgetCap, using only a single out of the five provided ground-truth captions.
Re-computing their CIDEr score using all five ground-truth captions changes the score from the 167 originally reported in the paper to 156.4, which is close to PaliGemma's 148.4.

\section{Image annotations work as well as prompts}\label{sec:app:widgetcaps_box}

The WidgetCaps requires captioning a specific widget in the image.
The widget is given by bounding-box coordinates, which are typically provided in the prompt, including in previous PaLI versions.

Here, we found that drawing a red box in the image during transfers, and not indicating it in the prompt, performs equally well.
For example, the best validation result when sweeping hyper-parameters in both settings equally, was 135.99 CIDEr for the drawn red box, and 135.28 for the box coordinates as {\tt <loc>} tokens in the prompt.
This difference is well within re-run variation.

\clearpage
\section{More details and results with Objaverse}\label{sec:app:objaverse}


\begin{figure*}[t]
  \centering
  \includegraphics[clip, trim=7cm 6.5cm 6cm 0cm, width=\textwidth]{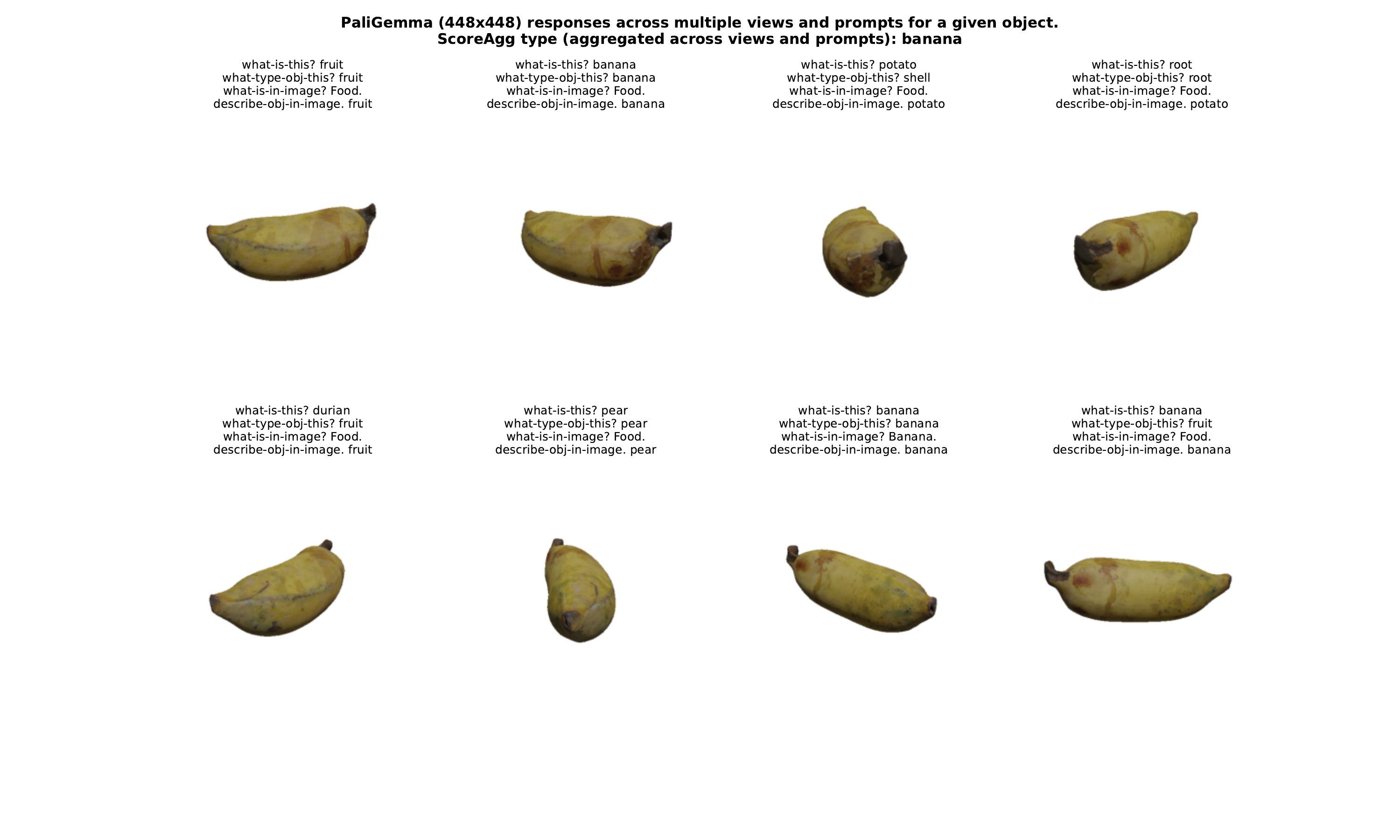}
  \caption{One representative example of PaliGemma's predictions across the different views.}
  \label{fig:app:banana}
\end{figure*}

\begin{table}[t]
\centering
\caption{Results of various Objaverse captioning methods. The mean, standard deviation, and standard error of the mean are computed across the 44\,k examples, after the per-example aggregation.}
\label{tbl:objaverse}
\vspace{-2mm}

\begin{tabular}{p{8cm}ccc}
\toprule
\textbf{Model} & \textbf{Score} (mean) & (stddev) & (stderr) \\
\toprule

CAP3D (Luo et al., 2023) & 36.6 & - & - \\
PaLI-X (55B, 756x756, mixed VQA transfer) & 59.0 & 29.1 & 0.1 \\
PaLI-3 (5B, 448x448, mixed VQA transfer) & \textbf{64.3} & 29.2 & 0.1 \\
\midrule
PaliGemma (3B, 224x224) & 58.4 & 29.5 & 0.1 \\
PaliGemma (3B, 224x224, VQAv2 transfer) & 62.7 & 28.2 & 0.1 \\
PaliGemma (3B, 448x448) & 60.4 & 28.9 & 0.1 \\
PaliGemma (3B, 448x448, VQAv2 transfer) & \textbf{62.8} & 28.3 & 0.1 \\

\bottomrule
\end{tabular}
\end{table}

Objaverse~\cite{deitke2023objaverse} is an internet-scale collection of 800\,k diverse but noisily or unannotated 3D models.
They were uploaded by 100\,k artists to the Sketchfab platform.
A subset of 44\,k objects called Objaverse-LVIS is accompanied by human-verified categories.
These can be used to validate object class predictions.
This amounts to only 5\% of examples with human-verified labels in a restricted subset of categories.

A compelling use of VLMs is to infer the type of each Objaverse object in an open-vocabulary setting, providing multiple views of each object to the model.
Furthermore, this is an interesting evaluation benchmark for VLMs: the images come from a more specialized distribution of 3D renders (often untextured), and the query of object type is something we should expect a VLM to handle well.

The previously most popular method, CAP3D~\cite{luo2023cap3d}, is an ad-hoc combination of BLIP2, CLIP, and GPT4, where GPT4 summarizes annotations created by BLIP2 and CLIP.
Given access to log-probabilities, \cite{kabra2024objaverse} show that the aggregating the predictions of each view informed by their log-probabilities ({\tt ScoreAgg}), largely outperforms CAP3D while being significantly simpler.
See \cite{kabra2024objaverse} for more details on the exact setup.

We additionally evaluate PaLI-3, and the open-weight PaliGemma base and VQAv2 fine-tune.
We use four VQA prompts to probe for the type of each object: (i) ``What is this?'' (ii) ``What type of object is this?'' (iii) ``What is in the image?'' (iv) ``Describe the object in the image''. This produces four sets of five sampled responses per view, which are aggregated using ScoreAgg to produce a final prediction.
The results are summarized in Table~\ref{tbl:objaverse} and show that PaliGemma has clear 3D object understanding out of the box, and the VQAv2 fine-tune performs even better.
Finally, we also show a representative example of raw predictions in Figure~\ref{fig:app:banana}; a banana for s(c)ale, if you will. 

\clearpage

\section{Multitask transfer}\label{sec:abl:multi}
\newcommand{\taskvalM}[2]{#1 {\footnotesize (#2\%)}}
\begin{table}[t]
\centering
\caption{Results when multitasking. Table~1 is the best achievable single-task result with per-task tuned hyper-paramters. ``Single Simple'' is then the per-task performance when using the same simplified hyper-parameter setting for all tasks. Finally, ``Multi'' is the multitasking setup where a single model performs all tasks, either with or without prefix indicating the task. The number in parenthesis is the relative regret versus the preceding column.}
\label{tbl:multitask}
\vspace{-2mm}

\begin{tabular}{m{4cm}cccc}
\toprule
\textbf{Metric} & \textbf{Table~1} & \textbf{Single Simple} & \textbf{Multi Prefix} & \textbf{Multi No Prefix} \\
\midrule


AI2D & \phantom{0}72.1 & \phantom{0}73.1 & \taskvalM{\phantom{0}72.3}{-1.0} & \taskvalM{\phantom{0}72.3}{-1.1} \\
COCOcap & 141.9 & 141.7 & \taskvalM{139.5}{-1.6} & \taskvalM{138.9}{-2.0} \\
NoCaps & 121.7 & 121.6 & \taskvalM{118.4}{-2.6} & \taskvalM{\phantom{0}98.9}{-18.7} \\
DocVQA (val) & \phantom{0}37.8 & \phantom{0}38.6 & \taskvalM{\phantom{0}32.2}{-16.7} & \taskvalM{\phantom{0}31.3}{-19.0} \\
GQA & \phantom{0}65.6 & \phantom{0}65.4 & \taskvalM{\phantom{0}64.9}{-0.7} & \taskvalM{\phantom{0}64.2}{-1.8} \\
xGQA (avg7) & \phantom{0}57.3 & \phantom{0}56.9 & \taskvalM{\phantom{0}55.8}{-2.0} & \taskvalM{\phantom{0}52.6}{-7.7} \\
InfoVQA (val) & \phantom{0}25.5 & \phantom{0}25.0 & \taskvalM{\phantom{0}20.8}{-17.0} & \taskvalM{\phantom{0}21.4}{-14.4} \\
OCR-VQA & \phantom{0}72.3 & \phantom{0}73.3 & \taskvalM{\phantom{0}71.4}{-2.6} & \taskvalM{\phantom{0}71.4}{-2.6} \\
OKVQA & \phantom{0}63.5 & \phantom{0}63.1 & \taskvalM{\phantom{0}67.0}{+6.2} & \taskvalM{\phantom{0}58.9}{-6.6} \\
RefCOCO (val) & \phantom{0}73.4 & \phantom{0}69.9 & \taskvalM{\phantom{0}68.4}{-2.1} & \taskvalM{\phantom{0}68.3}{-2.3} \\
RefCOCO+ (val) & \phantom{0}68.3 & \phantom{0}61.0 & \taskvalM{\phantom{0}61.2}{+0.3} & \taskvalM{\phantom{0}60.9}{-0.2} \\
RefCOCOg (val) & \phantom{0}67.7 & \phantom{0}63.7 & \taskvalM{\phantom{0}61.8}{-3.1} & \taskvalM{\phantom{0}61.8}{-3.0} \\
RSVQA-hr (test) & \phantom{0}92.6 & \phantom{0}92.7 & \taskvalM{\phantom{0}92.1}{-0.6} & \taskvalM{\phantom{0}92.2}{-0.5} \\
RSVQA-hr (test2) & \phantom{0}90.6 & \phantom{0}90.8 & \taskvalM{\phantom{0}90.0}{-0.9} & \taskvalM{\phantom{0}90.2}{-0.7} \\
RSVQA-lr & \phantom{0}92.6 & \phantom{0}93.7 & \taskvalM{\phantom{0}93.4}{-0.3} & \taskvalM{\phantom{0}93.0}{-0.8} \\
SciCap & 162.3 & \phantom{0}64.0 & \taskvalM{\phantom{0}74.7}{+16.7} & \taskvalM{\phantom{0}74.7}{+16.6} \\
Screen2Words & 117.6 & 114.7 & \taskvalM{111.6}{-2.7} & \taskvalM{110.6}{-3.5} \\
ST-VQA (val) & \phantom{0}61.6 & \phantom{0}62.0 & \taskvalM{\phantom{0}58.3}{-6.0} & \taskvalM{\phantom{0}57.8}{-6.8} \\
TallyQA (simple) & \phantom{0}81.7 & \phantom{0}81.3 & \taskvalM{\phantom{0}80.4}{-1.1} & \taskvalM{\phantom{0}80.7}{-0.7} \\
TallyQA (complex) & \phantom{0}69.6 & \phantom{0}69.6 & \taskvalM{\phantom{0}69.6}{-0.1} & \taskvalM{\phantom{0}69.9}{+0.3} \\
CountBenchQA & \phantom{0}81.9 & \phantom{0}83.5 & \taskvalM{\phantom{0}80.6}{-3.5} & \taskvalM{\phantom{0}78.3}{-6.2} \\
TextCaps & 127.5 & 128.3 & \taskvalM{122.5}{-4.5} & \taskvalM{122.3}{-4.6} \\
TextVQA (val) & \phantom{0}59.0 & \phantom{0}57.9 & \taskvalM{\phantom{0}56.0}{-3.2} & \taskvalM{\phantom{0}56.5}{-2.5} \\
VQAv2 (minival) & \phantom{0}82.1 & \phantom{0}81.9 & \taskvalM{\phantom{0}83.4}{+1.8} & \taskvalM{\phantom{0}82.9}{+1.2} \\
VizWizVQA (val) & \phantom{0}76.4 & \phantom{0}75.9 & \taskvalM{\phantom{0}75.7}{-0.2} & \taskvalM{\phantom{0}76.0}{+0.2} \\
WidgetCap & 136.1 & 136.4 & \taskvalM{129.1}{-5.3} & \taskvalM{131.5}{-3.6} \\
\midrule
Average & \phantom{0}84.6 & \phantom{0}80.2 & \taskvalM{\phantom{0}78.9}{-2.0} & \taskvalM{\phantom{0}77.6}{-3.5} \\


\bottomrule
\end{tabular}

\end{table}

Transferring to individual tasks by fine-tuning is good when all one cares about is having a model that solves a specific task.
However, it is often desirable to have a single generalist model with a conversational interface.
This is typically achieved by instruction tuning, \ie/ fine-tuning on a mixture of a diverse dataset.
We verify that PaliGemma is well-suited for this type of transfer: we transfer it to a mix of 27 of our datasets (excluding some tasks, such as multi-image, for simplicity).
We follow the ``simplified'' hyper-parameter from Section~\ref{sec:abl:simple} for the mixture, and mix uniformly such that each individual task goes through 10 epochs, no matter its size.

The results in Table~\ref{tbl:multitask} show that there is a dramatic change in a few tasks, while most win or lose just a little.
Surprisingly, the overall average shows the largest loss comes from changing to a unified hyper-parameter, followed by multitasking, and removing per-task prefix indicator comes with the smallest regret.
However, it should be noted that this multitasking setup was not tuned much.

\clearpage
\section{Inference}\label{sec:inference}

See table~\ref{tbl:inference} for inference measurements using the code from {\tt big\_vision}.

\begin{table}
\centering

\newcommand{\p}{\phantom{0}}

\caption{
Inference measurements on a TPUv3~\cite{tpucloud} with 8 devices (4 chips), prefilling 512 tokens, cache sized 640 tokens.
Even though FSDP sharding~\cite{fsdp} is very efficient at training time (Section~\ref{sec:pretrain_details}),
at inference time it has the downside that each device needs process at least one single example and needs to read all model parameters. It reduces the memory requirements but it still processes one single example as slow as using a single device with a larger memory.
Megatron-style sharding~\cite{megatron} on the other hand shards both the parameters \textit{and} the activations, and can make use of multiple accelerators in parallel to process a single example and significantly reduce the total number of memory reads per device.
}
\label{tbl:inference}

\small
\begin{tabular}{m{1.2cm}m{1.4cm}m{1cm}m{1cm}m{1cm}m{1.2cm}m{1cm}m{1cm}m{1cm}m{1cm}m{1cm}}
\toprule

\multirow[b]{2}{*}{\textbf{Sharding}} &
\multirow[b]{2}{*}{\textbf{Params}} &
\multirow[b]{2}{*}{\textbf{Batch}} &
\multicolumn{2}{c}{\textbf{Walltime [ms]}} &
\multicolumn{2}{c}{\textbf{Tokens/sec}} &
\multicolumn{2}{c}{\textbf{Utilization [\%]}} &
\multicolumn{2}{c}{\textbf{Memory [GiB]}} \\
& & & Prefill & Extend & Prefill & Extend & Prefill & Extend & Prefill & Extend \\


\midrule
FSDP       & { \tt float32    } & {\p}{\p}1 &    {\p}107 &    {\p}14.6 & {\p}4785 & {\p}{\p}{\p}68 &    34.9 & {\p}0.6 & {\p}2.04 & {\p}1.69 \\
           & { \tt            } & {\p}{\p}8 &    {\p}107 &    {\p}14.7 &    38282 &    {\p}{\p}545 &    34.9 & {\p}0.6 & {\p}2.04 & {\p}1.69 \\
           & { \tt            } &    {\p}64 &    {\p}772 &    {\p}14.7 &    42423 &       {\p}4367 &    38.6 & {\p}4.5 & {\p}2.67 & {\p}1.87 \\
           & { \tt            } &       512 &       6262 &    {\p}35.0 &    41866 &          14617 &    38.1 &    14.8 & {\p}7.48 & {\p}3.21 \\
           & { \tt bfloat16   } & {\p}{\p}1 &    {\p}101 & {\p}{\p}8.2 & {\p}5073 &    {\p}{\p}122 &    37.0 & {\p}1.0 & {\p}0.84 & {\p}0.64 \\
           & { \tt            } & {\p}{\p}8 &    {\p}101 & {\p}{\p}8.3 &    40555 &    {\p}{\p}969 &    36.9 & {\p}1.0 & {\p}0.84 & {\p}0.64 \\
           & { \tt            } &    {\p}64 &    {\p}764 &    {\p}10.2 &    42914 &       {\p}6258 &    39.1 & {\p}6.4 & {\p}1.48 & {\p}0.81 \\
           & { \tt            } &       512 &       6235 &    {\p}30.8 &    42043 &          16601 &    38.3 &    16.8 & {\p}6.28 & {\p}2.14 \\
\midrule
Megatron   & { \tt float32    } & {\p}{\p}1 & {\p}{\p}22 & {\p}{\p}8.2 &    23018 &    {\p}{\p}122 &    20.6 & {\p}0.1 & {\p}2.23 & {\p}1.61 \\
           & { \tt            } & {\p}{\p}8 &    {\p}105 &    {\p}10.3 &    39149 &    {\p}{\p}777 &    34.9 & {\p}0.7 & {\p}2.32 & {\p}2.01 \\
           & { \tt            } &    {\p}64 &    {\p}744 &    {\p}13.7 &    44055 &       {\p}4660 &    39.2 & {\p}4.1 & {\p}3.16 & {\p}2.33 \\
           & { \tt            } &       512 &       6360 &    {\p}41.2 &    41219 &          12427 &    36.7 &    11.5 & {\p}7.82 & {\p}3.75 \\
           & { \tt bfloat16   } & {\p}{\p}1 & {\p}{\p}17 & {\p}{\p}5.3 &    30006 &    {\p}{\p}189 &    26.9 & {\p}0.3 & {\p}0.92 & {\p}0.82 \\
           & { \tt            } & {\p}{\p}8 & {\p}{\p}99 & {\p}{\p}6.5 &    41284 &       {\p}1239 &    36.7 & {\p}1.2 & {\p}1.00 & {\p}0.80 \\
           & { \tt            } &    {\p}64 &    {\p}739 & {\p}{\p}9.4 &    44348 &       {\p}6835 &    39.4 & {\p}6.4 & {\p}1.85 & {\p}0.98 \\
           & { \tt            } &       512 &       6348 &    {\p}34.9 &    41297 &          14677 &    36.7 &    13.1 & {\p}6.54 & {\p}2.41 \\

\bottomrule
\end{tabular}

\end{table}

\clearpage

\section{Hyper-parameters}
\label{sec:app:hparams}
For each transfer task, we provide the exact hyper-parameters with which the final score and the publicly released checkpoint was obtained.
Tuning was done by looking at a validation-set if available, otherwise on a small ``minival'' held out from the training set.
Most tuning was done at 224px resolution, and either carried over as-is to higher resolutions, or further tuned only slightly at higher resolution.
Zero-shot evaluations were done on the checkpoint obtained for the fine-tune task and using hyper-parameters selected based on the fine-tune task validation metric. This can lead to worse performance as the hyper-parameters were not selected for generalization to different test distributions.
The full configuration of every transfer task can be found in the provided source-code.

\newcommand{\oddrow}{}
\newcommand{\evenrow}{\rowcolor{lightgray!20}}
{\scriptsize
\begin{longtable}{cccccccccc}
\toprule
\multirow{2}{*}{Task} & \multirow{2}{*}{Res} & \multirow{2}{*}{Epochs} & Batch & Learning & Weight & LLM & Label & Freeze & Decode \\
 & &  & size & rate & decay & dropout & smoothing & ViT \\
\midrule
\endhead
\bottomrule
\endfoot
\oddrow  COCOcap         & 224   & 5     & 256   & 1e-05 & 1e-06 & 0.0   & 0.0   & False & beam n=2 \\
\oddrow                  & 448   & 5     & 256   & 1e-05 & 1e-06 & 0.0   & 0.0   & False & beam n=2 \\
\evenrow COCO-35L        & 224   & 25 & 256   & 1e-05 & 1e-06 & 0.0   & 0.0   & False & greedy \\
\evenrow                 & 448   & 25 & 256   & 1e-05 & 1e-06 & 0.0   & 0.0   & False & greedy \\
\oddrow  Screen2Words    & 224   & 10    & 256   & 1e-05 & 0.0   & 0.3   & 0.2   & False & greedy \\
\oddrow                  & 448   & 10    & 256   & 1e-05 & 0.0   & 0.3   & 0.2   & False & greedy \\
\evenrow TextCaps        & 224   & 5     & 256   & 1e-05 & 1e-06 & 0.0   & 0.0   & False & beam n=3 \\
\evenrow                 & 448   & 5     & 256   & 1e-05 & 1e-06 & 0.0   & 0.0   & False & beam n=3 \\
\oddrow  SciCap          & 224   & 80    & 256   & 3e-05 & 0.0   & 0.1   & 0.1   & False & greedy \\
\oddrow                  & 448   & 80    & 256   & 3e-05 & 0.0   & 0.1   & 0.1   & False & greedy \\
\evenrow WidgetCap       & 224   & 4     & 64    & 3e-06 & 3e-07 & 0.1   & 0.1   & False & greedy \\
\evenrow                 & 448   & 4     & 64    & 3e-06 & 3e-07 & 0.1   & 0.1   & False & greedy \\
\oddrow  VQAv2           & 224   & 10    & 256   & 1e-05 & 1e-06 & 0.0   & 0.0   & False & greedy \\
\oddrow                  & 448   & 10    & 256   & 1e-05 & 0.0   & 0.0   & 0.0   & False & greedy \\
\evenrow OKVQA           & 224   & 10    & 128   & 5e-06 & 0.0   & 0.0   & 0.0   & False & greedy \\
\evenrow                 & 448   & 10    & 128   & 5e-06 & 0.0   & 0.0   & 0.0   & False & greedy \\
\oddrow  AOKVQA-MC       & 224   & 15    & 128   & 5e-06 & 0.0   & 0.0   & 0.0   & False & greedy \\
\oddrow                  & 448   & 15    & 128   & 5e-06 & 0.0   & 0.0   & 0.0   & False & greedy \\
\evenrow AOKVQA-DA       & 224   & 10    & 128   & 5e-06 & 0.0   & 0.0   & 0.0   & False & greedy \\
\evenrow                 & 448   & 10    & 128   & 5e-06 & 0.0   & 0.0   & 0.0   & False & greedy \\
\oddrow  GQA             & 224   & 1     & 256   & 1e-05 & 0.0   & 0.0   & 0.0   & False & greedy \\
\oddrow                  & 448   & 1     & 256   & 1e-05 & 0.0   & 0.0   & 0.0   & True  & greedy \\
\evenrow NLVR2           & 224   & 3     & 256   & 1e-05 & 1e-06 & 0.0   & 0.0   & False & greedy \\
\evenrow                 & 448   & 10    & 256   & 3e-06 & 3e-07 & 0.0   & 0.0   & False & greedy \\
\oddrow  AI2D            & 224   & 10    & 256   & 1e-05 & 1e-06 & 0.0   & 0.0   & False & greedy \\
\oddrow                  & 448   & 10    & 256   & 1e-05 & 1e-06 & 0.0   & 0.0   & False & greedy \\
\evenrow ScienceQA       & 224   & 20    & 128   & 1e-05 & 0.0   & 0.0   & 0.0   & True  & greedy \\
\evenrow                 & 448   & 20    & 128   & 1e-05 & 0.0   & 0.0   & 0.0   & True  & greedy \\
\oddrow  RSVQA-lr        & 224   & 3     & 256   & 3e-06 & 0.0   & 0.0   & 0.2   & False & greedy \\
\oddrow                  & 448   & 3     & 256   & 3e-06 & 0.0   & 0.0   & 0.2   & False & greedy \\
\evenrow RSVQA-hr        & 224   & 1     & 256   & 1e-05 & 0.0   & 0.0   & 0.0   & False & greedy \\
\evenrow                 & 448   & 1     & 256   & 1e-05 & 0.0   & 0.0   & 0.0   & False & greedy \\
\oddrow  ChartQA         & 224   & 30    & 256   & 1e-05 & 1e-06 & 0.1   & 0.2   & False & greedy \\
\oddrow                  & 448   & 30    & 256   & 1e-05 & 1e-06 & 0.1   & 0.2   & False & greedy \\
\evenrow VizWizVQA       & 224   & 10    & 256   & 1e-05 & 0.0   & 0.0   & 0.0   & False & greedy \\
\evenrow                 & 448   & 10    & 256   & 1e-05 & 0.0   & 0.0   & 0.0   & False & greedy \\
\oddrow  TallyQA         & 224   & 2     & 256   & 1e-05 & 0.0   & 0.0   & 0.0   & False & greedy \\
\oddrow                  & 448   & 2     & 256   & 1e-05 & 1e-06 & 0.0   & 0.0   & False & greedy \\
\evenrow OCR-VQA         & 224   & 3     & 128   & 3e-06 & 0.0   & 0.0   & 0.0   & False & greedy \\
\evenrow                 & 448   & 3     & 128   & 3e-06 & 0.0   & 0.0   & 0.0   & False & greedy \\
\evenrow                 & 896   & 3     & 128   & 1e-05 & 0.0   & 0.0   & 0.0   & False & greedy \\
\oddrow  TextVQA         & 224   & 5     & 256   & 3e-06 & 0.0   & 0.0   & 0.0   & False & greedy \\
\oddrow                  & 448   & 10    & 256   & 3e-06 & 3e-07 & 0.0   & 0.0   & False & greedy \\
\oddrow                  & 896   & 10    & 256   & 3e-06 & 0.0   & 0.0   & 0.0   & False & greedy \\
\evenrow DocVQA          & 224   & 10    & 256   & 1e-05 & 1e-06 & 0.0   & 0.0   & False & greedy \\
\evenrow                 & 448   & 10    & 256   & 1e-05 & 1e-06 & 0.0   & 0.0   & False & greedy \\
\evenrow                 & 896   & 10    & 256   & 1e-05 & 1e-06 & 0.0   & 0.0   & False & greedy \\
\oddrow  InfoVQA         & 224   & 3     & 256   & 1e-05 & 1e-06 & 0.0   & 0.4   & False & greedy \\
\oddrow                  & 448   & 3     & 128   & 1e-05 & 1e-06 & 0.0   & 0.4   & False & greedy \\
\oddrow                  & 896   & 3     & 32    & 3e-06 & 3e-07 & 0.0   & 0.4   & False & greedy \\
\evenrow ST-VQA          & 224   & 3     & 256   & 1e-05 & 1e-06 & 0.0   & 0.1   & False & greedy \\
\evenrow                 & 448   & 3     & 128   & 1e-05 & 1e-06 & 0.0   & 0.1   & False & greedy \\
\evenrow                 & 896   & 3     & 32    & 3e-06 & 3e-07 & 0.0   & 0.1   & False & greedy \\
\oddrow  RefCOCO         & 224   & 100   & 256   & 3e-05 & 0.0   & 0.1   & 0.3   & False & greedy \\
\oddrow                  & 448   & 100   & 256   & 1e-05 & 0.0   & 0.0   & 0.3   & False & greedy \\
\oddrow                  & 896   & 100   & 64    & 1e-05 & 0.0   & 0.0   & 0.3   & False & greedy \\
\evenrow ActivityNet-QA  & 224   & 1     & 128   & 1e-05 & 1e-06 & 0.0   & 0.0   & False & greedy \\
\oddrow  ActivityNet-CAP & 224   & 1     & 128   & 1e-05 & 1e-06 & 0.0   & 0.0   & True  & greedy \\
\evenrow MSRVTT-QA       & 224   & 1     & 128   & 1e-05 & 0.0   & 0.0   & 0.0   & True  & greedy \\
\oddrow  MSRVTT-CAP      & 224   & 20    & 128   & 1e-05 & 0.0   & 0.0   & 0.0   & True  & greedy \\
\evenrow MSVD-QA         & 224   & 1     & 128   & 3e-06 & 3e-07 & 0.0   & 0.0   & False & greedy \\
\oddrow  VATEX           & 224   & 10    & 128   & 3e-06 & 3e-07 & 0.0   & 0.0   & False & greedy \\
\end{longtable}
}

\clearpage

\section{Full per-task results of ablations}

\subsection{Pretraining duration}\label{sec:app:pt_duration}
\begin{figure*}[ht]
  \centering
  \includegraphics[width=\textwidth]{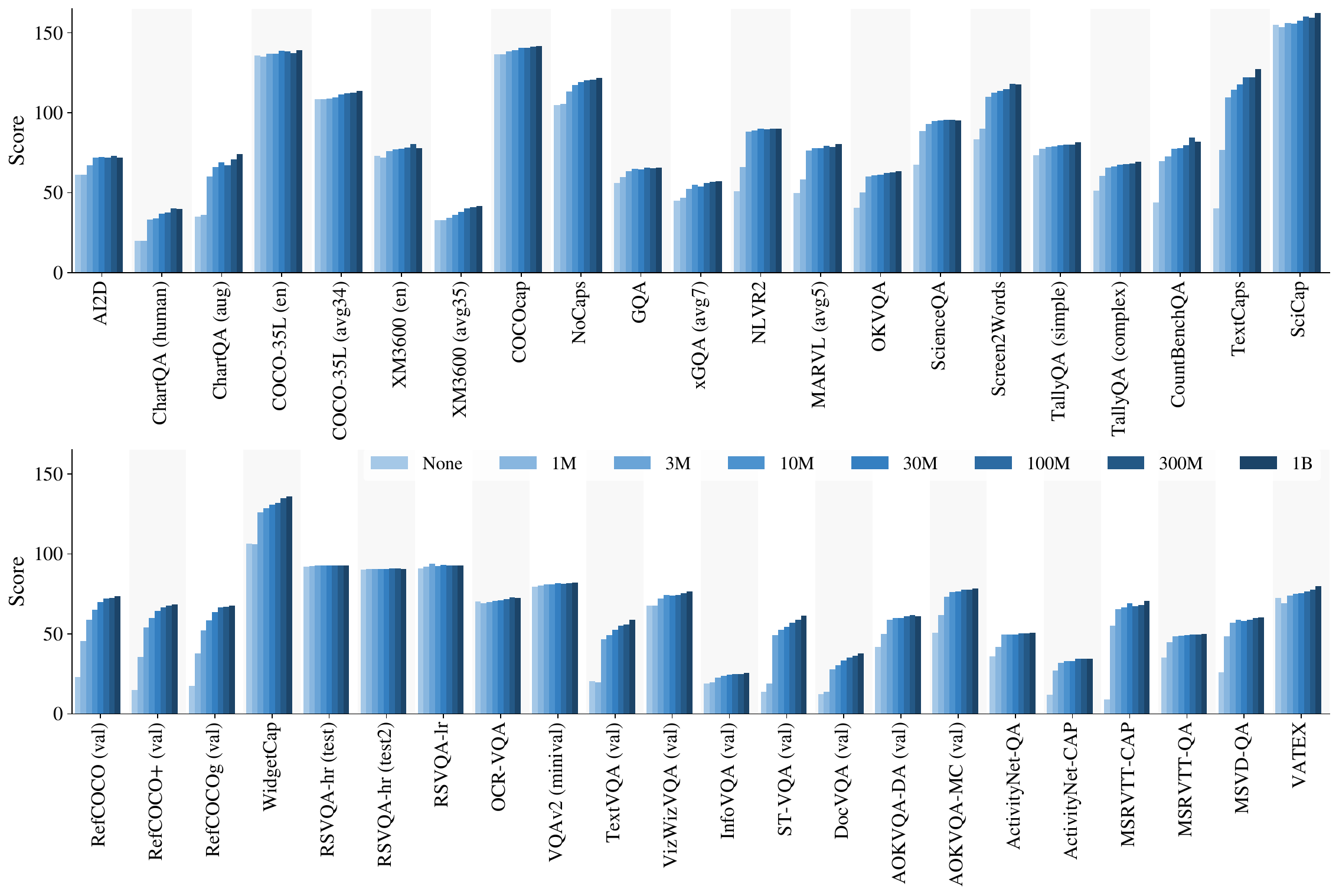}
  \caption{Most tasks benefit significantly from longer pretraining. The notable exception are the remote sensing tasks that have a significantly different image distribution (see examples in Appendix~\ref{sec:app:tasks}). Discussion in Section~\ref{sec:abl:duration}.}
  \label{fig:app:pt_duration}
\end{figure*}

See Figure~\ref{fig:app:pt_duration}.

\clearpage

\subsection{Masking and learning objective}\label{sec:app:learning}
\begin{figure*}[ht]
  \centering
  \includegraphics[width=\textwidth]{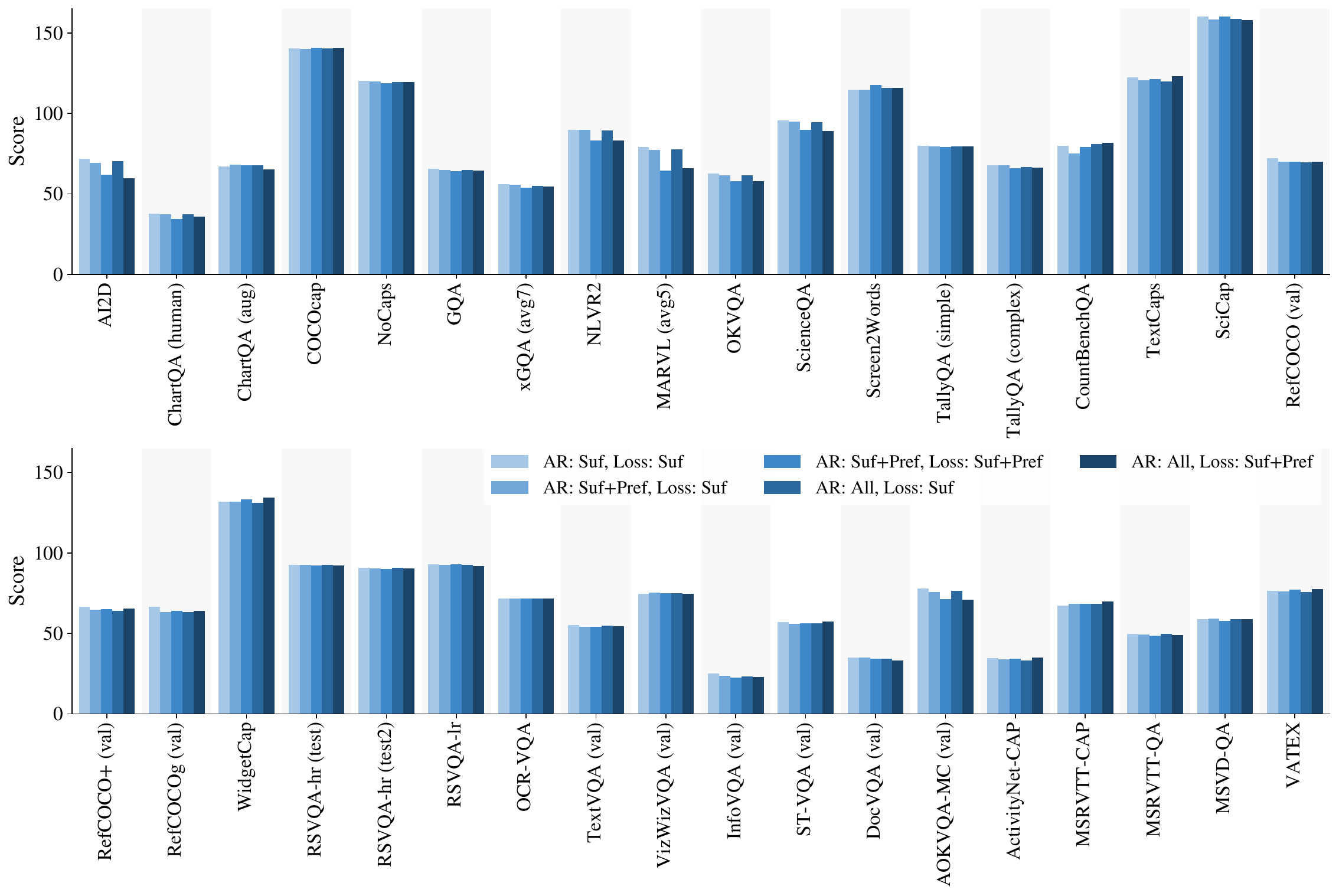}
  \caption{Per task results when changing the learning objective. Discussion in Section~\ref{sec:abl:objective}.}
  \label{fig:app:learning}
\end{figure*}

\begin{figure*}[ht]
  \centering
  \includegraphics[width=\textwidth]{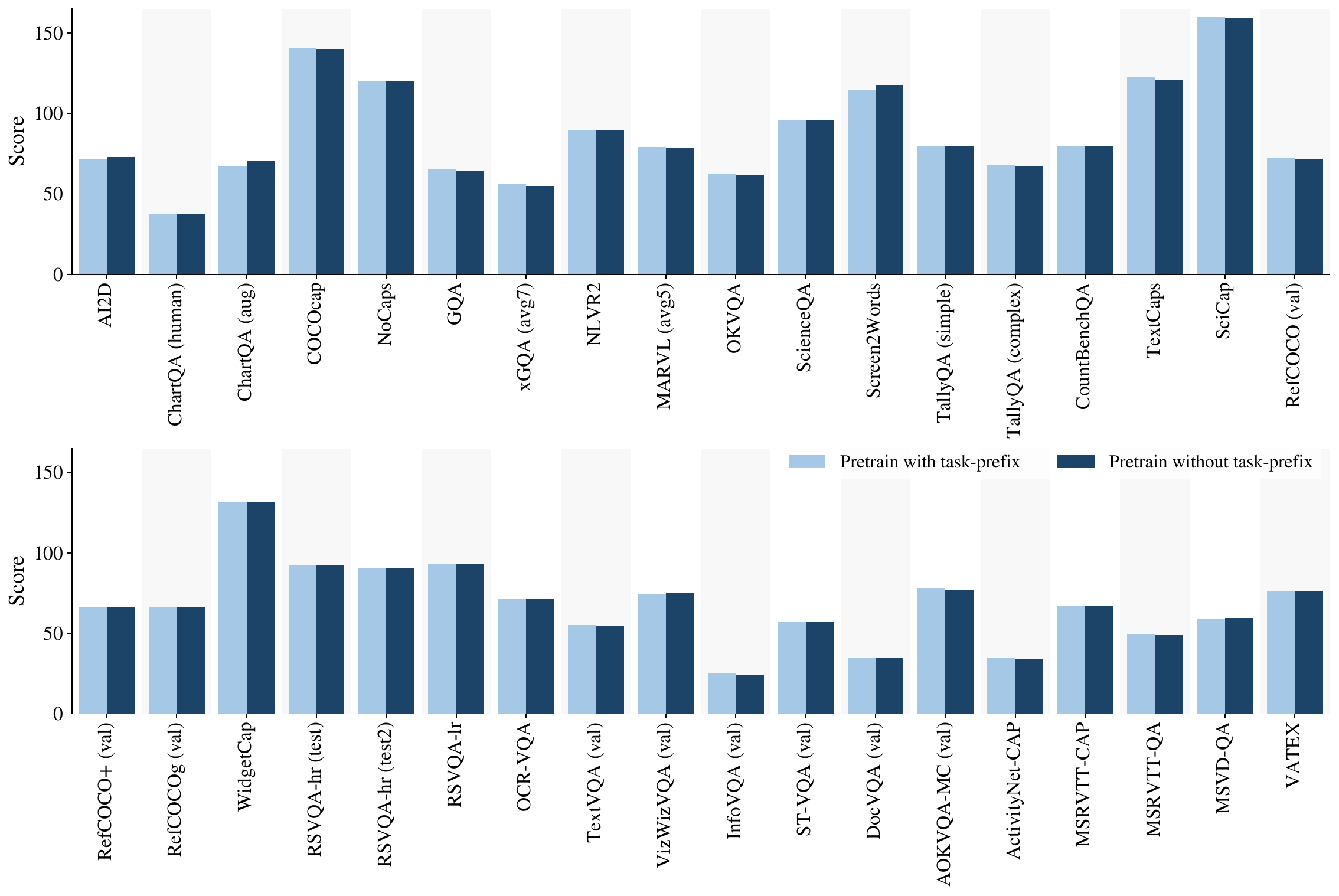}
  \caption{Per task results pretraining with or without task-prefix. Discussion in Section~\ref{sec:abl:objective}.}
  \label{fig:app:learning_prefix}
\end{figure*}

See Figure~\ref{fig:app:learning} and Figure~\ref{fig:app:learning_prefix}.

\clearpage

\subsection{To freeze or not to freeze?}\label{sec:app:freeze}
\begin{figure*}[ht]
  \centering
  \includegraphics[width=\textwidth]{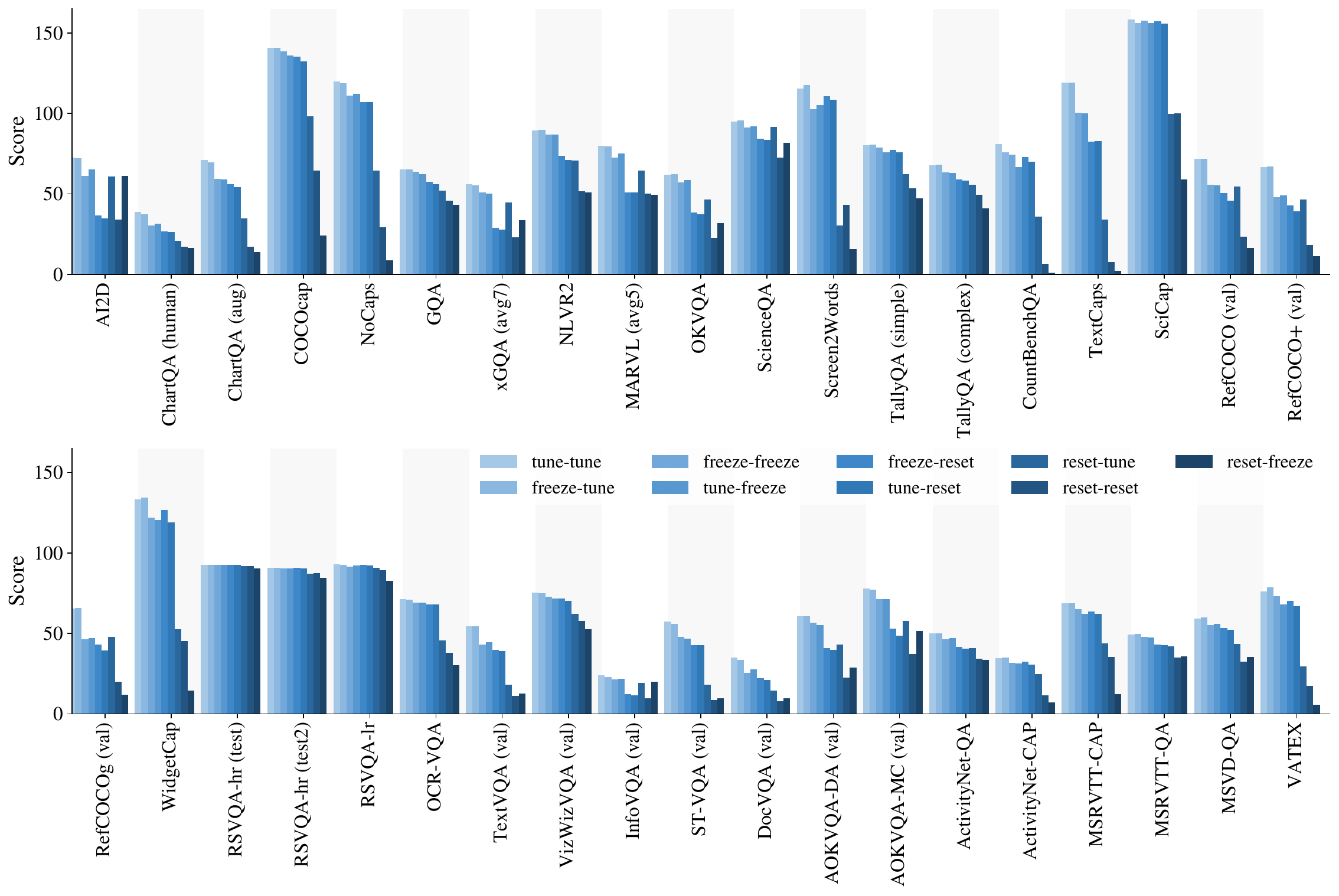}
  \caption{Per task results when using various freezing and resetting patterns. Discussion in Section~\ref{sec:abl:freeze}.}
  \label{fig:app:freeze}
\end{figure*}

See Figure~\ref{fig:app:freeze}.

\clearpage

\subsection{Image encoder: with or without?}\label{sec:app:enc}
\begin{figure*}[ht]
  \centering
  \includegraphics[width=\textwidth]{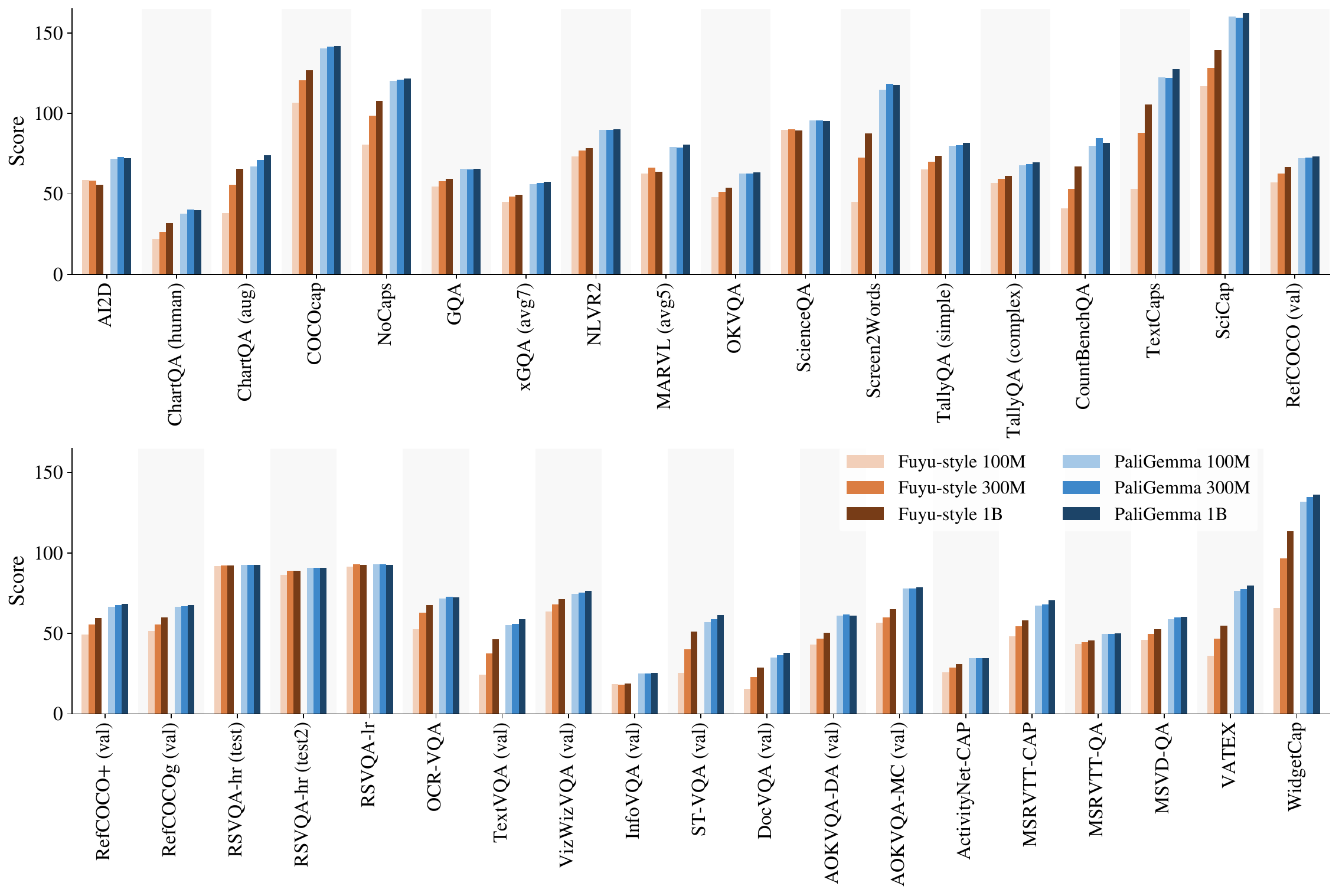}
  \caption{Per task results when using an image encoder vs not using one (fuyu-style). Discussion in Section~\ref{sec:abl:enc}.}
  \label{fig:app:enc}
\end{figure*}

See Figure~\ref{fig:app:enc}.

\clearpage

\subsection{Resolution or sequence length?}\label{sec:app:res_or_seqlen}
\begin{figure*}[ht]
  \centering
  \includegraphics[width=\textwidth]{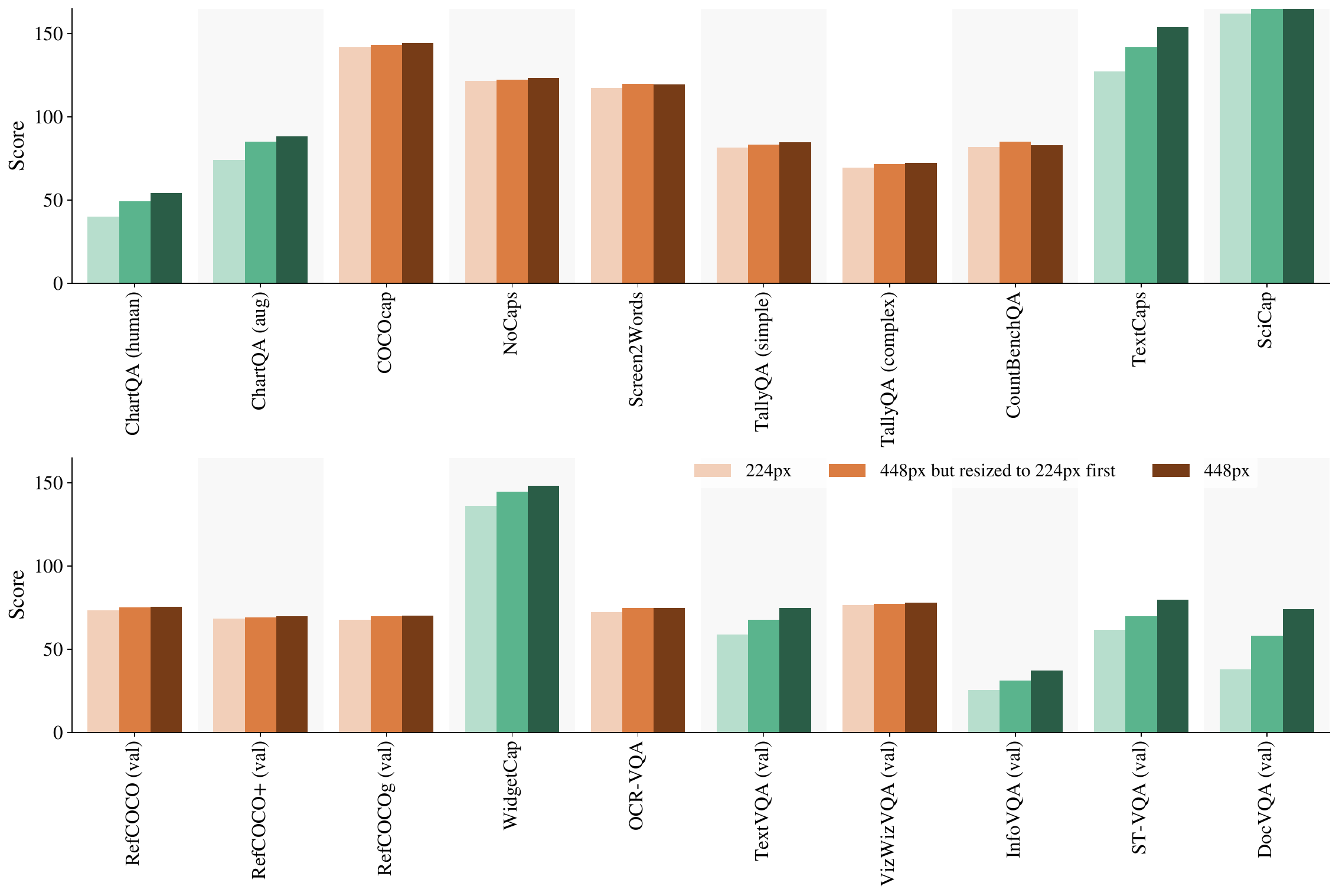}
  \caption{Per task results when the 448px model's input was first resized to 224. Orange are resolution-insensitive tasks, while green are resolution-sensitive tasks. Discussion in Section~\ref{sec:abl:res_or_seqlen}.}
  \label{fig:app:res_or_seqlen}
\end{figure*}

See Figure~\ref{fig:app:res_or_seqlen}.

\clearpage

\subsection{Resolution-specific checkpoints and windowing}\label{sec:app:res_window}
\begin{figure*}[ht]
  \centering
  \includegraphics[width=\textwidth]{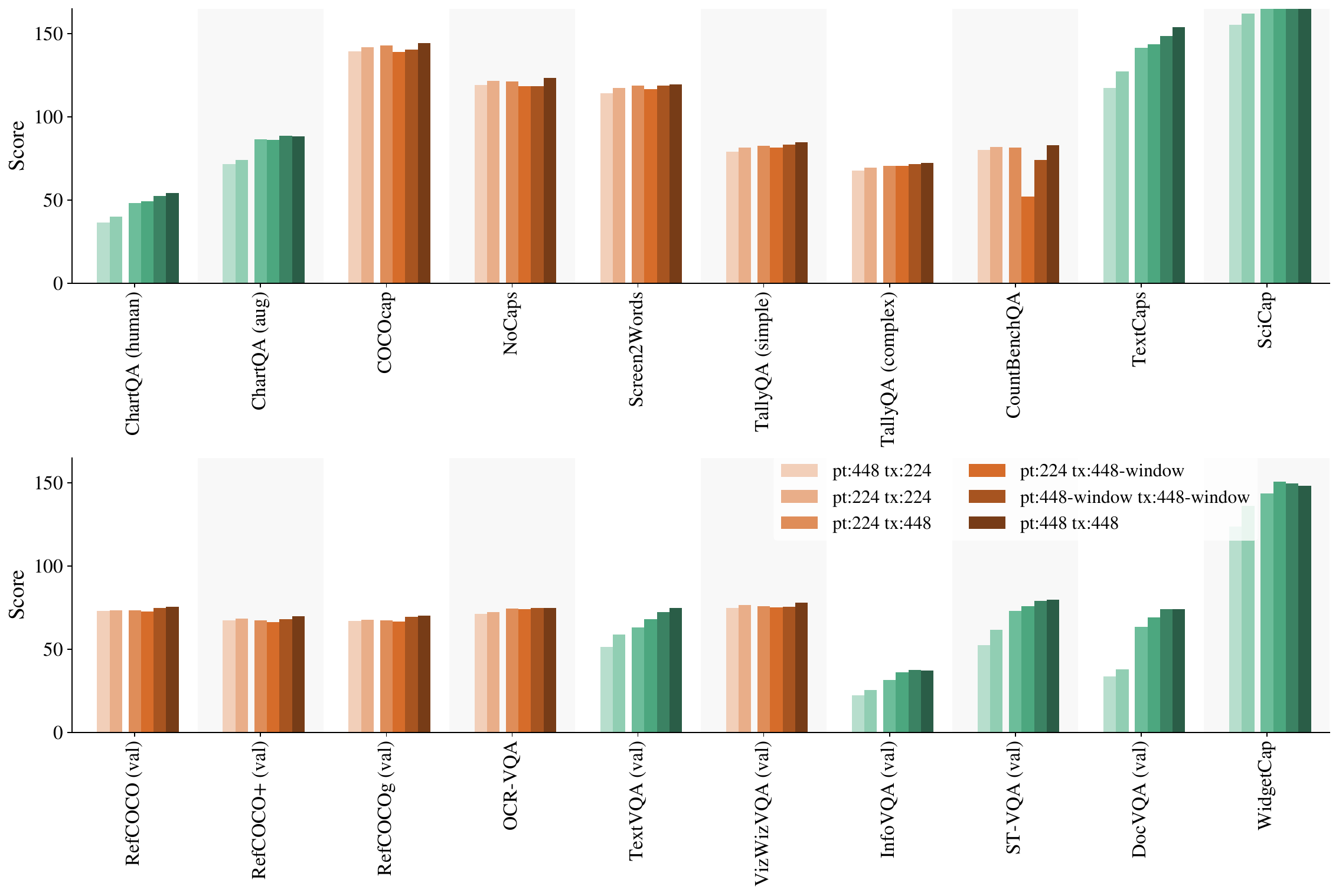}
  \caption{Per task results when the 448px model's input was first resized to 224. Orange are resolution-insensitive tasks, while green are resolution-sensitive tasks. Discussion in Sections~\ref{sec:abl:res_ckpt} and~\ref{sec:abl:windows}.}
  \label{fig:app:res_window}
\end{figure*}

See Figure~\ref{fig:app:res_window}.

\clearpage

\subsection{Stage2 mixture re-weighting}\label{sec:app:s2_reweight}
\begin{figure*}
  \centering
  \includegraphics[width=\textwidth]{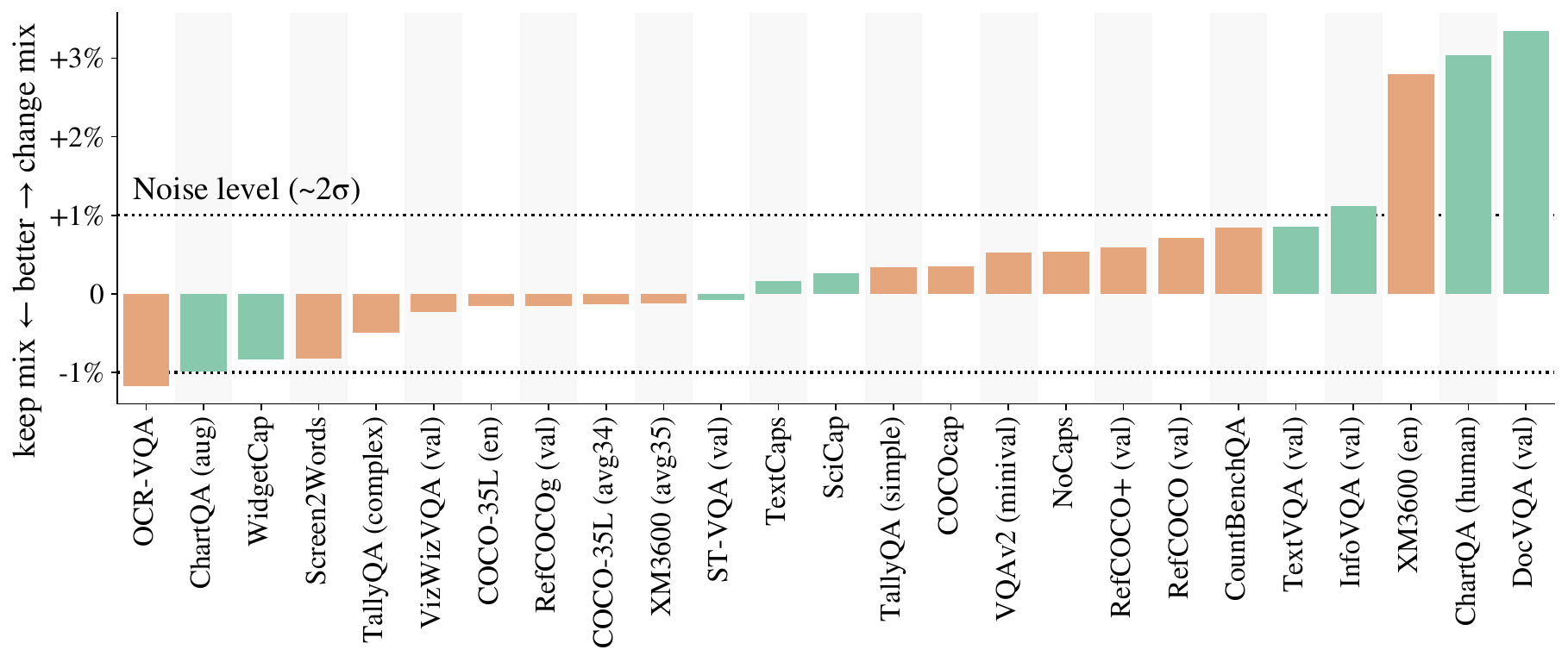}
  \caption{Per task change in result when performing Stage2 with the exact same pretraining task-mixture as Stage1. Most changes in result are below 2 standard-deviations of re-runs. Discussion in Section~\ref{sec:abl:s2mix}.}
  \label{fig:app:s2_reweight}
\end{figure*}

See Figure~\ref{fig:app:s2_reweight}.

\clearpage

\subsection{Transfer with simple hyper parameters}\label{sec:app:transfer_simple}
\begin{figure*}[ht]
  \centering
  \includegraphics[width=\textwidth]{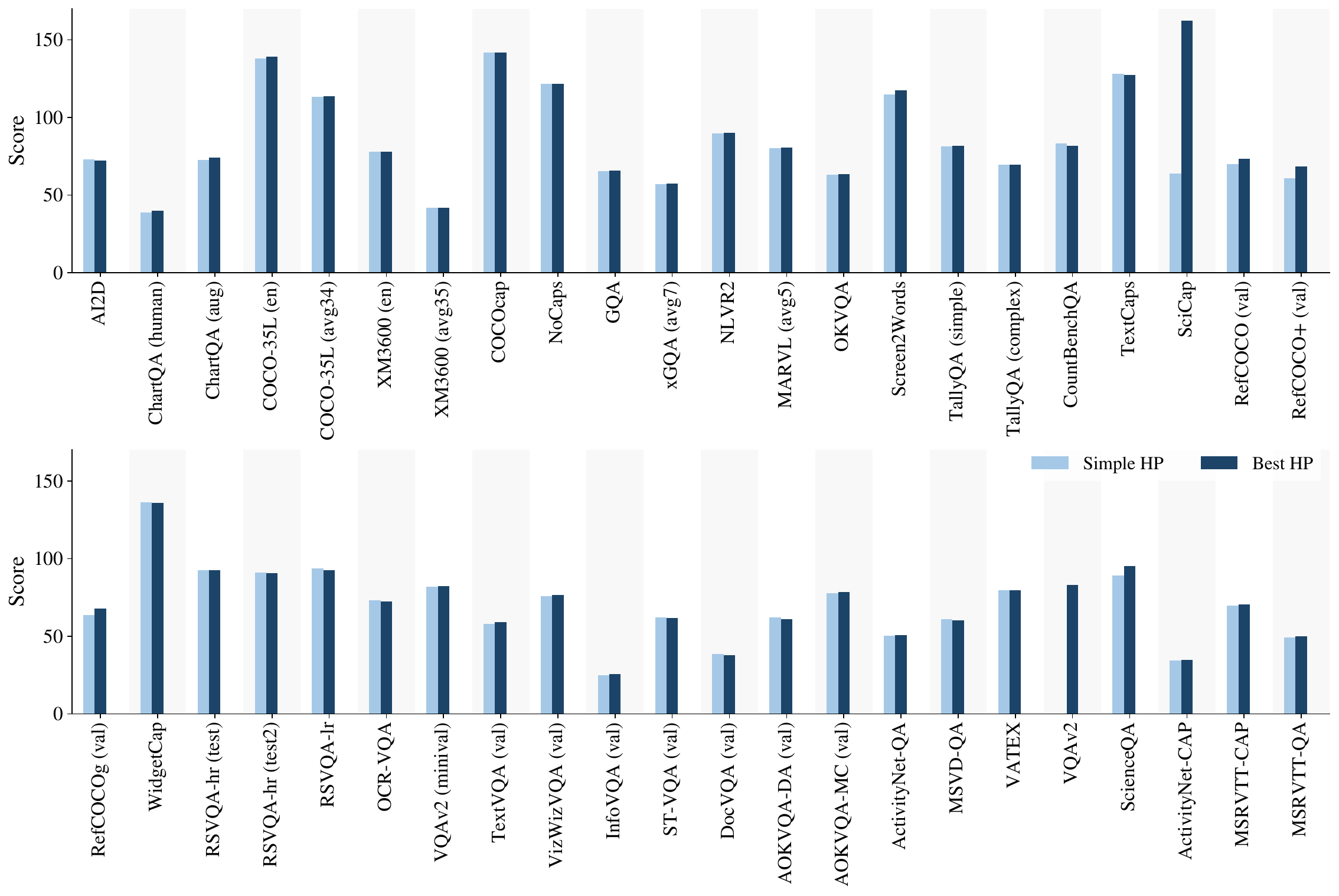}
  \caption{Per task results when transferring with unique  and simple hyper parameter setting for all tasks. For the majority of tasks it works good out of the box.}
  \label{fig:app:transfer_simple}
\end{figure*}

See Figure~\ref{fig:app:transfer_simple}.

\clearpage

\subsection{Transfer with limited examples}\label{sec:app:transfer_lowdata}
\begin{figure*}
  \centering
  \includegraphics[width=\textwidth]{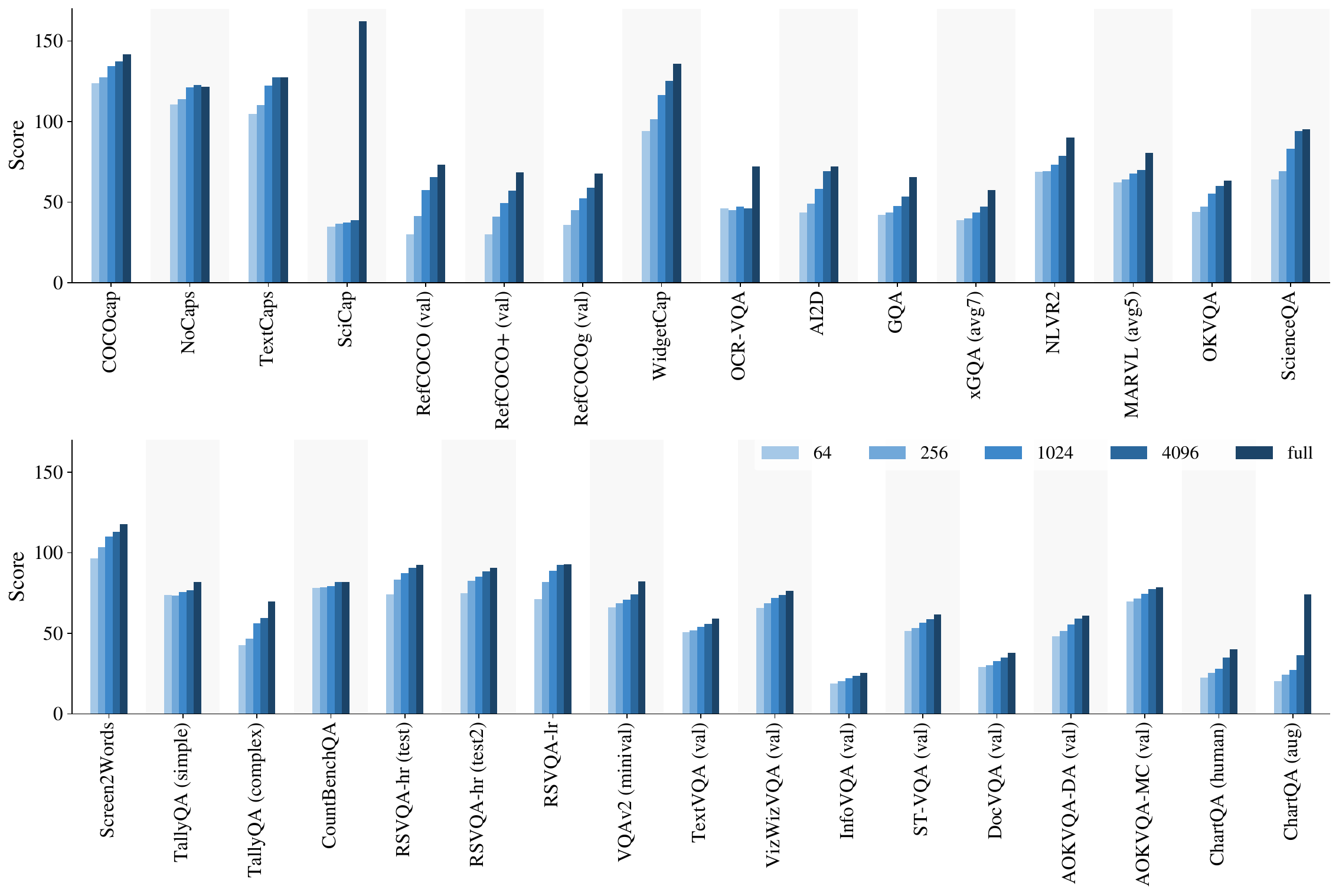}
  \caption{Per task results when transferring with limited number of examples. We report the max of the values obtained by sweeping learning rates, epochs and batch size.}
  \label{fig:app:transfer_lowdata}
\end{figure*}

See Figure~\ref{fig:app:transfer_lowdata}.

\end{document}